\PassOptionsToPackage{unicode}{hyperref}
\PassOptionsToPackage{hyphens}{url}

\documentclass[
]{article}
\usepackage{amsmath,amssymb}
\usepackage[utf8]{inputenc}
\usepackage{iftex}
\ifPDFTeX
  \usepackage[T1]{fontenc}
  \usepackage[utf8]{inputenc}
  \usepackage{textcomp} 
\else 
  \usepackage{unicode-math} 
  \defaultfontfeatures{Scale=MatchLowercase}
  \defaultfontfeatures[\rmfamily]{Ligatures=TeX,Scale=1}
\fi
\usepackage{lmodern}
\ifPDFTeX\else
\fi
\IfFileExists{upquote.sty}{\usepackage{upquote}}{}
\IfFileExists{microtype.sty}{
  \usepackage[]{microtype}
  \UseMicrotypeSet[protrusion]{basicmath} 
}{}
\makeatletter
\@ifundefined{KOMAClassName}{
  \IfFileExists{parskip.sty}{%
    \usepackage{parskip}
  }{
    \setlength{\parindent}{0pt}
    \setlength{\parskip}{6pt plus 2pt minus 1pt}}
}{
  \KOMAoptions{parskip=half}}
\makeatother
\usepackage{xcolor}
\ifLuaTeX
  \usepackage{selnolig}  
\fi
\usepackage{bookmark}
\IfFileExists{xurl.sty}{\usepackage{xurl}}{} 
\hypersetup{
  hidelinks,
  pdfcreator={LaTeX via pandoc}}
\usepackage{graphicx}
\usepackage{caption}
\usepackage{xcolor}
\usepackage{pifont}
\newcommand{\cmark}{\textcolor{green!80!black}{\ding{51}}}
\newcommand{\xmark}{\textcolor{red}{\ding{55}}}

\title{\phantomsection\label{_fm0tygoa1omj}{}\textbf{End-To-End Clinical
Trial Matching with}

\phantomsection\label{_dypn943sok7l}{}\textbf{Large Language Models}}
\author{}
\date{}

\begin{document}
\maketitle
\begin{center}

Dyke Ferber (1, 2), Lars Hilgers (2), Isabella C. Wiest (2, 3),
Marie-Elisabeth Leßmann (2, 4), Jan Clusmann (2, 5), Peter Neidlinger
(2),
Jiefu Zhu (2), Georg Wölflein (6), Jacqueline Lammert (7),
Maximilian Tschochohei (8), Heiko Böhme (9, 10, 11, 12), Dirk Jäger (1),
Mihaela Aldea (13), Daniel Truhn (14),
Christiane Höper (15), Jakob Nikolas Kather (1, 2, 4, +)
\end{center}

\vspace{1cm}

\begin{enumerate}
\def\labelenumi{\arabic{enumi}.}
\item
  Department of Medical Oncology, National Center for Tumor Diseases
  (NCT), Heidelberg University Hospital, Heidelberg, Germany
\item
  Else Kroener Fresenius Center for Digital Health, Technical University
  Dresden, Dresden, Germany
\item
  Department of Medicine II, Medical Faculty Mannheim, Heidelberg
  University, Mannheim, Germany
\item
  \hspace{0pt}\hspace{0pt}Department of Medicine I, University Hospital
  Dresden, Dresden, Germany
\item
  Department of Medicine III, University Hospital RWTH Aachen, Aachen,
  Germany
\item
  School of Computer Science, University of St Andrews, St Andrews,
  United Kingdom
\item
  Department of Gynecology and Center for Hereditary Breast and Ovarian
  Cancer, University Hospital rechts der Isar, Technical University of
  Munich (TUM), Munich, Germany
\item
  Google Cloud, Munich, Germany
\item
  National Center for Tumor Diseases (NCT/UCC), Dresden, Germany
\item
  German Cancer Research Center (DKFZ), Heidelberg, Germany
\item
  Medical Faculty and University Hospital Carl Gustav Carus, TUD Dresden
  University of Technology, Dresden, Germany
\item
  Helmholtz-Zentrum Dresden-Rossendorf (HZDR), Dresden, Germany
\item
  Department of Medical Oncology, Gustave Roussy, Villejuif, France;
  Paris Saclay University, Kremlin Bicêtre, France
\item
  Department of Diagnostic and Interventional Radiology, University
  Hospital Aachen, Germany
\item
  AstraZeneca GmbH, Germany
\end{enumerate}

\vspace{1cm}
+ Corresponding author: jakob-nikolas.kather@alumni.dkfz.de

Jakob Nikolas Kather, MD, MSc

Professor of Clinical Artificial Intelligence

Else Kröner Fresenius Center for Digital Health

Technische Universität Dresden

DE -- 01062 Dresden

Phone: +49 351 458-7558

Fax: +49 351 458 7236

Mail: jakob\_nikolas.kather@tu-dresden.de

ORCID ID: 0000-0002-3730-5348

\section{\texorpdfstring{\textbf{Abstract}}{Abstract}}\label{abstract}

\subsubsection{Background}\label{background}

Identifying suitable clinical trials for cancer patients is crucial to
advance treatment modalities and patient care. However, due to the
inconsistent format of medical free text documents and the often highly
complex logic in the trials eligibility criteria, this process is not
only extremely challenging for medical doctors, but also time-consuming
and prone to errors. This results in insufficient inclusion of oncology
patients in clinical trials, especially in a timely manner. The recent
advent of Large Language Models (LLMs) has demonstrated considerable
potential for interpreting electronic health records (EHRs), suggesting
that they hold great promise to facilitate accurate trial matching at
scale.

\subsubsection{Patients and Methods}\label{patients-and-methods}

We generated 51 realistic oncology-focused patient EHRs. For each, a
database of all 105,600 oncology-related clinical trials worldwide from
clinicaltrials.gov was accessed by GPT-4o to identify a pool of suitable
trial candidates with minimal human supervision. Patient eligibility was
then screened by the LLM on criterion-level across a selection of trials
from the candidate trial pool and compared against a baseline defined by
human experts. We then used criterion-level AI feedback to iterate over
discrepant AI and human results, refining the human ground truth where
necessary.

\subsubsection{Results}\label{results}

Our approach successfully identified relevant, human preselected
candidate trials in 93.3\% of test cases from all trials available
worldwide and achieved a preliminary accuracy of 88.0\% (1,398/1,589)
when matching patient-level information on a per-criterion-basis using
the initial human evaluation as baseline. Utilizing LLM feedback to
interactively re-evaluate human scores revealed that 39.3\% of criteria
that were initially considered incorrect according to the human baseline
were either ambiguous or inaccurately annotated by humans, leading to a
total model accuracy of 92.7\% after refining the human ground truth
eligibility definitions.

\subsubsection{Conclusion}\label{conclusion}

We present an end-to-end pipeline for clinical trial matching using
LLMs, demonstrating high precision in screening for appropriate clinical
trials at scale and matching selected candidate trials with high
precision to individual patients, even outperforming the performance of
qualified medical doctors. Additionally, our pipeline can operate both
fully autonomously or with human supervision and is not intrinsically
restricted to cancer, offering a scalable solution to enhance
patient-trial matching for the real world.

\section{\texorpdfstring{\textbf{Keywords}}{Keywords}}\label{keywords}
\noindent
\small
\small
Clinical Trial Matching \textbullet\ 
Oncology Trials \textbullet\ 
Eligibility Criteria \textbullet\
Artificial Intelligence \textbullet\ 
Large Language Model \textbullet\ 
GPT-4o

\section{\texorpdfstring{\textbf{Introduction}}{Introduction}}\label{introduction}

In oncology, clinical trials serve two purposes: they offer potential
therapeutic benefits to cancer patients across all disease stages,
ranging from early intervention to experimental treatments for those
with limited or exhausted standard care options\textsuperscript{1,2}.
They are also crucial to advance scientific research, as new treatments
can only be approved through rigorous clinical testing. However, the
practical realization of clinical trial enrollments still remains far
from satisfactory, from both patient and clinician perspectives. For
clinicians, identifying suitable trials is often
time-consuming\textsuperscript{3} and complex due to patient-related
factors such as performance status or comorbidities, logistical
challenges like regional trial availability, systemic issues including
lack of access to genomic testing, and the difficulty clinicians might
face in locating available trials, all of which contribute to low
enrollment rates of only 2-3\% of potential trial
candidates\textsuperscript{4}.

Overall, there are three primary reasons for this:

First, the sheer volume of data generated during oncologic treatments,
including hospital stay records, as well as genomic and imaging data,
often accumulate, drastically increasing the burden on
physicians\textsuperscript{5--7}. These data are typically fragmented
and unstructured\textsuperscript{8}, comprising free text, tabular
records, and more\textsuperscript{8}. Secondly, the complexity and
volume of clinical trials tailored to oncology further complicate the
process. There are approximately 500,000 studies registered on
ClinicalTrials.gov, out of which 105,732 are dedicated to patients with
cancer as of May 2024. Like patient records, trial information often
includes unstructured information, such as plain text eligibility
criteria and requires complex logical combinations of disease
conditions, histologic subtypes, molecular markers and
comorbidities\textsuperscript{9}. Thirdly, from a
patient\textquotesingle s perspective, due to the evolution of the
disease and the need to avoid patient attrition from deteriorating
clinical conditions, it is crucial that the time to inclusion and
initiation of treatment in clinical trials is minimized to the shortest
possible duration.

From a practical perspective, addressing these challenges requires
clinicians to follow a two-step process: first, they must screen for
potential trial candidates based on key patient criteria such as tumor
type, stage, mutations and availability within the patient's area of
residence; then, they need to perform detailed one-on-one matches of all
the patient's information with each candidate trial's eligibility
criteria.

So far, computational support tools designed to simplify this process
have focused on only one of these steps at a time. For the first step,
systems have primarily used embedding techniques, where patient and
trial text data are converted into a numerical representation space and
matched based on approximate mathematical
similarity\textsuperscript{10,11}. For the second step, most tools focus
on converting unstructured text from patient records and trial
information into a tabular-like format. For instance, Criteria2Query
uses machine learning and rule-based methods to parse inclusion and
exclusion criteria into a structured format accessible via database
queries\textsuperscript{12}.

Only recently, with advances in generative AI, particularly Large
Language Models (LLMs) like GPT-4\textsuperscript{13}, extracting and
structuring information from medical documents has been drastically
simplified\textsuperscript{14}. The potential of LLMs has also been
explored for matching patients to clinical trials based on comparing
eligibility criteria to patient records\textsuperscript{15}. For
instance, den Hamer et al.\textsuperscript{16} demonstrated that LLMs
can accurately provide eligibility labels such as `yes', `no', or
`unknown' when given both trial information and patient data as input at
the same time. In oncology, Wong et al.\textsuperscript{17} extended
this idea to account for complex logical conditions using a hierarchical
matching procedure, showing that GPT-4 can excel at this task even
without additional training. Fine-tuning LLMs on annotated trial data
has markedly improved their performance even further. This approach has
facilitated the development of a local, privacy-preserving model that
closely rivals the capabilities of proprietary, large cloud-based
LLMs\textsuperscript{18}. Recently, the same research team created
OncoLLM\textsuperscript{19}, a new model that significantly reduces the
performance gap with the current leading model, GPT-4.

Nevertheless, the aforementioned projects have several limitations:

First, they tend to focus on either step one or step two of the process,
rather than integrating both. Additionally, for step one, discrete
criteria such as recruitment status or intervention type take on only
discrete values (e.g. sex, location or recruitment status), which would
more effectively be managed through direct selection or filtering rather
than embedding-based approaches that rely on inexact similarity matches.
Second, all current LLM-based methods heavily rely on narrowly
engineered prompts, which can be lengthy and cumbersome (Wong et
al.\textsuperscript{17} report prompts of up to four pages). Third, due
to the free-text nature in which eligibility criteria and patient
information are processed by the model\textsuperscript{20}, there is no
guarantee that the responses will adhere strictly to the required
criteria structure.

We herein present a fully end-to-end pipeline for clinical trial
matching which we designed to overcome the aforementioned limitations.
Our approach is based on two principles: using LLMs as central reasoning
agents\textsuperscript{21} capable of taking actions and
programmatically enforcing trial eligibility criteria as structured
programming objects rather than plain free text, thereby ensuring the
model consistently outputs validly annotated
information\textsuperscript{22}.~

Our contributions are the following:

\begin{enumerate}
\def\labelenumi{\arabic{enumi}.}
\item
  To the best of our knowledge, we present the first \emph{truly}
  end-to-end pipeline for clinical trial matching, starting with
  searching relevant trial candidates for a given patient from all
  cancer trials available world wide and ending with fully-annotated
  trial eligibility criteria for a relevant set of trials.
\item
  We provide an extensive and profound benchmarking, encompassing 51
  oncology cases and matching over 1,580 single trial criteria that have
  been annotated by five human experts. We provide evidence that our
  pipeline excels in both, reliably filtering relevant trials from tens
  of thousands and providing highly accurate one-on-one eligibility
  matches with criterion-level feedback and explanations for users.
\item
  We demonstrate that LLMs can outperform medical
  doctors in clinical trial matching. Our findings reveal that nearly
  40\% of the initially contradictory answers between GPT-4o and
  physicians were accepted as valid responses upon refining the human
  baseline with criterion-level AI feedback, resulting in an overall
  criterion-level accuracy of 92.7\% for our pipeline.
\item
  By enforcing trial eligibility criteria as structured programming
  objects rather than relying on them as free-text inputs, we guarantee
  that the LLM always outputs precisely and validly annotated
  information.
\end{enumerate}

\section{\texorpdfstring{\textbf{Methods}}{Methods}}\label{methods}

\subsubsection{Clinical Trial
Composition}\label{clinical-trial-composition}

Data was sourced from ClinicalTrials.gov on May 13, 2024, by filtering
for the Condition/disease ``\emph{cancer}'', yielding a total of 105,600
registered clinical trials, provided in a JavaScript-Object Notation
(JSON) file. Subsequently, we programmatically filter each clinical
trial by selecting relevant metadata, including fields like recruitment
status, available centers (locations) or allowed disease conditions.
Next, to allow the generation of vector embeddings from free text, we
combine several metadata fields such as the brief and official titles,
detailed trial descriptions and brief summaries into a structured plain
text field.

\subsubsection{Database Generation}\label{database-generation}

When finding appropriate clinical trials for patients, physicians most
often need to initially filter by specific, structured criteria like the
locations of participating centers, recruitment status or allowed
disease conditions, while then also examining free text descriptions to
precisely match patient conditions to exclusion and inclusion criteria.
From a computational perspective, we are thus confronted with the fact
that certain attributes, such as discrete metadata fields that have a
set of discrete allowed options, require exact matches, whereas others
need to be matched based on free text: This requires handling the issue
of synonyms - such as recognizing that ``lung metastases'' and
``pulmonary metastases'' are equivalent - where exact pattern matches
are unsuitable.

To address these issues, we developed a hybrid database that effectively
combines exact field matching with vector proximity search to find
clinical trials that most closely correspond to patient descriptions in
representation space. For the former, we employed a local instance of a
No-SQL database\textsuperscript{23} (MongoDB), which offers several
advantages in this context, including high scalability, a flexible
schema design for sending nested requests, and robust performance when
handling large datasets. Next, we generate vector embeddings - numerical
representations - of the free, preprocessed text for each clinical trial
using the ``\emph{BAAI/bge-large-en-v1.5}'' embedding model locally,
producing vector embeddings with a dimensionality of 768 from text with
a maximum of up to 512 tokens each. As clinical trial information is
most often considerably longer, we performed text splits, including a
50-character overlap to ensure comprehensive coverage and avoid
information loss, such as by splitting text in the middle of a sentence.
We store all text embeddings in a local collection of a vector database
(ChromaDB\textsuperscript{24}) for efficient similarity search, using
cosine distance as the default search metric throughout our experiments.

\subsubsection{Clinical Case Generation}\label{clinical-case-generation}

Our experiments are based on published synthetic cases by Benary et
al.\textsuperscript{25}, which include ten fictional patient vignettes
representing seven different tumor types, primarily lung adenocarcinoma
(four cases), each annotated with various mutations (59 in total). To
create a more realistic setting, we extended these cases to full medical
EHR reports, using in-house original patient reports as templates, and
including clinical descriptions of patient diagnoses, comorbidities,
molecular information, short imaging descriptions from staging CT or MRI
scans and patient history at different levels of detail. To ensure
reliable matching of patients to existing clinical trials, we initially
selected candidate trials through manual search or by utilizing those
approved by a molecular tumor board from Lammert et
al\textsuperscript{26}. This led us to generating a total of 15 patient
cases, which we refer to as base cases in the following. We then aligned
the clinical case descriptions to either meet or contain conflicts with
the respective trial eligibility criteria. This procedure was performed
by first manually crafting patient reports based on medical expertise,
then utilizing ChatGPT (GPT-4) for iterative refinement of style,
grammar and language flow, leading to a total of 51 case vignettes. The
final versions of these were evaluated for clinical realism,
completeness and linguistic authenticity, and were approved by one
physician with expertise in oncology before performing the experiments.

\subsubsection{Trial Matching Pipeline
Specifications}\label{trial-matching-pipeline-specifications}

The pipeline consists of two main components: the hybrid
No-SQL-\&-Vector database and an LLM that acts at its core to
sequentially orchestrate database search, trial retrieval and finally
trial matching with patient information. We utilized the
``\emph{GPT-4o}'' model through the OpenAI integration in Python. Model
hyperparameters were kept at their default settings. As the LLM operates
programmatically to access the database, its outputs cannot be plain
text, but need to be valid programmatic data types and occasionally also
adhere to certain constraints, such as belonging to a fixed, discrete
set of options (like current recruitment status of a trial). Otherwise,
invalid requests would lead to failures in accessing the database. We
therefore constrain model output types by setting type hints in
pydantic\textsuperscript{27}.

\begin{figure}[ht!]
  \centering
  \includegraphics[width=\linewidth]{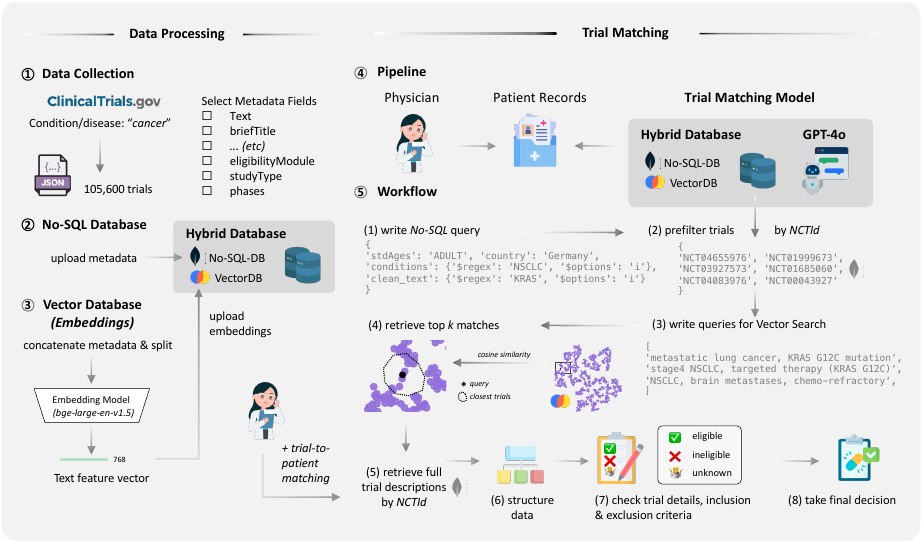}
  \caption{\textbf{High-level overview of the proposed trial matching
framework.}
This figure provides an overview of the entire trial matching pipeline,
divided into the data processing steps (left) and the workflow at
inference time (trial matching, right).
We begin with downloading clinical trial data from ClinicalTrials.gov in
structured JSON format (1). We define metadata fields used for No-SQL
database uploads and consecutive lookup operations (in our case, we use
MongoDB) (2). Additionally we select metadata fields such as detailed
trial titles, detailed descriptions and trial summaries that we
concatenate into plain text. Next, the embedding model converts
tokenized text into numerical representations (vectors), which we store
for similarity search in a vector database (ChromaDB), forming the final
hybrid DB structure (3). To accommodate the context window of the
embedding model (BAAI/bge-large-en-v1.5 in our case has a limit of 512
tokens), we split the concatenated text with a 50-character overlap
before forwarding it into the model.
The final model comprises the hybrid database and an LLM (GPT-4o) that
generates database requests (4). Next, we illustrate the entire pipeline
end-to-end (5), starting from a given (cancer) patient record that
encompasses clinical diagnoses, comorbidities, molecular alterations,
and patient history, from which the LLM extracts relevant information to
prefilter only valid trials via a No-SQL request (Step 1). Next, it
rewrites the patient information into \emph{n} short, precise queries
for search over the vector database, where we measure the cosine
distance between the query and all embedded clinical trial texts (Step
3) on the embedding level. From this procedure, only the top \emph{k}
closest trial samples are returned, with the additional constraint that
they must be within the prefiltered set of trial IDs (Step 4). From
these, the original trial descriptions, including their metadata are
re-instantiated (Step 5). GPT-4o then inspects the suitability of a
patient for a given trial by element-wise evaluation of the available
patient information against the descriptions and criteria from the
respective trial (Step 7). The trial can then be reviewed with
criteria-level decisions and model explanations by the physician (Step
8).}
  \label{fig:Figure 1}
\end{figure}

The entire pipeline, which we illustrate in \textbf{Figure 1}, consists
of a sequential chain of LLM requests, where each LLM call is executed
as a structured Chain-of-Thought (CoT) module: Upon invocation with a
plain text description of a cancer patient and a user instruction (with
varying levels of detail), the LLM extracts relevant metadata to
prefilter the database in a No-SQL fashion. For all discrete attributes,
we provide all available options as enforced type hints in a zero-shot
manner; for open-ended free text search terms like disease conditions or
keywords (for instance to filter free text for specific mutations) we
provide manually crafted, few-shot examples. Next, each LLM output is
converted into a valid No-SQL database query to extract all matching
trials by their National Clinical Trial identifier (NCTId).
Subsequently, to enhance the diversity of the retrieval step and remove
uninformative information from the patient descriptions, GPT-4o is
instructed to generate a maximum of five different queries from the main
patient information, where the retrieval is constrained to filter only
from the preselected pool of trials by NCTId. This step is performed
iteratively over all queries. We end up with a collection of \emph{n}
top-matching trials from the vector search, from which we use NCTIds to
retrieve full trial information. As outlined later, we additionally
experimented with using reranking (Cohere
rerank-english-v3.0\textsuperscript{28}), which we omit from our final
pipeline due to lack of additional benefits. Instead, for each trial,
the LLM processes the fields containing brief trial summaries and
detailed descriptions and discards trials that are deemed irrelevant to
the patient. It then structures eligibility criteria programmatically
with up to two levels of nested conditions and performs an
element-by-element match of the structured inclusion and exclusion
criteria to the patient information, returning only boolean values (True
if patient is eligible according to a criterion, False otherwise) or
unknown if the information provided to the LLM was insufficient to make
a decision. To guide the model's response in handling edge cases, we
define few-shot examples: For instance, if a potential comorbidity is
not mentioned in the patient's EHR, the model is instructed to assume
its absence unless the eligibility criteria require explicit exclusion.
However, if any documented symptoms or indications in the EHR make the
comorbidity plausible, the model should indicate that the information is
insufficient (unknown). Additionally, for each single criterion, we
receive an explanation by the model based on Chain-of-Thought reasoning.

One constraint we make during testing the model is that we permit it to
include trials in an active but not currently recruiting status as an
explicit design choice to ensure consistency with the trials described
previously\textsuperscript{26}.

\subsubsection{Human evaluations}\label{human-evaluations}

Evaluations of all 51 trial candidates were conducted by five
professionals experienced in medical oncology. To ensure one-to-one
matches, the same criteria splits defined by GPT-4o for each trial were
used for human annotations, with evaluations categorized as eligible,
not eligible, or unknown. These ratings were performed using a
browser-based interface that provided access to the full patient EHR,
the trial NCTId, the trial's official title, brief summary, and GPT-4o
structured inclusion and exclusion criteria (Supplementary Figure 1).
Each human evaluator worked independently, with results later aggregated
using a majority vote as the aggregation criterion. During the second
stage, where discrepant AI-human results were compared, we collected
consensus responses through discussions of the model's criterion-level
explanations among three of the original evaluators, leading to either
acceptance or rejection of the model's response.

\section{\texorpdfstring{\textbf{Results}}{Results}}\label{results-1}

\subsubsection{Target Trial Identification
Performance}\label{target-trial-identification-performance}

We hypothesized that the process of filtering relevant trials on
clinicaltrials.gov could be optimized using GPT-4o to write No-SQL
database queries, thereby reducing the manual burden on physicians. We
evaluate this idea on a subset of 15 base cases from our EHR collection,
using either clinical trials from Lammert et al\textsuperscript{26} or
potential target trials manually identified from clinicaltrials.gov. All
prompts are provided in Supplementary Table 1.

\begin{figure}[ht!]
  \centering
  \includegraphics[width=\linewidth]{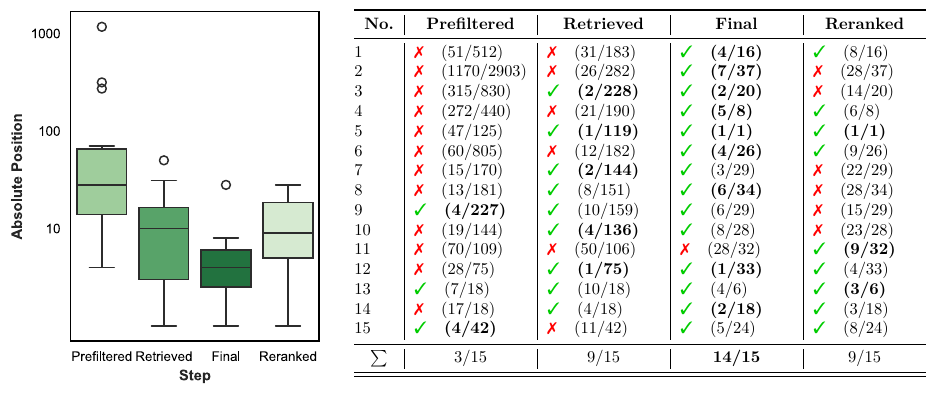}
  \caption{\textbf{Absolute and top-k positions of target trials along
the search process.} We illustrate the average (left) and single (right) positions of 15 trial candidates, one for each of the patient base cases during the entire search process. These trial candidates were either selected
manually or taken from the work of Lammert et al. At
each step, only candidate trials from the previous step are used to
further narrow down the collection of trials and identify the ones with
best potential matching. A checkmark (\cmark) indicates that an expected
trial is found within the top ten possible candidates, while a cross (\xmark)
signifies that the trial does not fall within the top 10. In brackets,
we denote the absolute position of the trial first, followed by the
total number of trials that are selected as potential candidates for
each step. In instances where GPT-4o employs multi-query retrieval, we
aggregate the total number of trials and select the best (lowest) target
trial position at each step.
Herein \emph{prefiltering} refers to the initial selection of trials
from the entire database of 105,600 oncology trials by No-SQL query
selection, based on categorical criteria like study type, overall
status, and locations and trial specific free text search terms
including keywords (primarily mutations) and conditions (mostly disease
types). \emph{Retrieved} denotes trials that are selected through vector
search via embedding lookup. \emph{Final} denotes selecting the top n of
closest trials and prescreening of relevant trials through GPT-4o by
comparing patient summaries and trial brief descriptions. Lastly,
\emph{reranking} is performed using the Cohere rerank-english-v3.0
model. As an example, in case 4, after prefiltering, the model
identified 440 potential candidates, with the relevant target trial
positioned at 272. At the final step, the pool was narrowed down to 8
trials, with the target trial ranked 5th. The lowest positions across
all steps, indicating the highest similarity to the patient's case, are
highlighted in bold. Our results show that retrieval and model selection
are the crucial steps in downsampling candidate trials and prioritizing
on the most relevant ones. We highlight this finding by evaluating the
trial lookup process by the number of expected target trials that are
represented in the top 10 trials found by the LLM (right). Notably,
while prefiltering is crucial for narrowing down the number of trials to
be evaluated from an initial 105,600, it has only a negligible impact on
the relevance ranking of trials. This ranking is predominantly
determined by the cosine similarity search during the retrieval phase
and GPT-4o prescreening (\emph{final}), as it positions all of the
relevant trial candidates among the top ten trials for each of the
respective patient queries except one. Unexpectedly, reranking has no
additional effect on the sorting of relevant trials, such that we omit
it during all further work in this project. This process ends in a final
collection of top trial candidates that are further passed to one-on-one eligibility criteria checks.}
  \label{fig:Figure 2}
\end{figure}

Our results indicate that using GPT-4o is sufficient to write a No-SQL
query that filters all (15/15) potentially relevant trials - those that
were preselected via manual search by humans for each patient base case
- thus narrowing the initial pool of over 100,000 trials to a few
hundred candidates (\textbf{Figure 2, left}). However, due to the
variability in the number of trials for different conditions - such as
rare mutations or tumor types yielding only a handful of trials, while
others result in hundreds - it is not always feasible to process all
filtered trials directly through an LLM. We therefore employed vector
similarity search to enrich trials with highest potential relevance by
calculating the cosine distance between trial information and patient
EHRs in a representation space. We selected the top k=50 trials with the
lowest cosine distance. These trials were then processed by GPT-4o,
which was instructed to discard any irrelevant trials that falsely
appeared relevant due to semantic overlap (\textbf{Figure 2, right}). As
an example, consider a patient with ``non-small cell lung cancer'' and a
clinical trial that is eligible only for ``small cell lung cancer.''
Despite the high semantic similarity (low cosine distance) between these
terms, the patient would be ineligible for the trial. This discrepancy
is accounted for by instructing GPT-4o to discard such trials, ensuring
only relevant trials are selected.

Our results demonstrate that this combined approach is highly effective,
reducing the number of candidate trials from hundreds to 20-30. Notably,
14 out of the 15 target trials (93.3\%) fall within the top 10 trial
options, and 10 out of 15 are ranked within the top 5 trials
(\textbf{Figure 2, right}). Additionally, we evaluated the potential
benefits of incorporating reranking models. Although these models have
shown promising results in optimizing text retrieval tasks and relevance
sorting for efficiency\textsuperscript{29}, we did not observe
significant improvements when applied to the full text of the trials
using Cohere's rerank-english-v3.0\textsuperscript{28}. Therefore, we
omitted reranking and considered the selected trials from the previous
step as final.

Our findings demonstrate the potential of combining No-SQL database and
vector similarity search with GPT-4o to effectively reduce the number of
trials to a few candidate options, ensuring that only the most relevant
ones are prioritized for each patient.

\subsubsection{Inclusion and Exclusion Criteria
Accuracy}\label{inclusion-and-exclusion-criteria-accuracy}

Next, we evaluated the criterion-level accuracy of GPT-4o across all 51
oncology EHRs for one target trial each, resulting in a total of 1,589
evaluable criteria, including both flat and nested ones. We show an
example of how GPT-4o internally structures these eligibility criteria
in Supplementary Table 2 and 3 and provide an example of GPT-4o's full
trial annotations including the unaltered eligibility criteria and
criterion-level AI reasoning in Supplementary Table 4. For each
criterion, the model was instructed to return one of three responses:
True if the patient was eligible based on that criterion alone, False if
the patient was not eligible, or ``unknown'' if the available data was
inadequate to make a decision. In cases involving nested criteria, where
the criterion ``header'' was not directly evaluable (e.g., ``All
patients:'' or ``(At least) one of the following:''), the model was
additionally instructed to provide a global criterion result that
reflects the logical aggregation of the nested criteria. We use the
majority answer from annotations, generated by five independent
board-licensed physicians on all 1,589 criteria as a human baseline for
comparing to the model's performance, which we highlight in
\textbf{Figure 3}. Notably, as elaborated later, we do not speak of
human annotations as ground truth. Our results demonstrate that GPT-4o
achieves an overall - preliminary - accuracy of 88.0\% (calculated as
the number of criteria where human and LLM decisions agree, divided by
the total number of criteria, \textbf{Figure 3}), with similar
performance when considering inclusion and exclusion criteria separately
(87.5\% and 88.6\% respectively). All patient cases can be found in
Supplementary Table 5.

\begin{figure}[ht!]
  \centering
  \includegraphics[width=\linewidth]{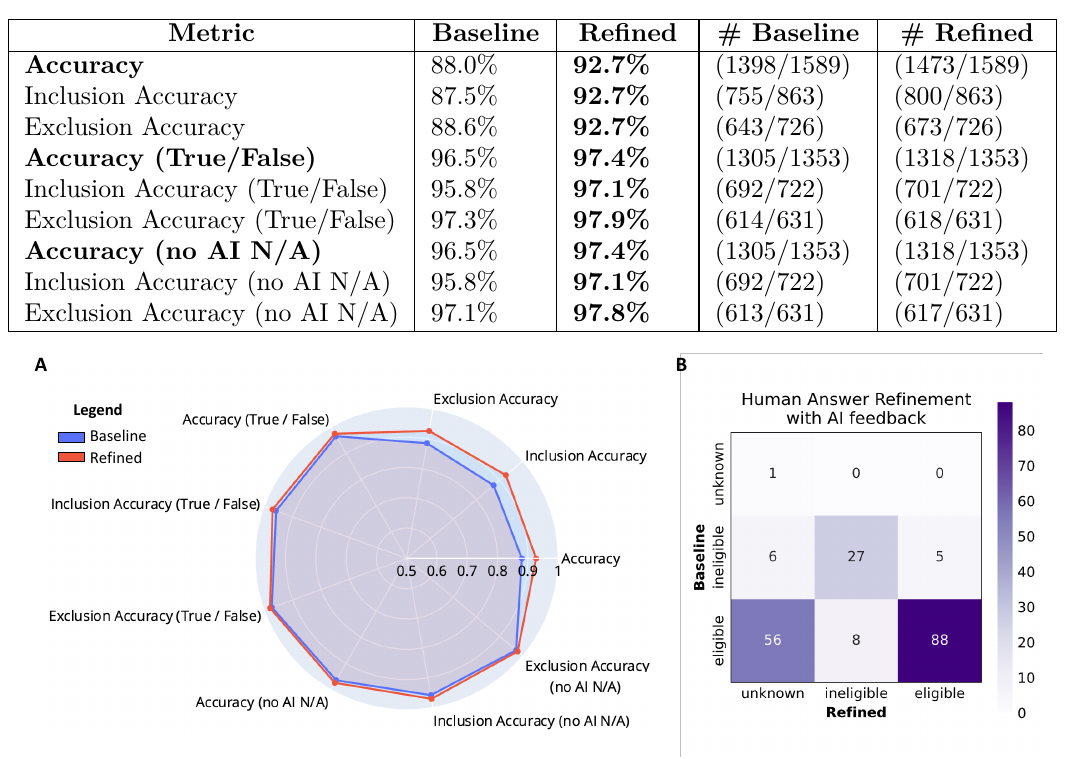}
  \caption{\textbf{Criterion-level patient-to-trial matching
performance.} We illustrate the end-to-end criteria-level decision
making performance of GPT-4o in our oncology trial matching pipeline
(A), evaluating it on 1,589 criteria across 51 cases. We first compared
the model performance to a baseline annotated by five independent
licensed physicians, resulting in an overall accuracy of 88.0\%
(Baseline). We then assessed accuracy for criteria where model and human
reference answers aligned on True or False only (``True/False'') and on
a subset of the data where N/A answers from the model were excluded
(``no AI N/A''). Subsequently, we re-evaluated the human baseline with
three of the original observers using criteria-level AI feedback on all
questions where LLM and human majority votes did not align (n=191),
showing an overall 4.7\% increment in accuracy on the newly formed
refined human ``ground truth'' (``Refined'').
We additionally highlight the shifts in corrections made by human
annotators during the refinement stage, as shown in the transition
matrix in B. The diagonal represents annotations that remained unchanged
(116/191), resulting in a remaining 75 out of 191 cases where physicians
accepted the AI response. Notably, the majority of corrections happened
when physicians initially deemed a patient eligible for a certain criterion but agreed with the LLM that sufficient information was actually missing (74.7\%, 56/75). More importantly, we observed that refinements were also made for changes
between eligible and ineligible decisions and vice versa, occurring in
10.7\% (8/75) and 6.7\% (5/75) of cases, respectively. Additionally, we
identified scenarios where human annotators corrected decisions from
eligible to ineligible (10.7\%, 8/75) and from ineligible to eligible
(6.7\%, 6/75), highlighting the potential of using AI feedback to
correct human mistakes.
}
  \label{fig:Figure3}
\end{figure}

Additionally, we find that GPT-4o achieved a 96.5\% accuracy when
focusing solely on True or False answers by either the model or human
observers (``True/False''). The same observation is made when excluding
model N/A answers only, which led to 96.5\% of the answers being
considered correct upon comparison to the human annotations (``no AI
N/A''). We consider excluding model N/A outputs as an even better
indicator of the model's performance as outputs that point out its
inability to answer the criterion due to insufficient patient or trial
information are less critical in real-world settings than incorrectly
assigning ineligibility or eligibility.

In summary, we can show that besides finding relevant trial candidates,
GPT-4o can next evaluate patient eligibility on these selected trials
with very high criterion-level accuracy.

\subsubsection{Refining human baseline with AI
feedback}\label{refining-human-baseline-with-ai-feedback}

To better understand the reasons behind differences in model versus
human annotations at the criterion level, we re-evaluated all 191 cases
where answers did not align. This process was performed by three of the
original observers, who debated these discrepancies: We found that
39.3\% (75 out of 191) of the initially conflicting answers were
accepted after considering the model's reasoning and re-assessing the
patient case and specific criteria. Following this refinement of human
baseline, our trial matching pipeline showed a 4.7\% improvement in
performance, achieving an overall accuracy of 92.7\%. Furthermore, our
model consistently performed above the 97\% accuracy threshold when
focusing on True and False answers only (``True/False'') or excluding
model responses referring to missing information (``no AI N/A'') from
the measurement (\textbf{Figure 3A}).

We next investigated the types of refinements human annotators made upon
reviewing model answers (\textbf{Figure 3B}): We found that a
substantial number of corrections to human annotations were necessary
when annotators initially considered a patient eligible or ineligible
for a certain criterion, while relevant information to make a decision
was absent in reality (74.7\%, 56/75 and 8\%, 6/75 respectively). More
importantly however, we found scenarios in which human annotators
corrected eligible to ineligible decisions (10.7\%, 8/75) and ineligible
to eligible (6.7\%, 6/75), indicating instances where human annotators
made substantial mistakes that could be corrected using AI feedback.

In summary, we herein show that GPT-4o can match, and likely even exceed
the performance of qualified physicians in evaluating patient trial
eligibility.

\section{\texorpdfstring{\textbf{Discussion}}{Discussion}}\label{discussion}

In this work, we describe and validate a fully end-to-end approach for
leveraging LLMs for clinical trial matching using oncology cases as an
example. Overall, we demonstrate how GPT-4o can first effectively screen
potential trial candidates from a collection of over one hundred
thousand clinical trials registered on clinicaltrials.gov and secondly
match selected candidates on a criterion-by-criterion basis to patient
records.

This has several real-world advantages: From a clinical perspective,
physicians must filter out over 99.9\% of irrelevant trials due to
differing tumor types, disease stages, or distant locations.
Additionally, they must consider what type of trials they are
specifically looking for: Should they target a particular molecular
alteration? Are they seeking trials for treatment-naive patients, or for
those refractory to other therapies? Consequently, physicians are often
forced to rely on ad hoc searches rather than structured methods to find
suitable clinical trials. Given the inherent capabilities of
state-of-the-art LLMs, we show that the process of filtering relevant
trials by keyword-based search can be automated using GPT-4o, which can
itself write queries for a No-SQL database, guided without or with human
supervision, such as ``Please find a Phase 1 trial for the patient in
Germany'' or ``Could you please find a clinical trial for the patient's
KRAS mutation (or all of the patient's mutations)?''

Our approach leverages the robust capabilities of GPT-4o in generating
valid computer code, allowing programmatic access to trial databases
with only optional human guidance. For instance, GPT-4o can request
trials in specific locations or target particular mutations, or
combinations of both if given as instruction from medical professionals.
This makes our system highly scalable and flexible, extending its
applicability beyond pre-selecting trials from a single center, as
demonstrated by Gupta et al.\textsuperscript{19}

Moreover, we are convinced of the inherent reasoning capabilities of
LLMs\textsuperscript{21}, particularly looking toward future
advancements, allowing them to handle complex logic internally. In
contrast, the approach by Wong et al.\textsuperscript{17} explicitly
enforces LLMs to rewrite eligibility logic into structured Disjunctive
Normal Form (DNF), imposing constraints on the model by limiting the
combination of categories such as disease state, histology, and
biomarkers through logical conditions (and, or, all, any, not etc.).
This method also alters the original trial criteria, complicating human
evaluation. Our approach ensures that the model's output can be mapped
back to the original criteria on a one-by-one basis, with each criterion
accompanied by a detailed chain of reasoning explaining the model's
decision. This allows medical doctors to fact-check each decision of the
model, ensuring explainability and trust. By understanding why the model
reaches a particular conclusion and identifying potential errors we can
better understand capabilities and limitations of current LLMs in
managing EHR data.

Additionally, we demonstrate that our system's performance can match and
under certain conditions even surpass that of human experts in criterion
decision tasks. Although not directly comparable due to different trials
and patient data (Gupta et al. utilize real-world de-identified EHR
cases) and potential variations in how trial criteria are processed, our
overall pipeline achieves an accuracy of 92.7\% and 97.4\% when
excluding N/A samples. This exceeds the results others have previously
achieved with GPT-4, reporting accuracies of 68\% and 72\%,
respectively\textsuperscript{19}.

Moreover, our program-rather-than-prompt strategy ensures that responses
consistently adhere to the required format, reducing the burden of
finding optimal, often specific and narrow prompts. We therefore can
guarantee that regardless of the length or complexity of the criteria,
we receive validly annotated and unaltered criteria back from the LLM,
which is not the case if criteria are handled as plain, free text.

Also, this approach allows the broad transferability of our system to
other medical domains, with minimal need for adjustments, such as only
the need for addressing domain-specific edge cases.

Nevertheless, our study has several limitations: In real-world
scenarios, under current regulatory restrictions, GPT-4o is not a
suitable candidate due to its cloud-based nature, which necessitates
transferring sensitive patient data to proprietary servers. Thus, we
consider GPT-4o as a best-in-class model suitable for proof-of-concept
purposes. We anticipate that local model solutions will catch up in
performance in the near future, making them more suitable for clinical
application. Additionally, real-world patient data will be required to
fully validate applicability of our system, incorporating longer and
even more diverse patient documents. For instance, laboratory values may
be nested in spreadsheets, and imaging data might be separate, with all
relevant patient information distributed across various documents.
Moreover, we aim to evaluate the model's ability to accurately rank and
prioritize the most relevant trials, enabling doctors to quickly
identify the best options for their patients. Although our system
currently provides scores based on the number and ratio of fulfilled
eligibility criteria, we have not yet established a sophisticated
measure for quantitative evaluation. We plan to develop and refine this
using real-world data in the near future.

Despite these challenges, our work demonstrates that an LLM can
autonomously narrow down relevant trials from thousands to a manageable
handful and accurately match these trials criterion by criterion. To our
knowledge, our study is the closest in mirroring the real-world scenario
of how medical doctors interact with clinical trial databases like
clinicaltrials.gov. This evidence suggests significant potential for our
approach, particularly as we show, for the first time, that AI feedback
can enhance the performance of medical specialists in identifying
suitable clinical trials for their patients.

\section{\texorpdfstring{\textbf{Acknowledgements}}{Acknowledgements}}\label{acknowledgements}

We thank OpenAI for supporting our work through a researcher
access grant.

\section{\texorpdfstring{\textbf{References}}{References}}\label{references}

1. Bouzalmate-Hajjaj, A., Massó Guijarro, P., Khan, K. S.,
Bueno-Cavanillas, A. \& Cano-Ibáñez, N. Benefits of Participation in
Clinical Trials: An Umbrella Review. \emph{Int. J. Environ. Res. Public
Health} \textbf{19}, (2022).

2. Unger, J. M., Cook, E., Tai, E. \& Bleyer, A. The Role of Clinical
Trial Participation in Cancer Research: Barriers, Evidence, and
Strategies. \emph{Am Soc Clin Oncol Educ Book} \textbf{35}, 185--198
(2016).

3. Penberthy, L. T., Dahman, B. A., Petkov, V. I. \& DeShazo, J. P.
Effort required in eligibility screening for clinical trials. \emph{J.
Oncol. Pract.} \textbf{8}, 365--370 (2012).

4. Unger, J. M., Vaidya, R., Hershman, D. L., Minasian, L. M. \& Fleury,
M. E. Systematic Review and Meta-Analysis of the Magnitude of
Structural, Clinical, and Physician and Patient Barriers to Cancer
Clinical Trial Participation. \emph{J. Natl. Cancer Inst.} \textbf{111},
245--255 (2019).

5. Oxentenko, A. S., West, C. P., Popkave, C., Weinberger, S. E. \&
Kolars, J. C. Time spent on clinical documentation: a survey of internal
medicine residents and program directors. \emph{Arch. Intern. Med.}
\textbf{170}, 377--380 (2010).

6. Rule, A., Bedrick, S., Chiang, M. F. \& Hribar, M. R. Length and
Redundancy of Outpatient Progress Notes Across a Decade at an Academic
Medical Center. \emph{JAMA Netw Open} \textbf{4}, e2115334 (2021).

7. Moy, A. J. \emph{et al.} Measurement of clinical documentation burden
among physicians and nurses using electronic health records: a scoping
review. \emph{J. Am. Med. Inform. Assoc.} \textbf{28}, 998--1008 (2021).

8. Kong, H.-J. Managing Unstructured Big Data in Healthcare System.
\emph{Healthc. Inform. Res.} \textbf{25}, 1--2 (2019).

9. Bradley, J., Kelly, K. \& Stinchcombe, T. E. The Ever-Increasing
Number of Trial Eligibility Criteria: Time to Bend the Curve.
\emph{Journal of thoracic oncology: official publication of the
International Association for the Study of Lung Cancer} vol. 12
1459--1460 (2017).

10. Zhang, X., Xiao, C., Glass, L. M. \& Sun, J. DeepEnroll:
Patient-Trial Matching with Deep Embedding and Entailment Prediction.
\emph{arXiv {[}cs.AI{]}} (2020).

11. Gao, J., Xiao, C., Glass, L. M. \& Sun, J. COMPOSE: Cross-Modal
Pseudo-Siamese Network for Patient Trial Matching. in \emph{Proceedings
of the 26th ACM SIGKDD International Conference on Knowledge Discovery
\& Data Mining} 803--812 (Association for Computing Machinery, New York,
NY, USA, 2020).

12. Yuan, C. \emph{et al.} Criteria2Query: a natural language interface
to clinical databases for cohort definition. \emph{J. Am. Med. Inform.
Assoc.} \textbf{26}, 294--305 (2019).

13. OpenAI \emph{et al.} GPT-4 Technical Report. \emph{arXiv
{[}cs.CL{]}} (2023).

14. Wiest, I. C. \emph{et al.} From Text to Tables: A Local Privacy
Preserving Large Language Model for Structured Information Retrieval
from Medical Documents. \emph{medRxiv} 2023.12.07.23299648 (2023)
doi:10.1101/2023.12.07.23299648.

15. Jin, Q. \emph{et al.} Matching Patients to Clinical Trials with
Large Language Models. \emph{ArXiv} (2024).

16. den Hamer, D. M., Schoor, P., Polak, T. B. \& Kapitan, D. Improving
Patient Pre-screening for Clinical Trials: Assisting Physicians with
Large Language Models. \emph{arXiv {[}cs.LG{]}} (2023).

17. Wong, C. \emph{et al.} Scaling Clinical Trial Matching Using Large
Language Models: A Case Study in Oncology. in \emph{Proceedings of the
8th Machine Learning for Healthcare Conference} (eds. Deshpande, K. et
al.) vol. 219 846--862 (PMLR, 11-\/-12 Aug 2023).

18. Nievas, M., Basu, A., Wang, Y. \& Singh, H. Distilling large
language models for matching patients to clinical trials. \emph{J. Am.
Med. Inform. Assoc.} (2024) doi:10.1093/jamia/ocae073.

19. Gupta, S. K. \emph{et al.} PRISM: Patient Records Interpretation for
Semantic Clinical Trial Matching using Large Language Models.
\emph{arXiv {[}cs.CL{]}} (2024).

20. Wornow, M. \emph{et al.} Zero-Shot Clinical Trial Patient Matching
with LLMs. \emph{arXiv {[}cs.CL{]}} (2024).

21. Truhn, D., Reis-Filho, J. S. \& Kather, J. N. Large language models
should be used as scientific reasoning engines, not knowledge databases.
\emph{Nat. Med.} \textbf{29}, 2983--2984 (2023).

22. Singhvi, A. \emph{et al.} DSPy Assertions: Computational Constraints
for Self-Refining Language Model Pipelines. \emph{arXiv {[}cs.CL{]}}
(2023).

23. Cattell, R. Scalable SQL and NoSQL data stores. \emph{SIGMOD Rec.}
\textbf{39}, 12--27 (2011).

24. Chroma. https://www.trychroma.com/.

25. Benary, M. \emph{et al.} Leveraging Large Language Models for
Decision Support in Personalized Oncology. \emph{JAMA Netw Open}
\textbf{6}, e2343689 (2023).

26. Lammert, J. \emph{et al.} Expert-guided large language models for
clinical decision support in precision oncology. (2024)
doi:10.2139/ssrn.4855985.

27. Welcome to pydantic - pydantic. https://docs.pydantic.dev/latest/.

28. Introducing Rerank 3: A new foundation model for efficient
enterprise search \& retrieval. \emph{Cohere}
https://cohere.com/blog/rerank-3.

29. Sasazawa, Y., Yokote, K., Imaichi, O. \& Sogawa, Y. Text Retrieval
with Multi-Stage Re-Ranking Models. \emph{arXiv {[}cs.IR{]}} (2023).

\subsection{\texorpdfstring{\textbf{Data availability
statement}}{Data availability statement}}\label{data-availability-statement}

All clinical trial information used can be accessed and downloaded
manually via https://clinicaltrials.gov/ as detailed in ``\emph{Methods
- Clinical Trial Composition}''. Note that available trials and
information on existing trials will change over time. We release all 51
synthetic EHR notes, that are based on case vignettes published by
Benary et al.\textsuperscript{25} in Supplementary Data Table 5.

\subsection{\texorpdfstring{\textbf{Code availability
statement}}{Code availability statement}}\label{code-availability-statement}

All methods necessary to reproduce our results are extensively
documented. While we plan to enhance our pipeline further, we are
committed to offering researchers access to our findings and
methodologies in the near future. We release codes from the current
implementation for research purposes upon publication in a scientific
journal here: https://github.com/Dyke-F/llm-trials.

\subsection{\texorpdfstring{\textbf{Ethics
statement}}{Ethics statement}}\label{ethics-statement}

This study does not include confidential information. All research
procedures were conducted exclusively on publicly accessible, anonymized
patient data and in accordance with the Declaration of Helsinki,
maintaining all relevant ethical standards. The overall analysis was
approved by the Ethics commission of the Medical Faculty of the
Technical University Dresden (BO-EK-444102022).

\subsection{\texorpdfstring{\textbf{Statement on use of Artificial
Intelligence
Tools}}{Statement on Use of Artificial Intelligence Tools}}\label{statement-on-use-of-artificial-intelligence-tools}

In accordance with the COPE (Committee on Publication Ethics) position
statement of 13 February 2023
(https://publicationethics.org/cope-position-statements/ai-author), the
authors hereby disclose the use of the following artificial intelligence
models during the writing of this article. GPT-4 (OpenAI) for checking
spelling and grammar.

\subsection{\texorpdfstring{\textbf{Author
Contributions}}{Author Contributions}}\label{author-contributions}

DF designed and performed the experiments, evaluated and interpreted the
results and wrote the initial draft of the manuscript. DF, LH and PN
developed the case vignettes. ICW, JC, ML, LH and DF analyzed the
results; ICW, LH and DF performed the eligibility re-evaluation. JZ
designed the web interface for eligibility evaluation. CH provided
expertise for the discussion of the implications of the findings. All
authors contributed to writing the manuscript. MA, DJ, DT and JNK
supervised the study.

\subsection{\texorpdfstring{\textbf{Funding}}{Funding}}\label{funding}

JNK is supported by the German Cancer Aid (DECADE, 70115166), the German
Federal Ministry of Education and Research (PEARL, 01KD2104C; CAMINO,
01EO2101; SWAG, 01KD2215A; TRANSFORM LIVER, 031L0312A; TANGERINE,
01KT2302 through ERA-NET Transcan; Come2Data, 16DKZ2044A; DEEP-HCC,
031L0315A), the German Academic Exchange Service (SECAI, 57616814), the
German Federal Joint Committee (TransplantKI, 01VSF21048) the European
Union's Horizon Europe and innovation programme (ODELIA, 101057091;
GENIAL, 101096312), the European Research Council (ERC; NADIR,
101114631), the National Institutes of Health (EPICO, R01 CA263318) and
the National Institute for Health and Care Research (NIHR, NIHR203331)
Leeds Biomedical Research Centre. The views expressed are those of the
author(s) and not necessarily those of the NHS, the NIHR or the
Department of Health and Social Care. This work was funded by the
European Union. Views and opinions expressed are however those of the
author(s) only and do not necessarily reflect those of the European
Union. Neither the European Union nor the granting authority can be held
responsible for them. JC is supported by the
Mildred-Scheel-Postdoktorandenprogramm of the German Cancer Aid (grant
\#70115730). DT is funded by the German Federal Ministry of Education
and Research (TRANSFORM LIVER, 031L0312A), the European Union's Horizon
Europe and innovation programme (ODELIA, 101057091), and the German
Federal Ministry of Health (SWAG, 01KD2215B). GW is supported by Lothian
NHS. JL is supported by the TUM School of Medicine and Health Clinician
Scientist Program (project no. H-08). CH contributed to this work in her
personal interest outside of her employment at AstraZeneca GmbH. The
views expressed are those of the author(s) and not necessarily those of
AstraZeneca, the NHS, the NIHR or the Department of Health and Social
Care. No other funding is disclosed by any of the authors.

\subsection{\texorpdfstring{\textbf{Competing
Interests}}{Competing Interests}}\label{competing-interests}

JNK declares consulting services for Owkin, France; DoMore Diagnostics,
Norway; Panakeia, UK, and Scailyte, Basel, Switzerland; furthermore JNK
holds shares in Kather Consulting, Dresden, Germany; and StratifAI GmbH,
Dresden, Germany, and has received honoraria for lectures and advisory
board participation by AstraZeneca, Bayer, Eisai, MSD, BMS, Roche,
Pfizer and Fresenius. DT received honoraria for lectures by Bayer and
holds shares in StratifAI GmbH, Germany. ICW received honoraria from
AstraZeneca. The authors have no additional financial or non-financial
conflicts of interest to disclose.
\clearpage

\setcounter{figure}{0}
\captionsetup[figure]{labelformat=empty}

\captionsetup{font=footnotesize} 
\captionsetup[figure]{labelfont=bf}

\section{\texorpdfstring{\textbf{Supplementary
Data}}{Supplementary Data}}\label{supplementary-data}

\textbf{Supplementary Table 1.} Model Instructions for Trial Search.

\textbf{Supplementary Table 2.} Unstructured eligibility criteria for
NCT02227251: \emph{Selinexor (KPT-330) in Patients With
Relapsed/\hspace{0pt}Refractory Diffuse Large B-Cell Lymphoma (DLBCL)}.

\textbf{Supplementary Table 3.} LLM-structured eligibility criteria for
NCT02227251: \emph{Selinexor (KPT-330) in Patients With
Relapsed/\hspace{0pt}Refractory Diffuse Large B-Cell Lymphoma (DLBCL)}.

\textbf{Supplementary Table 4.} Fully evaluated clinical trial
eligibility criteria for Patient 1.1.1.

\textbf{Supplementary Figure 1.} Web-based user interface for assessing
trial eligibility criteria on a per patient basis.

\textbf{Supplementary Table 5.} Clinical Case EHRs.

\clearpage

\clearpage

\begin{figure}[ht!]
  \centering
  \includegraphics[width=0.9\linewidth]{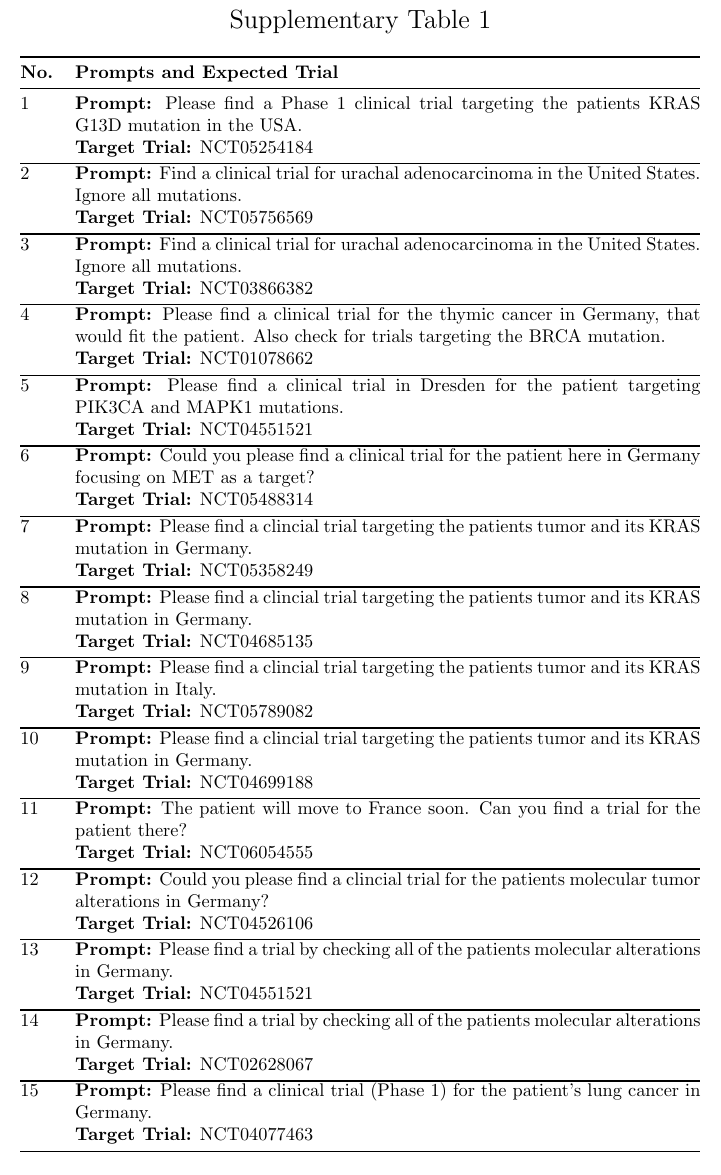}
  \caption{\textbf{Supplementary Table 1: Model instructions for Trial Search.} We show the instructions (prompts) as provided to the model throughout the initial search process and the expected trial (which is not shown to the model). Corresponding results can be found in Figure 2.
}
  \label{fig:SupplementaryTable1}
\end{figure}

\clearpage

\begin{figure}[ht!]
  \centering
  \includegraphics[width=0.95\linewidth]{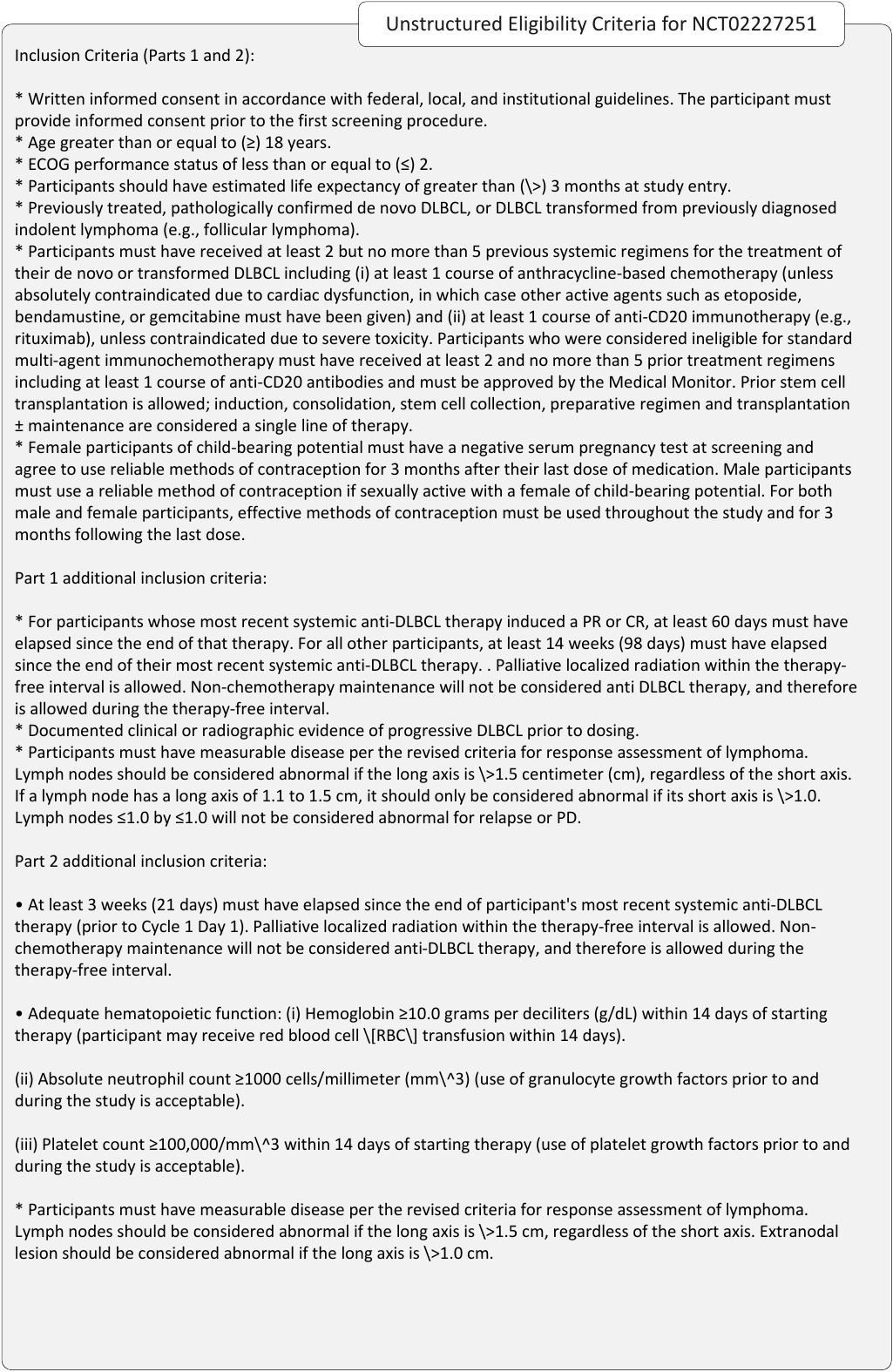}
  \label{fig:SupplementaryTable2}
\end{figure}

\clearpage

\begin{figure}[ht!]
  \centering
  \includegraphics[width=0.95\linewidth]{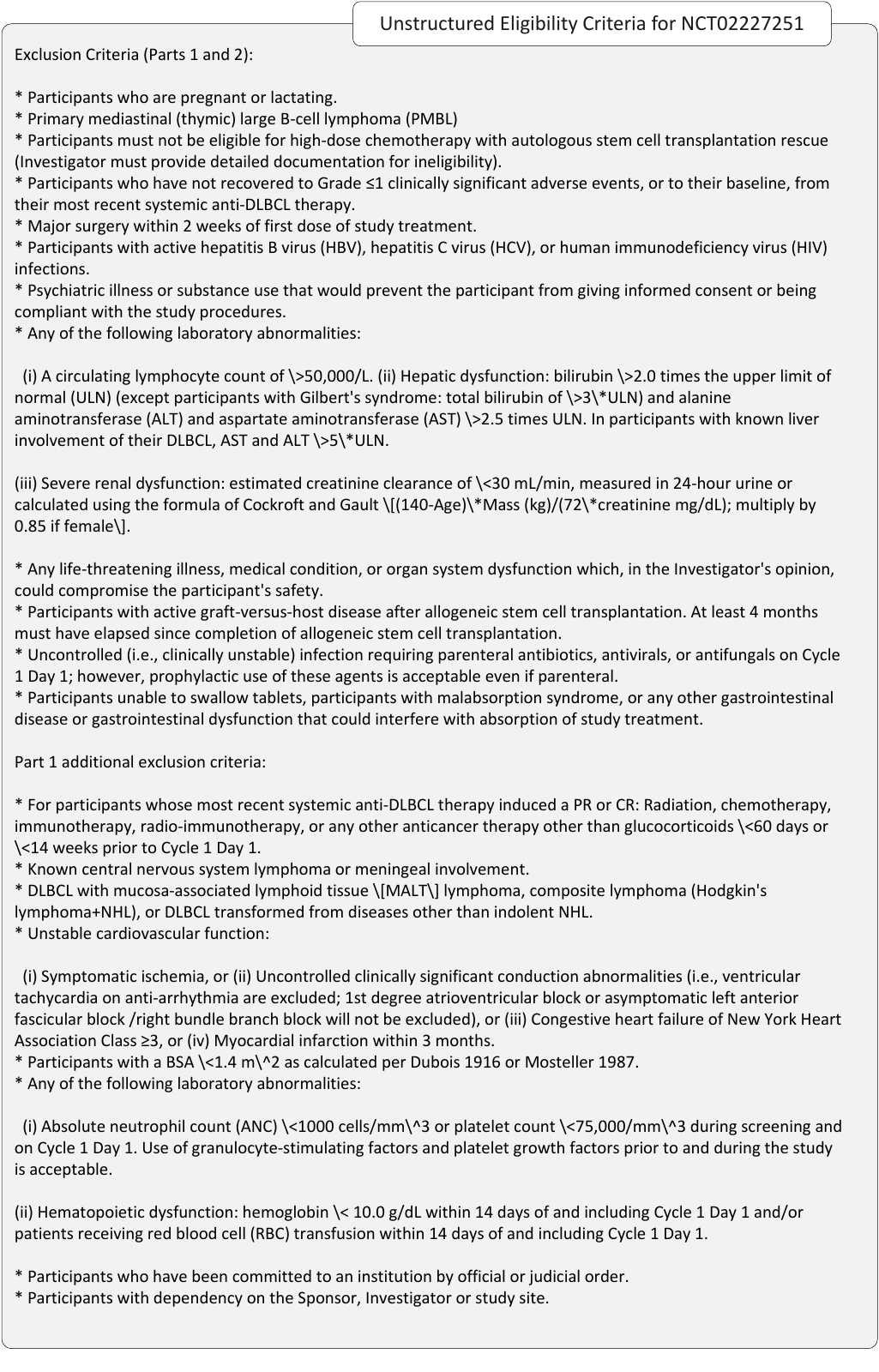}
  \label{fig:SupplementaryTable3}
\end{figure}

\clearpage

\begin{figure}[ht!]
  \centering
  \includegraphics[width=0.95\linewidth]{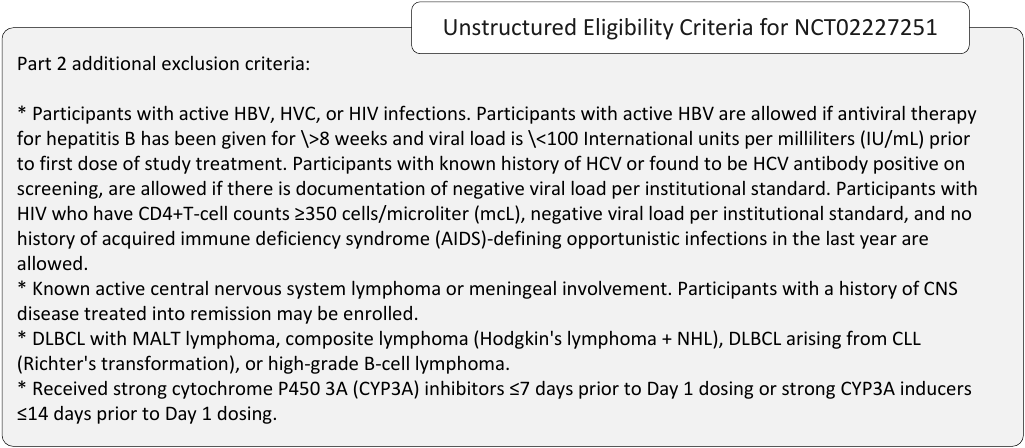}
  \caption{\textbf{Supplementary Table 2. Unstructured eligibility criteria:} Selinexor (KPT-330) in Patients With Relapsed/Refractory Diffuse Large B-Cell Lymphoma (DLBCL). This trial's eligibility criteria highlight the challenges faced by LLMs. Criteria are presented as free text and include multiple, nested inclusion and exclusion conditions and hiearchies (e.g. any, all). Formatting inconsistencies are common, with variations in line breaks, paragraphs, and enumeration symbols (e.g., bullets and roman numerals). Also, criteria headers can even be mixed with conditions (e.g. adequate hematopoietic function: (i) ...).}
  \label{fig:SupplementaryTable4}
\end{figure}

\begin{figure}[ht!]
  \centering
  \includegraphics[width=0.95\linewidth]{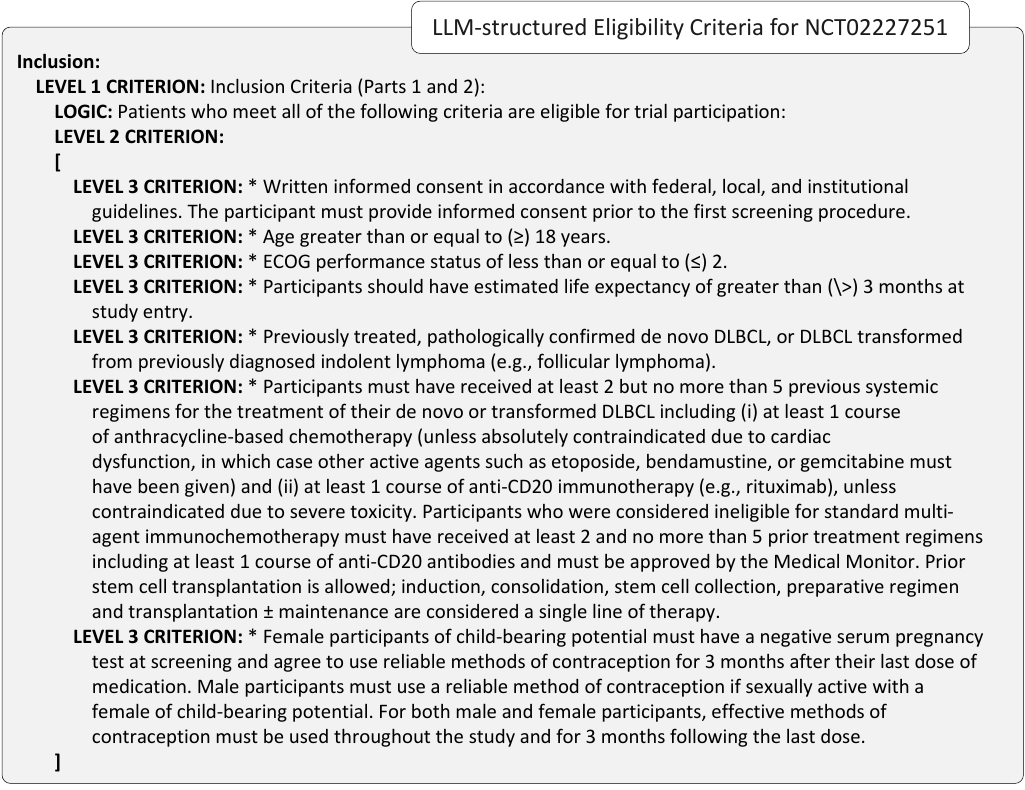}
  \label{fig:SupplementaryTable5}
\end{figure}

\clearpage

\begin{figure}[ht!]
  \centering
  \includegraphics[width=0.95\linewidth]{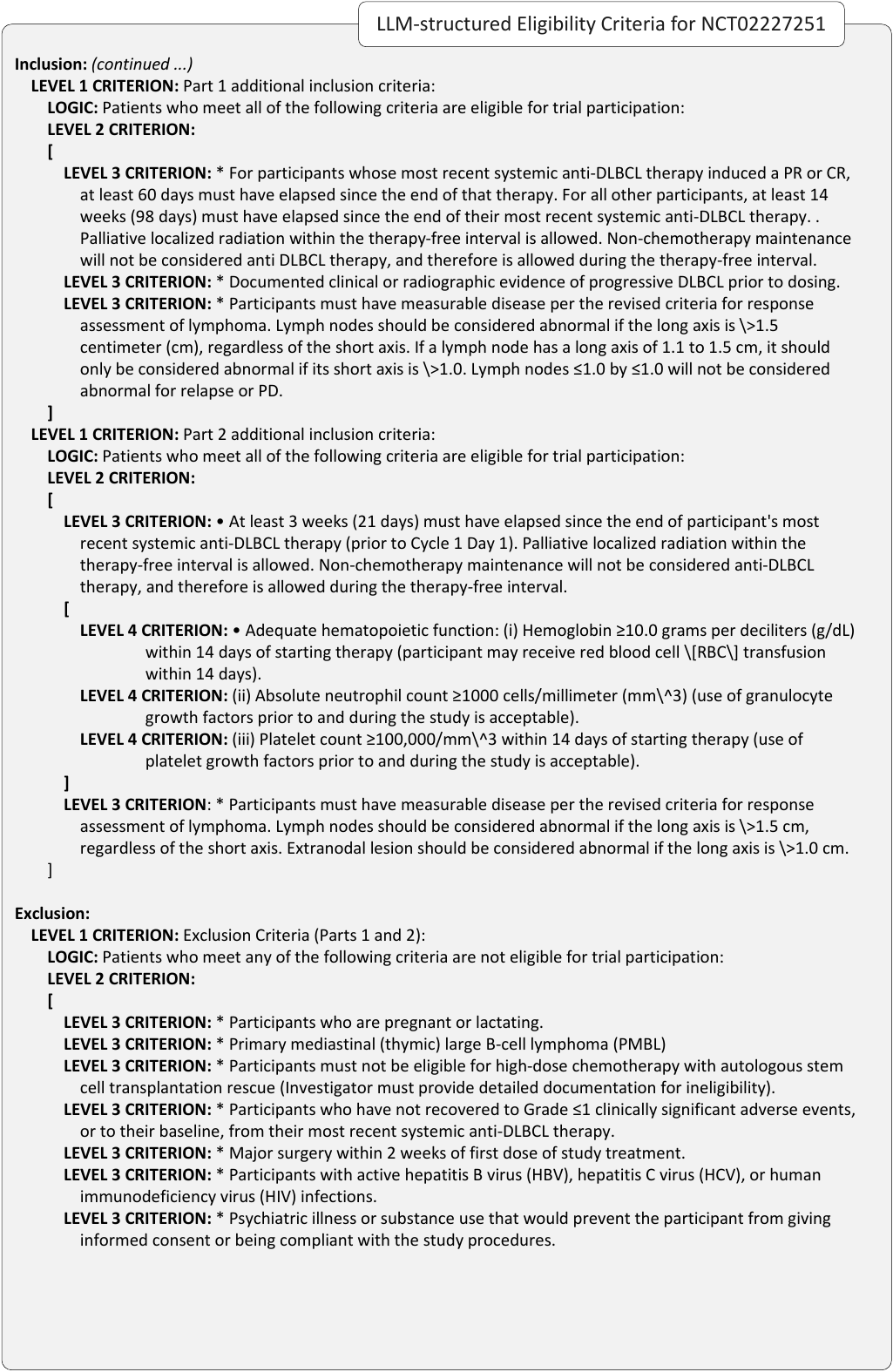}
  \label{fig:SupplementaryTable6}
\end{figure}

\clearpage

\begin{figure}[ht!]
  \centering
  \includegraphics[width=0.95\linewidth]{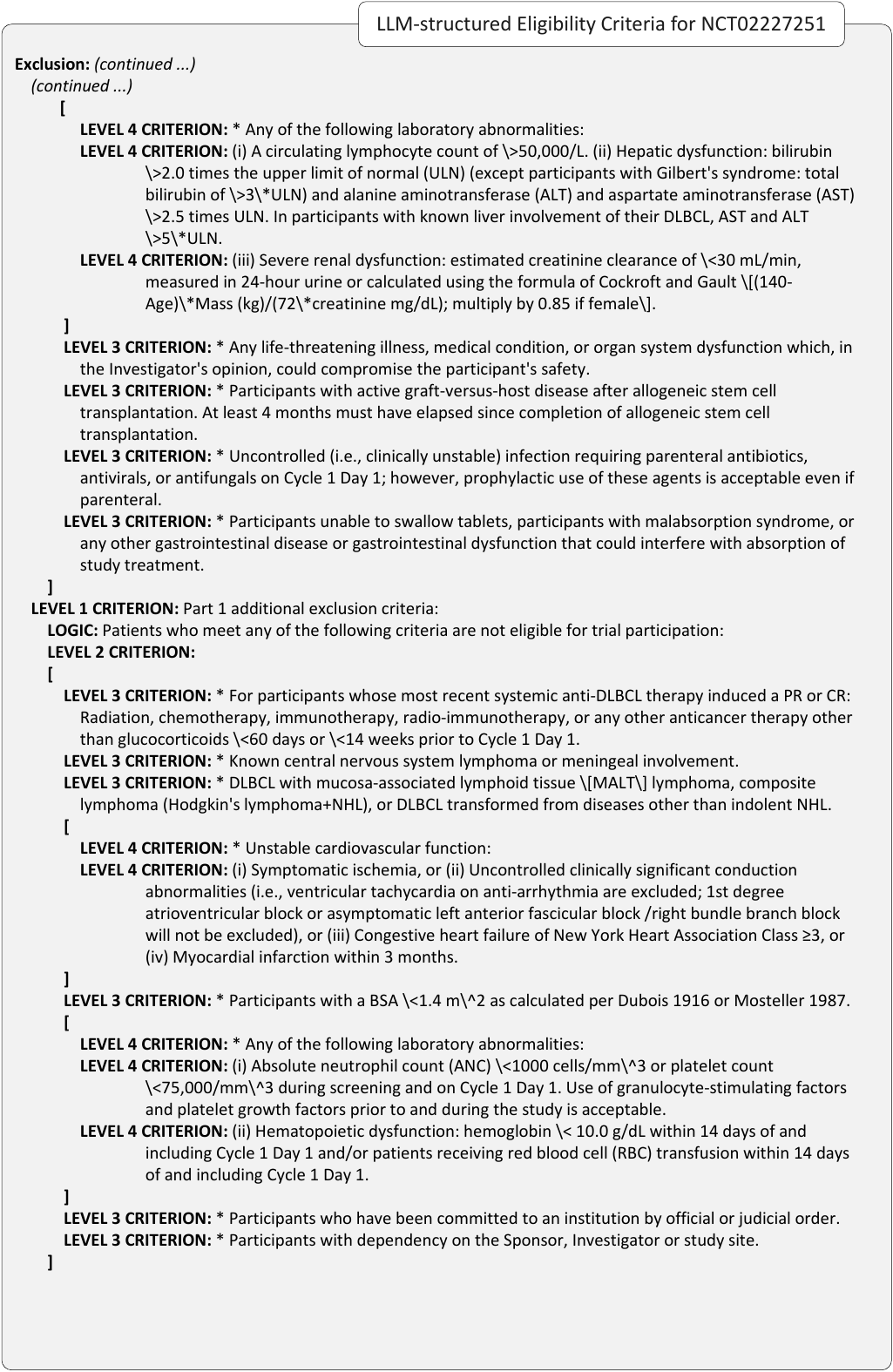}
  \label{fig:SupplementaryTable7}
\end{figure}

\clearpage

\begin{figure}[ht!]
  \centering
  \includegraphics[width=0.95\linewidth]{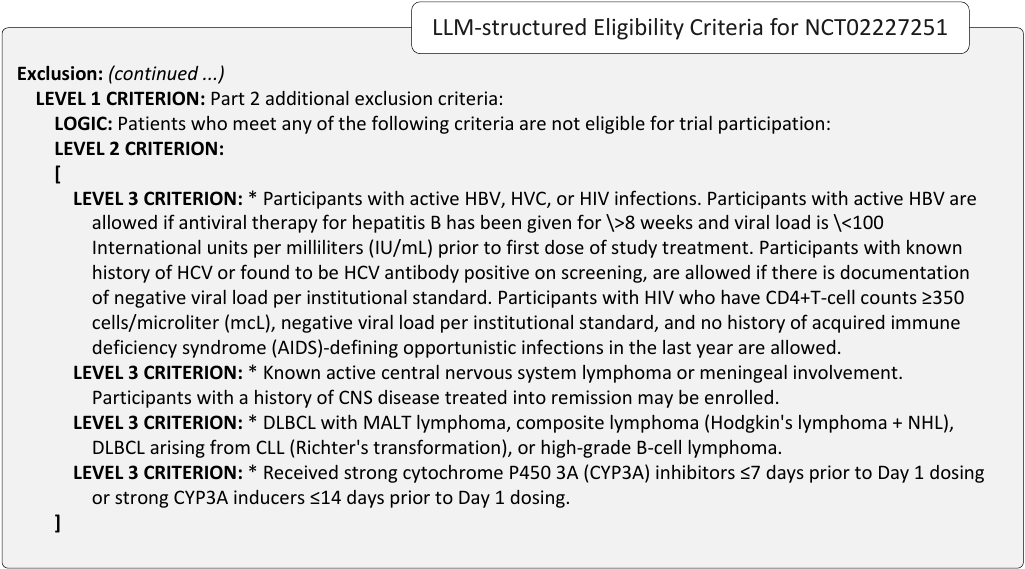}
  \label{fig:SupplementaryTable8}
    \caption{\textbf{Supplementary Table 3. LLM-structured eligibility criteria:} Selinexor (KPT-330) in Patients With Relapsed/Refractory Diffuse Large B-Cell Lymphoma (DLBCL). This table provides an example of structured eligibility criteria using GPT-4o. From high to low levels, the eligibility criteria are organized hierarchically, starting with separating inclusion and exclusion fields. Each field is then further divided into logical conditions and individual criteria. The criteria themselves can be either single conditions (Level 3) or nested criteria (Level 4). To achieve this structure, a strategy analogous to Disjunctive Normal Form (DNF) is employed, resulting in a hierarchical tree. Our results indicate that reasonable structuring can be achieved with two levels of nesting, although at the most granular level, single statements (conditions) could serve as the leaves, while all logical connectors (AND, OR, ALL, ANY) are placed as the edges. We will explore further optimization and refinement strategies of this approach in future iterations. Please note that the annotations (LEVEL X CRITERION and LOGIC) are only for readability purposes here, while the criteria hierarchy in reality is nested within the program code.}
  
\end{figure}

\clearpage

\begin{figure}[ht!]
  \centering
  \includegraphics[width=\linewidth]{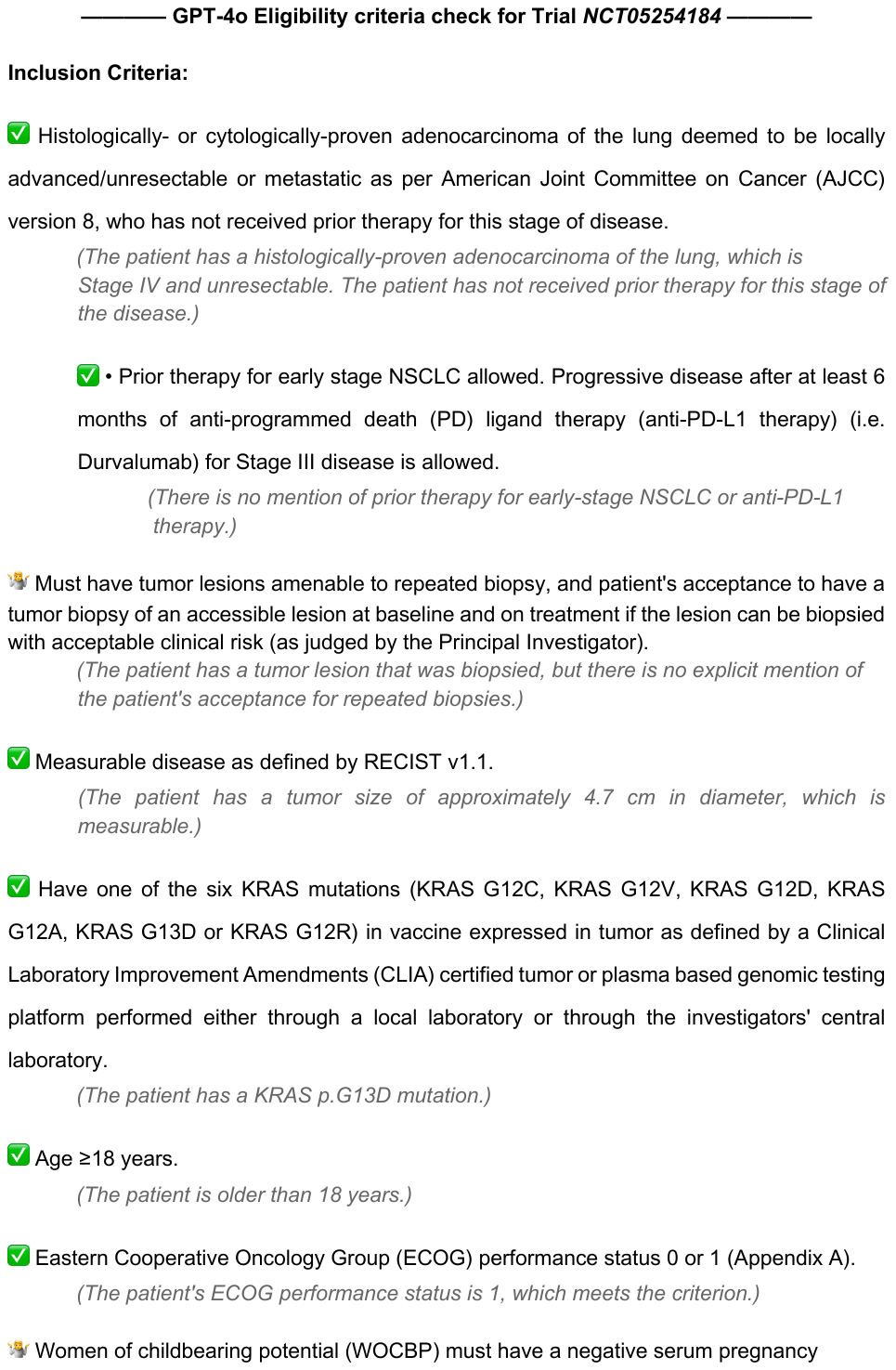}
  \label{fig:RatedTrial1}
\end{figure}

\clearpage

\begin{figure}[ht!]
  \centering
  \includegraphics[width=\linewidth]{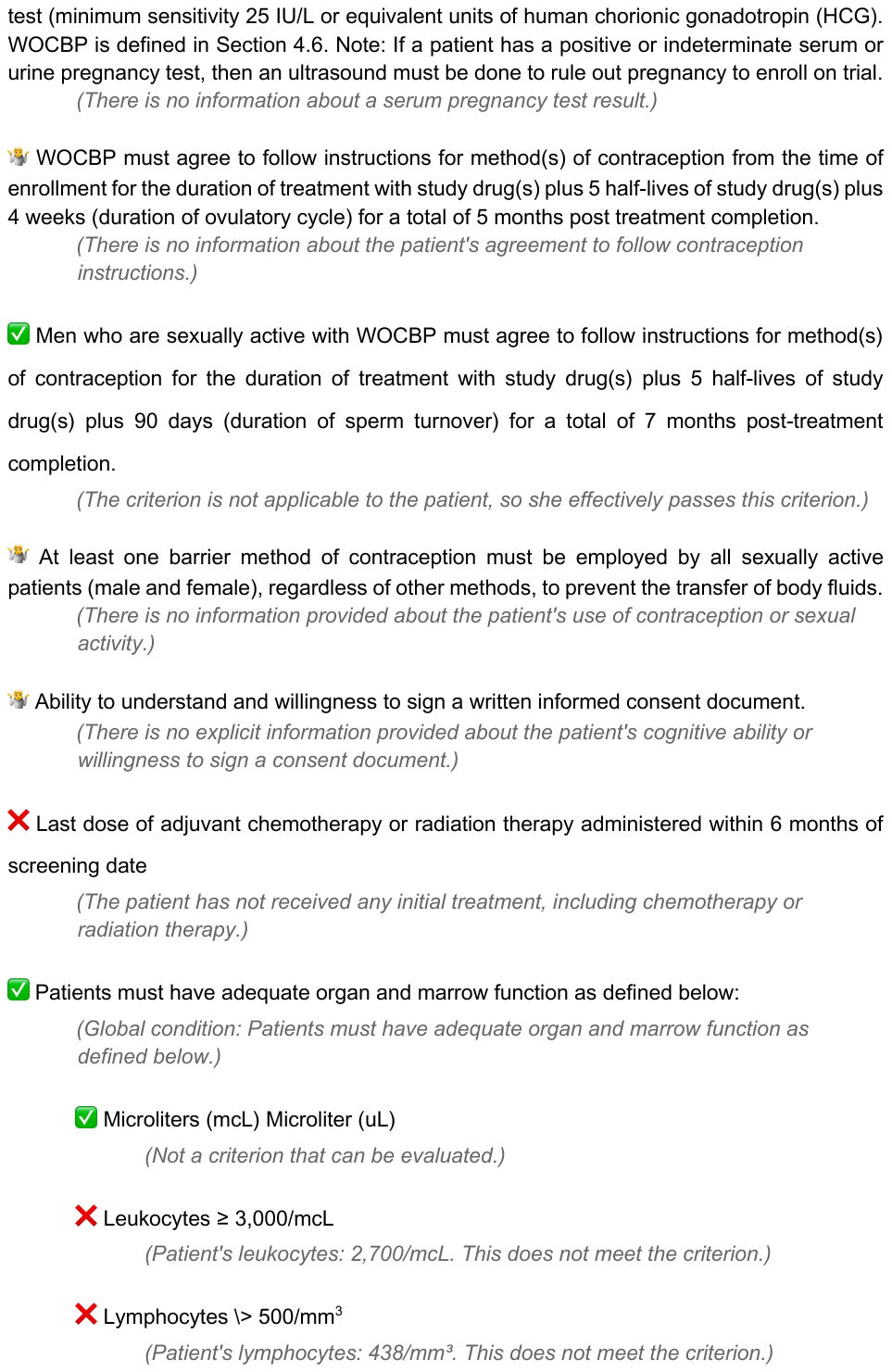}
  \label{fig:RatedTrial2}
\end{figure}

\clearpage

\begin{figure}[ht!]
  \centering
  \includegraphics[width=\linewidth]{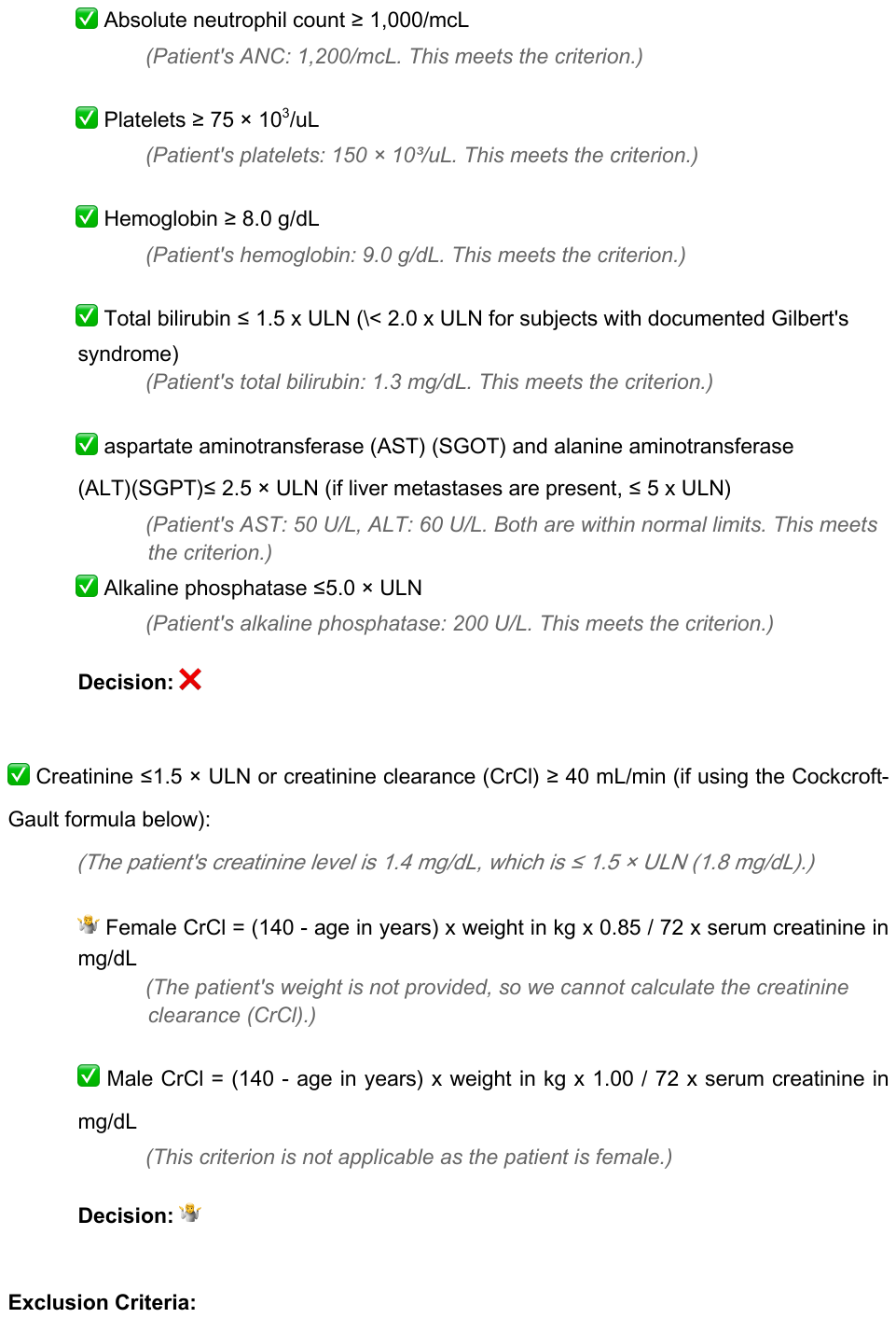}
  \label{fig:RatedTrial3}
\end{figure}

\clearpage

\begin{figure}[ht!]
  \centering
  \includegraphics[width=\linewidth]{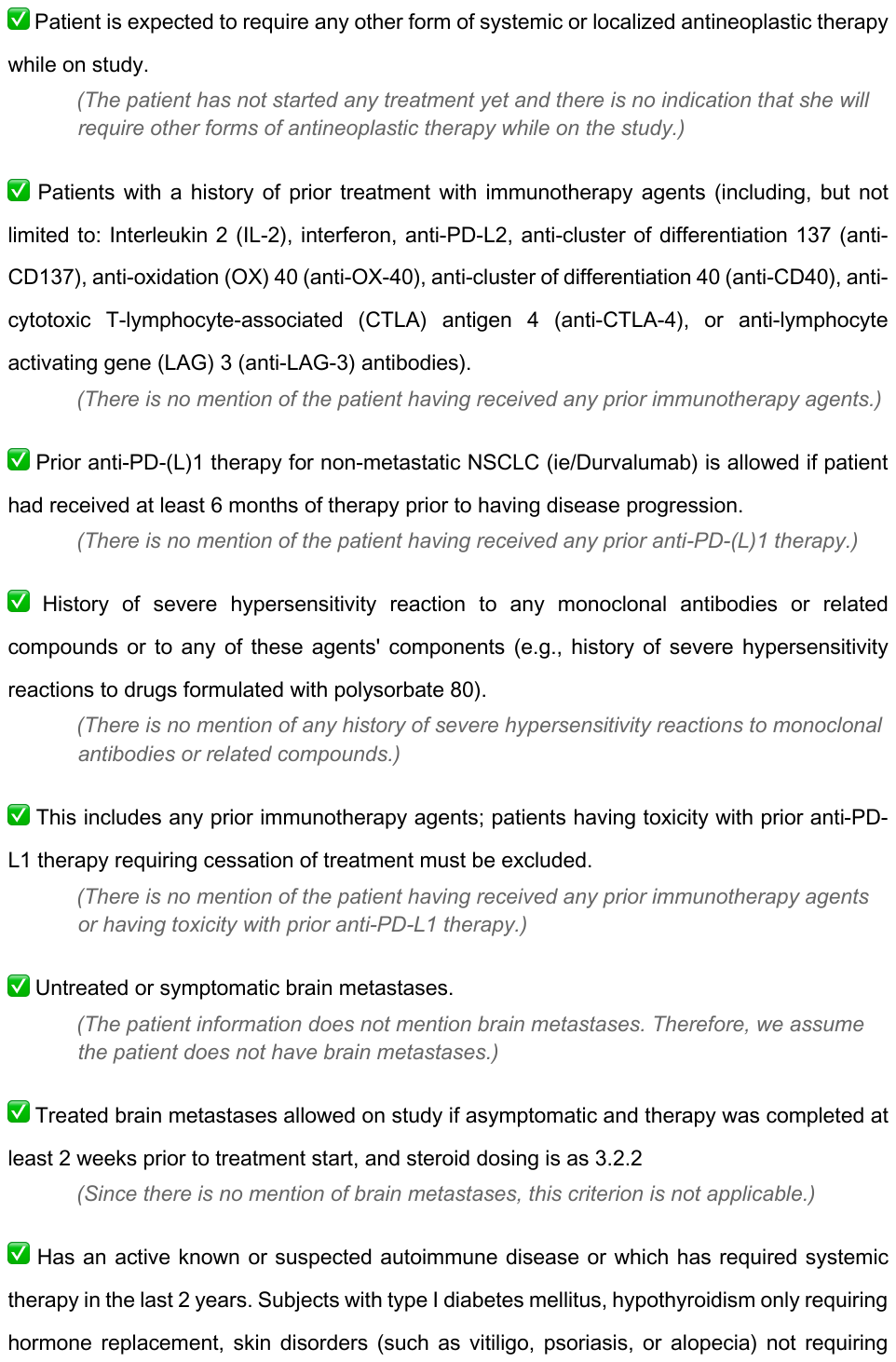}
  \label{fig:RatedTrial4}
\end{figure}

\clearpage

\begin{figure}[ht!]
  \centering
  \includegraphics[width=\linewidth]{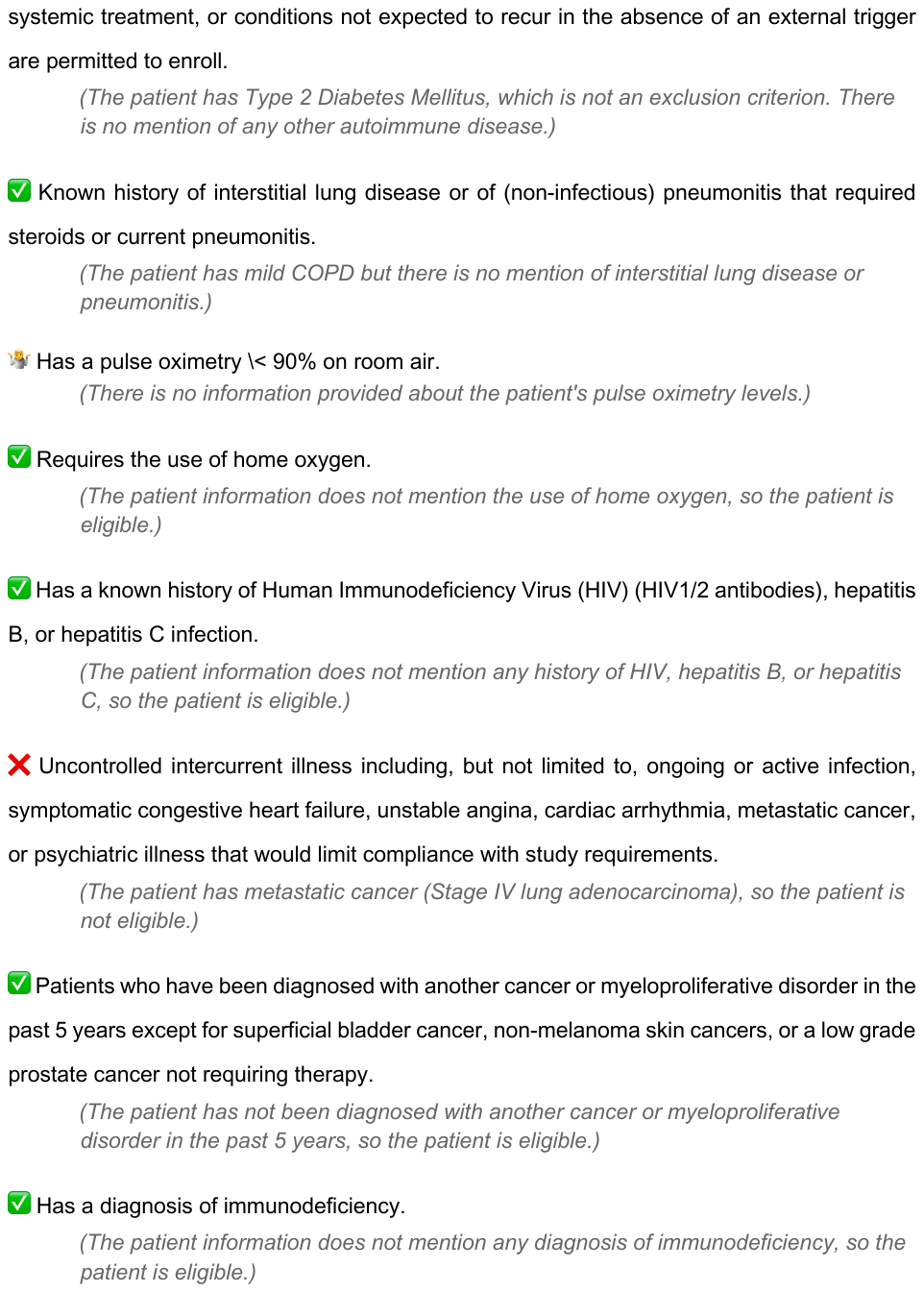}
  \label{fig:RatedTrial5}
\end{figure}

\clearpage

\begin{figure}[ht!]
  \centering
  \includegraphics[width=\linewidth]{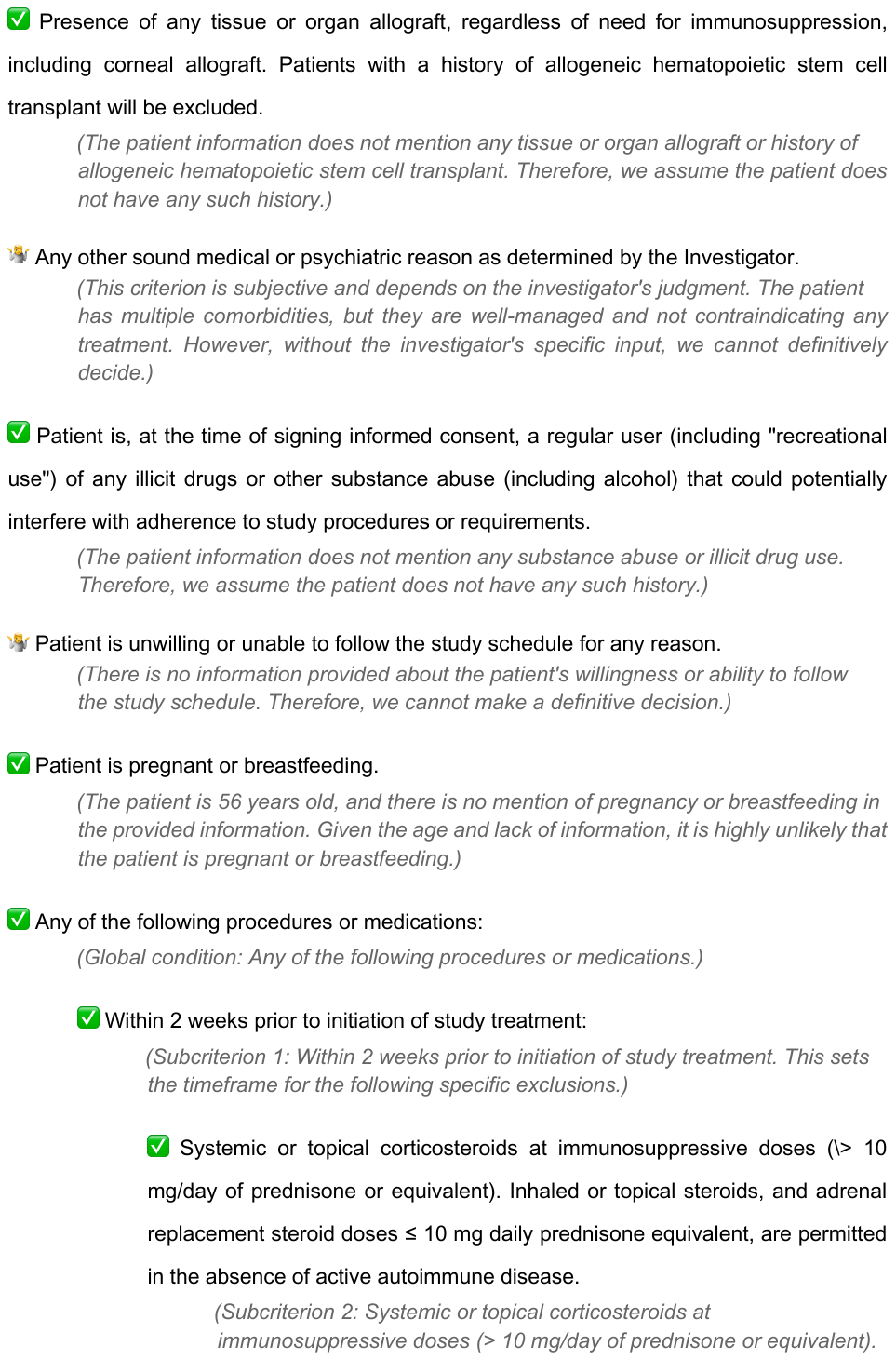}
  \label{fig:RatedTrial6}
\end{figure}

\clearpage

\begin{figure}[ht!]
  \centering
  \includegraphics[width=\linewidth]{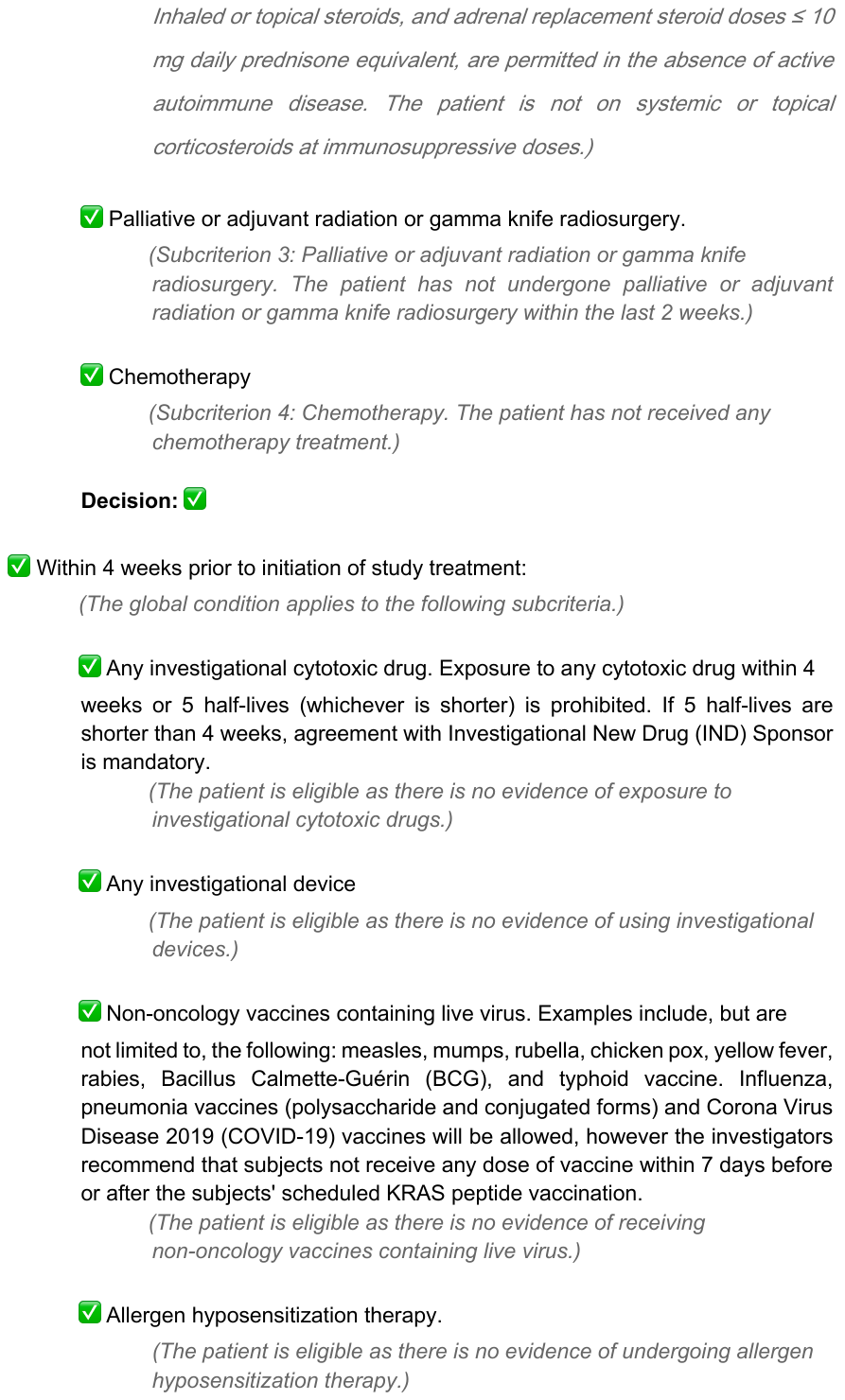}
  \label{fig:RatedTrial7}
\end{figure}

\clearpage

\begin{figure}[htbp]
  \centering
  \includegraphics[width=\linewidth]{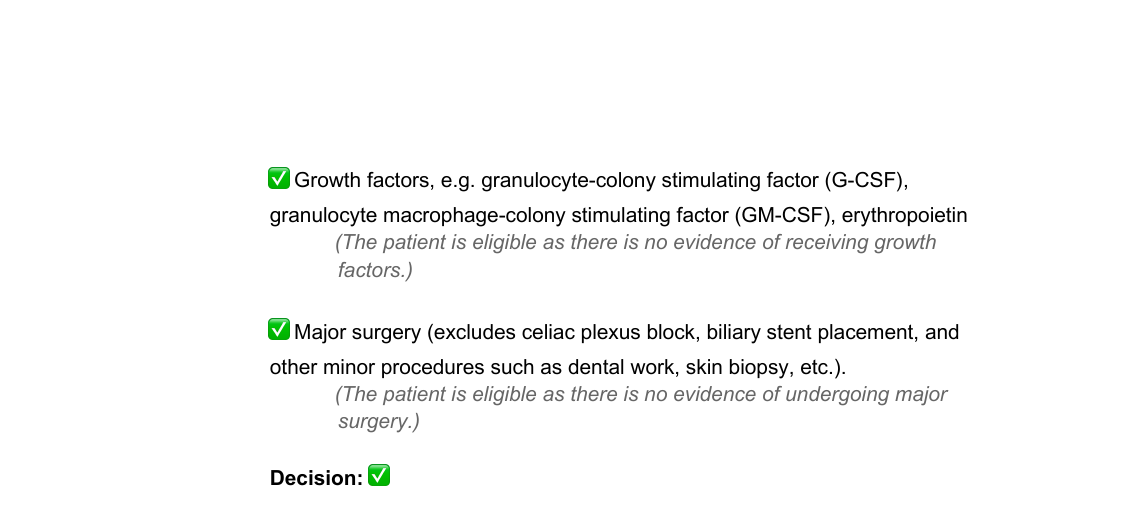}
  \caption{\textbf{Supplementary Table 4. Fully evaluated clinical trial eligibility criteria for Patient 1.1.1.} This table presents the fully annotated (evaluated) eligibility criteria for the clinical trial “KRAS-Targeted Vaccine With Nivolumab and Ipilimumab for Patients With NSCLC” (NCT05254184) as determined by our model for patient 1.1.1 (see Supplementary Table 5). The shown trial is a single-institution, Phase 1 study aimed to evaluate the safety and feasibility of a mutant-KRAS peptide vaccine in combination with Nivolumab and Ipilimumab for the first-line treatment of patients with advanced stage III/IV unresectable KRAS-mutated non-small cell lung cancer, and to estimate progression-free survival and T cell responses in the peripheral blood. The model’s decisions are indicated on a single criterion level basis and include a final “Decision” statement whenever encountering a situation where multiple criteria are nested. Symbols denote patient eligibility (\raisebox{-0.5mm}{\includegraphics[height=1em]{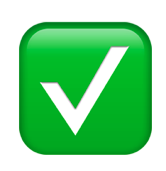}}), ineligibility (\raisebox{-0.5mm}{\includegraphics[height=1em]{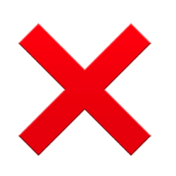}}), and insufficient information (\raisebox{-0.5mm}{\includegraphics[height=1em]{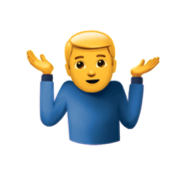}}). Gray italic text provides the model’s comments (reasoning) on its decision for each criterion. Note that we did not perform any attempts to optimize for the model's outputs at this step, which could include referencing the patient's EHR as source text or providing a more detailed chain of thought reasoning. Please also note that the order in which independent criteria are displayed here is not necessarily the same as on clinicaltrials.gov, which is a side effect of running evaluation across criteria in parallel to speed up the process.}
  \label{fig:RatedTrial8}
\end{figure}
\clearpage

\begin{figure}[ht!]
  \centering
  \includegraphics[width=\linewidth]{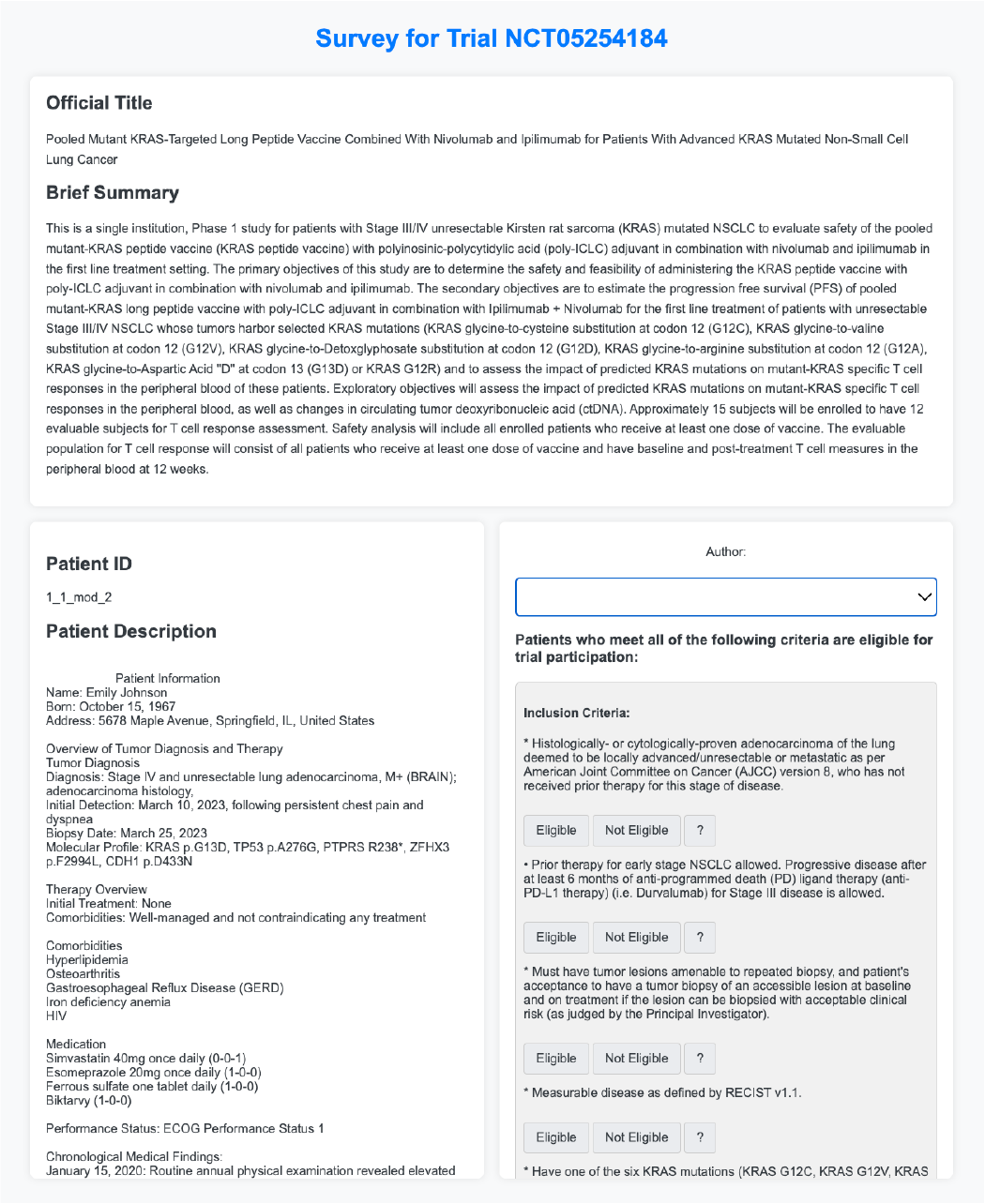}
  \label{fig:SupplementaryFigure1}
    \caption{\textbf{Supplementary Figure 1:} Web-based user interface for assessing trial eligibility criteria on a per patient basis. The top section shows the trial title and a free-text summary. On the left is the full text patient description. The right side features an interactive interface for selecting trial criteria.}
\end{figure}
\clearpage

\textbf{Supplementary Table 5. Clinical Cases EHRs.}\\
\textbf{Note:} Color highlighting is for illustration only, and not provided to the model.
\\
\\
\\
\begin{center}
\textbf{===== Patient 1.1 =====}
\end{center}

\textbf{Patient Information\\
}Name: Emily Johnson\\
Born: October 15, 1967\\
Address: 5678 Maple Avenue, Springfield, IL, United States

\textbf{Overview of Tumor Diagnosis and Therapy}

\textbf{Tumor Diagnosis\\
}Diagnosis: Stage IV and unresectable lung adenocarcinoma;
adenocarcinoma histology, PD-L1\\
Initial Detection: March 10, 2023, following persistent chest pain and
dyspnea\\
Biopsy Date: March 25, 2023\\
Molecular Profile: KRAS p.G13D, TP53 p.A276G, PTPRS R238*, ZFHX3
p.F2994L, CDH1 p.D433N

\textbf{Therapy Overview\\
}Initial Treatment: None\\
Comorbidities: Well-managed and not contraindicating any treatment

\textbf{Comorbidities}

Hyperlipidemia

Osteoarthritis

Gastroesophageal Reflux Disease (GERD)

Iron deficiency anemia

~

\textbf{Medication}

Simvastatin 40mg once daily (0-0-1)

Esomeprazole 20mg once daily (1-0-0)

Ferrous sulfate one tablet daily (1-0-0)

~

\textbf{Performance Status:} ECOG Performance Status 1

\textbf{Chronological Medical Findings:}

\textbf{January 15, 2020:} Routine annual physical examination revealed
elevated blood pressure. Diagnosed with hypertension and prescribed
Lisinopril 10 mg daily.

\textbf{March 20, 2020:} Complaints of frequent heartburn and acid
reflux. Diagnosed with GERD and prescribed Esomeprazole 20 mg daily.

\textbf{February 5, 2021:} Follow-up for hypertension showed
well-controlled blood pressure. Lisinopril dosage maintained.

\textbf{May 25, 2021:} Complained of knee pain and stiffness. Diagnosed
with osteoarthritis. Recommended over-the-counter NSAIDs for pain
management.

\textbf{April 15, 2022:} Routine check-up revealed elevated blood
glucose levels. Diagnosed with Type 2 Diabetes Mellitus and prescribed
Metformin 500 mg twice daily.

\textbf{August 10, 2022:} Routine cholesterol check indicated high
cholesterol levels. Diagnosed with hyperlipidemia and prescribed
Simvastatin 20 mg daily.

\textbf{November 15, 2022:} Follow-up for diabetes and hyperlipidemia.
Dosages adjusted: Metformin increased to 1000 mg twice daily,
Simvastatin increased to 40 mg daily.

\textbf{February 20, 2023:} Complained of shortness of breath and
chronic cough. Diagnosed with mild COPD. Prescribed Salbutamol inhaler.

\textbf{July 10, 2023:} Routine follow-up showed stable condition with
controlled comorbidities. Blood pressure, blood sugar, and cholesterol
levels within target ranges.

~

\textbf{March 10, 2024:} Experienced persistent chest pain and shortness
of breath. Chest X-ray and CT scan revealed a mass in the right lung.

\textbf{March 15, 2024:} CT Angiography, Pulmonary Arteries: Tumor Size:
Approximately 4.7 cm in diameter. Bronchial Obstruction: Partial
obstruction of the right main bronchus leading to atelectasis of the
right upper lobe. Urgent suspicion of a tumor-atelectasis complex in the
right upper lobe of the lung. Mucus present in the lower lobe bronchi on
the right. Suspicion of mediastinal lymph node metastases. No evidence
of pulmonary artery embolism. Lymph Nodes: Enlarged, FDG-positive lymph
nodes in the mediastinum.

\textbf{March 25, 2024:} CT-guided lung biopsy: Diagnosed with non-small
cell lung cancer (NSCLC), adenocarcinoma. Molecular diagnostics: KRAS
p.G13D, TP53 p.A276G, PTPRS R238*, ZFHX3 p.F2994L, CDH1 p.D433N.

\textbf{April 20, 2024:} Detailed assessment of health status. ECOG
performance status 1.~

\textbf{April 21, 2024: Routine Lab:}

Leukocytes 4,200/mcL, Lymphocytes 600/mm³, Absolute Neutrophil Count
(ANC) 1,200/mcL

Platelets 150 × 10³/uL

Hemoglobin 9.0 g/dL

Total Bilirubin 1.3 mg/dL

Aspartate Aminotransferase (AST) 50 U/L

Alanine Aminotransferase (ALT) 60 U/L

Alkaline Phosphatase 200 U/L

Creatinine 1.4 mg/dL

L-thyroxin (T4) 8.5 µg/dL

Thyroid Stimulating Hormone (TSH) 2.0 µIU/mL

Blood Glucose 90 mg/dL

Cholesterol 180 mg/dL

HCG within normal range.

\textbf{April 22, 2024:} Molecular tumor board: Recommendation for trial
inclusion.

~
\begin{center}
\textbf{===== Patient 1.1.1 =====}
\end{center}
\textbf{Patient Information\\
}Name: Emily Johnson\\
Born: October 15, 1967\\
Address: 5678 Maple Avenue, Springfield, IL, United States

\textbf{Overview of Tumor Diagnosis and Therapy}

\textbf{Tumor Diagnosis\\
}Diagnosis: Stage IV and unresectable lung adenocarcinoma;
adenocarcinoma histology, PD-L1\\
Initial Detection: March 10, 2023, following persistent chest pain and
dyspnea\\
Biopsy Date: March 25, 2023\\
Molecular Profile: KRAS p.G13D, TP53 p.A276G, PTPRS R238*, ZFHX3
p.F2994L, CDH1 p.D433N

\textbf{Therapy Overview\\
}Initial Treatment: None\\
Comorbidities: Well-managed and not contraindicating any treatment

\textbf{Comorbidities}

Hyperlipidemia

Osteoarthritis

Gastroesophageal Reflux Disease (GERD)

Iron deficiency anemia

~

\textbf{Medication}

Simvastatin 40mg once daily (0-0-1)

Esomeprazole 20mg once daily (1-0-0)

Ferrous sulfate one tablet daily (1-0-0)

~

\textbf{Performance Status:} ECOG Performance Status 1

\textbf{Chronological Medical Findings:}

\textbf{January 15, 2020:} Routine annual physical examination revealed
elevated blood pressure. Diagnosed with hypertension and prescribed
Lisinopril 10 mg daily.

\textbf{March 20, 2020:} Complaints of frequent heartburn and acid
reflux. Diagnosed with GERD and prescribed Esomeprazole 20 mg daily.

\textbf{February 5, 2021:} Follow-up for hypertension showed
well-controlled blood pressure. Lisinopril dosage maintained.

\textbf{May 25, 2021:} Complained of knee pain and stiffness. Diagnosed
with osteoarthritis. Recommended over-the-counter NSAIDs for pain
management.

\textbf{April 15, 2022:} Routine check-up revealed elevated blood
glucose levels. Diagnosed with Type 2 Diabetes Mellitus and prescribed
Metformin 500 mg twice daily.

\textbf{August 10, 2022:} Routine cholesterol check indicated high
cholesterol levels. Diagnosed with hyperlipidemia and prescribed
Simvastatin 20 mg daily.

\textbf{November 15, 2022:} Follow-up for diabetes and hyperlipidemia.
Dosages adjusted: Metformin increased to 1000 mg twice daily,
Simvastatin increased to 40 mg daily.

\textbf{February 20, 2023:} Complained of shortness of breath and
chronic cough. Diagnosed with mild COPD. Prescribed Salbutamol inhaler.

\textbf{July 10, 2023:} Routine follow-up showed stable condition with
controlled comorbidities. Blood pressure, blood sugar, and cholesterol
levels within target ranges.

~

\textbf{March 10, 2024:} Experienced persistent chest pain and shortness
of breath. Chest X-ray and CT scan revealed a mass in the right lung.

\textbf{March 15, 2024:} CT Angiography, Pulmonary Arteries: Tumor Size:
Approximately 4.7 cm in diameter. Bronchial Obstruction: Partial
obstruction of the right main bronchus leading to atelectasis of the
right upper lobe. Urgent suspicion of a tumor-atelectasis complex in the
right upper lobe of the lung. Mucus present in the lower lobe bronchi on
the right. Suspicion of mediastinal lymph node metastases. No evidence
of pulmonary artery embolism. Lymph Nodes: Enlarged, FDG-positive lymph
nodes in the mediastinum.

\textbf{March 25, 2024:} CT-guided lung biopsy: Diagnosed with non-small
cell lung cancer (NSCLC), adenocarcinoma. Molecular diagnostics: KRAS
p.G13D, TP53 p.A276G, PTPRS R238*, ZFHX3 p.F2994L, CDH1 p.D433N.

\textbf{April 20, 2024:} Detailed assessment of health status. ECOG
performance status 1.~

\textbf{April 21, 2024: Routine Lab:}

\colorbox{orange}{Leukocytes 2,700/mcL, Lymphocytes 438/mm³}, Absolute Neutrophil Count
(ANC) 1,200/mcL

Platelets 150 × 10³/uL

Hemoglobin 9.0 g/dL

Total Bilirubin 1.3 mg/dL

Aspartate Aminotransferase (AST) 50 U/L

Alanine Aminotransferase (ALT) 60 U/L

Alkaline Phosphatase 200 U/L

Creatinine 1.4 mg/dL

L-thyroxin (T4) 8.5 µg/dL

Thyroid Stimulating Hormone (TSH) 2.0 µIU/mL

Blood Glucose 90 mg/dL

Cholesterol 180 mg/dL

\colorbox{orange}{HCG +++++}

\textbf{April 22, 2024:} Molecular tumor board: Recommendation for trial
inclusion.

~

\begin{center}
\textbf{===== Patient 1.1.2 =====}
\end{center}
\textbf{Patient Information\\
}Name: Emily Johnson\\
Born: October 15, 1967\\
Address: 5678 Maple Avenue, Springfield, IL, United States

~

\textbf{Overview of Tumor Diagnosis and Therapy}

\textbf{Tumor Diagnosis\\
}Diagnosis: Stage IV and unresectable lung adenocarcinoma, \colorbox{orange}{M+ (BRAIN)};
adenocarcinoma histology,

Initial Detection: March 10, 2023, following persistent chest pain and
dyspnea\\
Biopsy Date: March 25, 2023\\
Molecular Profile: KRAS p.G13D, TP53 p.A276G, PTPRS R238*, ZFHX3
p.F2994L, CDH1 p.D433N

~

\textbf{Therapy Overview\\
}Initial Treatment: None\\
Comorbidities: Well-managed and not contraindicating any treatment

~

\textbf{Comorbidities}

Hyperlipidemia

Osteoarthritis

Gastroesophageal Reflux Disease (GERD)

Iron deficiency anemia

\colorbox{orange}{HIV}

~

\textbf{Medication}

Simvastatin 40mg once daily (0-0-1)

Esomeprazole 20mg once daily (1-0-0)

Ferrous sulfate one tablet daily (1-0-0)

\colorbox{orange}{Biktarvy (1-0-0)}

~

\textbf{Performance Status:} ECOG Performance Status 1

~

\textbf{Chronological Medical Findings:}

\textbf{January 15, 2020:} Routine annual physical examination revealed
elevated blood pressure. Diagnosed with hypertension and prescribed
Lisinopril 10 mg daily.

\textbf{March 20, 2020:} Complaints of frequent heartburn and acid
reflux. Diagnosed with GERD and prescribed Esomeprazole 20 mg daily.

\textbf{February 5, 2021:} Follow-up for hypertension showed
well-controlled blood pressure. Lisinopril dosage maintained.

\textbf{May 25, 2021:} Complained of knee pain and stiffness. Diagnosed
with osteoarthritis. Recommended over-the-counter NSAIDs for pain
management.

\textbf{April 15, 2022:} Routine check-up revealed elevated blood
glucose levels. Diagnosed with Type 2 Diabetes Mellitus and prescribed
Metformin 500 mg twice daily.

\textbf{August 10, 2022:} Routine cholesterol check indicated high
cholesterol levels. Diagnosed with hyperlipidemia and prescribed
Simvastatin 20 mg daily.

\textbf{November 15, 2022:} Follow-up for diabetes and hyperlipidemia.
Dosages adjusted: Metformin increased to 1000 mg twice daily,
Simvastatin increased to 40 mg daily.

\textbf{February 20, 2023:} Complained of shortness of breath and
chronic cough. Diagnosed with mild COPD. Prescribed Salbutamol inhaler.

\textbf{July 10, 2023:} Routine follow-up showed stable condition with
controlled comorbidities. Blood pressure, blood sugar, and cholesterol
levels within target ranges.

\textbf{March 10, 2024:} Experienced persistent chest pain and shortness
of breath. Chest X-ray and CT scan revealed a mass in the right lung.

\textbf{March 15, 2024:} CT Angiography, Pulmonary Arteries: Tumor Size:
Approximately 4.7 cm in diameter. Bronchial Obstruction: Partial
obstruction of the right main bronchus leading to atelectasis of the
right upper lobe. Urgent suspicion of a tumor-atelectasis complex in the
right upper lobe of the lung. Mucus present in the lower lobe bronchi on
the right. Suspicion of mediastinal lymph node metastases. No evidence
of pulmonary artery embolism. Lymph Nodes: Enlarged, FDG-positive lymph
nodes in the mediastinum.

\colorbox{orange}{\textbf{March 20, 2024:} MRI-Brain: three metastatic lesions consistent with primary } \colorbox{orange}{lung cancer. Lesions are located in the left frontal lobe, left parietal lobe, and } \colorbox{orange}{left occipital lobe, measuring 1.2 cm, 1.5 cm, and 1.8 cm in diameter, respectively.}\\
\colorbox{orange}{Surrounding vasogenic edema is noted, causing mild mass effect on adjacent}\\
\colorbox{orange}{brain structures. No evidence of hemorrhage or hydrocephalus observed.}

\textbf{March 25, 2024:} CT-guided lung biopsy: Diagnosed with non-small
cell lung cancer (NSCLC), adenocarcinoma. Molecular diagnostics: KRAS
p.G13D, TP53 p.A276G, PTPRS R238*, ZFHX3 p.F2994L, CDH1 p.D433N.

\textbf{April 20, 2024:} Detailed assessment of health status. ECOG
performance status 1.~

\textbf{April 21, 2024: Routine Lab:}

Leukocytes 4,200/mcL, Lymphocytes 600/mm³, Absolute Neutrophil Count
(ANC) 1,200/mcL

Platelets 150 × 10³/uL

Hemoglobin 9.0 g/dL

Total Bilirubin 1.3 mg/dL

Aspartate Aminotransferase (AST) 50 U/L

Alanine Aminotransferase (ALT) 60 U/L

Alkaline Phosphatase 200 U/L

Creatinine 1.4 mg/dL

L-thyroxin (T4) 8.5 µg/dL

Thyroid Stimulating Hormone (TSH) 2.0 µIU/mL

Blood Glucose 90 mg/dL

Cholesterol 180 mg/dL

HCG within normal range.

\colorbox{orange}{SO2 (room air) 87\%}

\textbf{April 22, 2024:} Molecular tumor board: Recommendation for trial
inclusion.

~
\begin{center}
\textbf{===== Patient 2.1 =====}
\end{center}

\textbf{Patient Information\\
}Name: Sarah Mitchell\\
Born: June 12, 1998\\
Address: 8765 Pine Street, Springfield, IL

~

\textbf{Overview of Tumor Diagnosis and Therapy}

\textbf{Tumor Diagnosis\\
}Diagnosis: Stage IV urachal adenocarcinoma; m+ (LYM, PULMONARY)\\
Initial Detection: January 18, 2024, following persistent hematuria and
abdominal discomfort\\
Biopsy Date: January 28, 2024\\
Molecular Profile: KRAS p.G12V, BCORL p.R1332*, TP53 p.H214fs\emph{7,
CDKN2C p.L65F, MAP3K1 p.T949\_E950insT, MYCN p.E47fs}8, CTNNA1
p.K577\_L578\textgreater TKL, JAK1 p.I597M, FANCL p.T367fs*12+, PIK3CA
amplification (n\textgreater6), MYC amplification (n\textgreater6),
MYCL1 amplification (n\textgreater6), SOX2 amplification
(n\textgreater6), MUTYH amplification (n\textgreater6)

~

\textbf{Therapy Overview\\
}Chemotherapy: Began February 1 - April 22, 2024 (Cisplatin + 5-FU)

~

\textbf{Comorbidities}

Seasonal Allergies

~

\textbf{Medication}

Cetirizine 10mg as needed

~

\textbf{Performance Status:} ECOG Performance Status 1

~

\textbf{Chronological Medical Findings:}

\textbf{January 18, 2024:} Presented with hematuria and abdominal
discomfort. Abdominal ultrasound revealed a mass in the bladder dome.

\textbf{January 22, 2024:} CT scan abdomen/pelvis: Mass located at the
bladder dome, measuring approximately 3.0 cm in diameter. Evidence of
local invasion into surrounding structures + several enlarged local
lymph nodes are noted, with the largest lymph node located near the
pelvic sidewall, measuring 1.5 cm.

Chest CT scan: Multiple metastatic lesions are present in both lungs.
The largest metastasis is located peripherally in the left lung and
measures approximately 3.1 cm in diameter. Other smaller metastatic
nodules scattered throughout both lung fields.

\textbf{January 28, 2024:} Multiple Biopsies of bladder (CT-guided):
Histology confirmed urachal adenocarcinoma. Molecular panel sequencing
revealed mutations: KRAS p.G12V, BCORL p.R1332*, TP53 p.H214fs\emph{7,
CDKN2C p.L65F, MAP3K1 p.T949\_E950insT, MYCN p.E47fs}8, CTNNA1
p.K577\_L578\textgreater TKL, JAK1 p.I597M, FANCL p.T367fs*12+, PIK3CA
amplification (n\textgreater6), MYC amplification (n\textgreater6),
MYCL1 amplification (n\textgreater6), SOX2 amplification
(n\textgreater6), MUTYH amplification (n\textgreater6).

\textbf{February 1, 2024:} Initiated chemotherapy with Cisplatin + 5-FU.

\textbf{April 23, 2024:} CT scan abdomen/pelvis: Mass located at the
bladder dome, now measuring approximately 4.5 cm in diameter. Increased
evidence of local invasion into surrounding structures. Several enlarged
local lymph nodes are noted, with the largest lymph node located near
the pelvic sidewall, now measuring 2.0 cm. Chest CT scan: Increased
number and size of metastatic lesions in both lungs. The largest
metastasis is located peripherally in the left lung and now measures
approximately 4.0 cm in diameter. Numerous other metastatic nodules,
with some showing an increase in size, are scattered throughout both
lung fields.

\textbf{April 25, 2024:} Detailed assessment of health status confirmed
adequate organ function. Routine labs within normal limits: ANC
4,500/mcL, platelet count 250,000/mcL, total bilirubin 0.8 mg/dL,
AST/ALT within normal limits, creatinine 0.8 mg/dL, hemoglobin 14.0
g/dL, serum albumin 4.0 g/dL, lipase and amylase within normal limits.
Serum HCG test negative.~

\textbf{April 28, 2024:} Tumor board review recommended considering
eligibility for clinical trials due to limited response to standard and
investigational therapies. Patient in good clinical condition, willing
to participate in a trial.

~
\begin{center}
\textbf{===== Patient 2.1.1 =====}
\end{center}
~

\textbf{Patient Information\\
}Name: Sarah Mitchell\\
Born: June 12, 1998\\
Address: 8765 Pine Street, Springfield, IL

~

\textbf{Overview of Tumor Diagnosis and Therapy}

\textbf{Tumor Diagnosis\\
}Diagnosis: Stage IV urachal adenocarcinoma; m+ (LYM, PULMONARY)\\
Initial Detection: January 18, 2024, following persistent hematuria and
abdominal discomfort\\
Biopsy Date: January 28, 2024\\
Molecular Profile: KRAS p.G12V, BCORL p.R1332*, TP53 p.H214fs\emph{7,
CDKN2C p.L65F, MAP3K1 p.T949\_E950insT, MYCN p.E47fs}8, CTNNA1
p.K577\_L578\textgreater TKL, JAK1 p.I597M, FANCL p.T367fs*12+, PIK3CA
amplification (n\textgreater6), MYC amplification (n\textgreater6),
MYCL1 amplification (n\textgreater6), SOX2 amplification
(n\textgreater6), MUTYH amplification (n\textgreater6)

~

\textbf{Therapy Overview\\
}Chemotherapy: Began February 1 - April 22, 2024 (Cisplatin + 5-FU)

~

\textbf{Comorbidities}

Seasonal Allergies

\colorbox{orange}{Platin-induced Neuropathy (April 2024)}

~

\textbf{Medication}

Cetirizine 10mg as needed

~

\textbf{Performance Status:} ECOG Performance Status 1

~

\textbf{Chronological Medical Findings:}

\textbf{January 18, 2024:} Presented with hematuria and abdominal
discomfort. Abdominal ultrasound revealed a mass in the bladder dome.

\textbf{January 22, 2024:} CT scan abdomen/pelvis: Mass located at the
bladder dome, measuring approximately 3.0 cm in diameter. Evidence of
local invasion into surrounding structures + several enlarged local
lymph nodes are noted, with the largest lymph node located near the
pelvic sidewall, measuring 1.5 cm.

Chest CT scan: Multiple metastatic lesions are present in both lungs.
The largest metastasis is located peripherally in the left lung and
measures approximately 3.1 cm in diameter. Other smaller metastatic
nodules scattered throughout both lung fields.

\textbf{January 28, 2024:} Multiple Biopsies of bladder (CT-guided):
Histology confirmed urachal adenocarcinoma. Molecular panel sequencing
revealed mutations: KRAS p.G12V, BCORL p.R1332*, TP53 p.H214fs\emph{7,
CDKN2C p.L65F, MAP3K1 p.T949\_E950insT, MYCN p.E47fs}8, CTNNA1
p.K577\_L578\textgreater TKL, JAK1 p.I597M, FANCL p.T367fs*12+, PIK3CA
amplification (n\textgreater6), MYC amplification (n\textgreater6),
MYCL1 amplification (n\textgreater6), SOX2 amplification
(n\textgreater6), MUTYH amplification (n\textgreater6).

\textbf{February 1, 2024:} Initiated chemotherapy with Cisplatin + 5-FU.\\
\colorbox{orange}{Chemotherapy abrogated before completion of the last cycle due to severe}\\
\colorbox{orange}{neuropathy limiting daily activities.}

\textbf{April 23, 2024:} CT scan abdomen/pelvis: Mass located at the
bladder dome, now measuring approximately 4.5 cm in diameter. Increased
evidence of local invasion into surrounding structures. Several enlarged
local lymph nodes are noted, with the largest lymph node located near
the pelvic sidewall, now measuring 2.0 cm. Chest CT scan: Increased
number and size of metastatic lesions in both lungs. The largest
metastasis is located peripherally in the left lung and now measures
approximately 4.0 cm in diameter. Numerous other metastatic nodules,
with some showing an increase in size, are scattered throughout both
lung fields.

\colorbox{orange}{cMRI: 1 single brain metastasis in the frontal lobe.}

\textbf{April 25, 2024:} Detailed assessment of health status confirmed
adequate organ function. Routine labs within normal limits: ANC
4,500/mcL, platelet count 250,000/mcL, total bilirubin 0.8 mg/dL,
AST/ALT within normal limits, creatinine 0.8 mg/dL, hemoglobin 14.0
g/dL, serum albumin 4.0 g/dL, lipase and amylase within normal limits.
Serum HCG test negative.~

\textbf{April 28, 2024:} Tumor board review recommended considering
eligibility for clinical trials due to limited response to standard and
investigational therapies.\\
\colorbox{orange}{Patient in overall good clinical condition,
however persistent neuropathy}\\
\colorbox{orange}{(no improvements).} She is willing to
participate in clinical trials.

~
\begin{center}
\textbf{===== Patient 2.1.2 =====}~
\end{center}
\textbf{Patient Information\\
}Name: Sarah Mitchell\\
Born: June 12, 1998\\
Address: 8765 Pine Street, Springfield, IL

~

\textbf{Overview of Tumor Diagnosis and Therapy}

\textbf{Tumor Diagnosis\\
}Diagnosis: Stage IV urachal adenocarcinoma; m+ (LYM, PULMONARY)\\
Initial Detection: January 18, 2024, following persistent hematuria and
abdominal discomfort\\
Biopsy Date: January 28, 2024\\
Molecular Profile: KRAS p.G12V, BCORL p.R1332*, TP53 p.H214fs\emph{7,
CDKN2C p.L65F, MAP3K1 p.T949\_E950insT, MYCN p.E47fs}8, CTNNA1
p.K577\_L578\textgreater TKL, JAK1 p.I597M, FANCL p.T367fs*12+, PIK3CA
amplification (n\textgreater6), MYC amplification (n\textgreater6),
MYCL1 amplification (n\textgreater6), SOX2 amplification
(n\textgreater6), MUTYH amplification (n\textgreater6)

~

\textbf{Therapy Overview\\
}Chemotherapy: Began February 1 - April 22, 2024 (Cisplatin + 5-FU)

~

\textbf{Comorbidities}

\colorbox{orange}{Diabetes type II}

Seasonal Allergies

~

\textbf{Medication}

Cetirizine 10mg as needed\\
\colorbox{orange}{Ceftriaxone 1g 1-0-0}\\
Metformin 500mg BID (paused)\\
Glyburide 5mg BID (paused)\\
Insulin under monitoring\\

~

\textbf{Performance Status:} ECOG Performance Status 1

~

\textbf{Chronological Medical Findings:}

\textbf{January 18, 2024:} Presented with hematuria and abdominal
discomfort. Abdominal ultrasound revealed a mass in the bladder dome.

\textbf{January 22, 2024:} CT scan abdomen/pelvis: Mass located at the
bladder dome, measuring approximately 3.0 cm in diameter. Evidence of
local invasion into surrounding structures + several enlarged local
lymph nodes are noted, with the largest lymph node located near the
pelvic sidewall, measuring 1.5 cm.

Chest CT scan: Multiple metastatic lesions are present in both lungs.
The largest metastasis is located peripherally in the left lung and
measures approximately 3.1 cm in diameter. Other smaller metastatic
nodules scattered throughout both lung fields.

\textbf{January 28, 2024:} Multiple Biopsies of bladder (CT-guided):
Histology confirmed urachal adenocarcinoma. Molecular panel sequencing
revealed mutations: KRAS p.G12V, BCORL p.R1332*, TP53 p.H214fs\emph{7,
CDKN2C p.L65F, MAP3K1 p.T949\_E950insT, MYCN p.E47fs}8, CTNNA1
p.K577\_L578\textgreater TKL, JAK1 p.I597M, FANCL p.T367fs*12+, PIK3CA
amplification (n\textgreater6), MYC amplification (n\textgreater6),
MYCL1 amplification (n\textgreater6), SOX2 amplification
(n\textgreater6), MUTYH amplification (n\textgreater6).

\textbf{February 1, 2024:} Initiated chemotherapy with Cisplatin + 5-FU.

\textbf{April 23, 2024:} CT scan abdomen/pelvis: Mass located at the
bladder dome, now measuring approximately 4.5 cm in diameter. Increased
evidence of local invasion into surrounding structures. Several enlarged
local lymph nodes are noted, with the largest lymph node located near
the pelvic sidewall, now measuring 2.0 cm. Chest CT scan: Increased
number and size of metastatic lesions in both lungs. The largest
metastasis is located peripherally in the left lung and now measures
approximately 4.0 cm in diameter. Numerous other metastatic nodules,
with some showing an increase in size, are scattered throughout both
lung fields.

\textbf{April 25, 2024:} Detailed assessment of health status confirmed
adequate organ function. Routine labs within normal limits: ANC
4,500/mcL, platelet count 250,000/mcL, total bilirubin 0.8 mg/dL,
AST/ALT within normal limits, creatinine 0.8 mg/dL, hemoglobin 14.0
g/dL, serum albumin 4.0 g/dL, lipase and amylase within normal limits.
Serum HCG test negative.~

\textbf{April 28, 2024:} Tumor board review recommended considering
eligibility for clinical trials due to limited response to standard and
investigational therapies. Patient in good clinical condition, willing
to participate in a trial.

\colorbox{orange}{\textbf{May 1, 2024:} Patient presents with fever, flank pain, and dysuria.}\\
\colorbox{orange}{Hospitalized for further evaluation and treatment.}\\
\colorbox{orange}{Ultrasound: Enlarged kidney with signs of}\\
\colorbox{orange}{inflammation, consistent with pyelonephritis.}\\
\colorbox{orange}{Blood culture: Pending. Urine culture: Pending. Started on IV antibiotics:}\\
\colorbox{orange}{Ceftriaxone 1g.~ CRP: 15 mg/dL. Leukocytes:
18,000/mcL. HbA1c 8.3\%.}\\
\colorbox{orange}{Paused Metformin/Glyburide, started on insulin with close monitoring}

~
\begin{center}
\textbf{===== Patient 2.2 =====}
\end{center}
\textbf{Patient Information\\
}Name: Sarah Mitchell\\
Born: June 12, 1998\\
Address: 8765 Pine Street, Springfield, IL

~

\textbf{Overview of Tumor Diagnosis and Therapy}

\textbf{Tumor Diagnosis\\
}Diagnosis: Stage IV urachal adenocarcinoma; m+ (LYM, PULMONARY)\\
Initial Detection: January 18, 2024, following persistent hematuria and
abdominal discomfort\\
Biopsy Date: January 28, 2024\\
Molecular Profile: KRAS p.G12V, BCORL p.R1332*, TP53 p.H214fs\emph{7,
CDKN2C p.L65F, MAP3K1 p.T949\_E950insT, MYCN p.E47fs}8, CTNNA1
p.K577\_L578\textgreater TKL, JAK1 p.I597M, FANCL p.T367fs*12+, PIK3CA
amplification (n\textgreater6), MYC amplification (n\textgreater6),
MYCL1 amplification (n\textgreater6), SOX2 amplification
(n\textgreater6), MUTYH amplification (n\textgreater6)

~

\textbf{Therapy Overview\\
}Chemotherapy: Began February 1 - April 22, 2024 (Cisplatin + 5-FU)

~

\textbf{Comorbidities}

Seasonal Allergies

~

\textbf{Medication}

Cetirizine 10mg as needed

~

\textbf{Performance Status:} ECOG Performance Status 1

~

\textbf{Chronological Medical Findings:}

\textbf{January 18, 2024:} Presented with hematuria and abdominal
discomfort. Abdominal ultrasound revealed a mass in the bladder dome.

\textbf{January 22, 2024:} CT scan abdomen/pelvis: Mass located at the
bladder dome, measuring approximately 3.0 cm in diameter. Evidence of
local invasion into surrounding structures + several enlarged local
lymph nodes are noted, with the largest lymph node located near the
pelvic sidewall, measuring 1.5 cm.

Chest CT scan: Multiple metastatic lesions are present in both lungs.
The largest metastasis is located peripherally in the left lung and
measures approximately 3.1 cm in diameter. Other smaller metastatic
nodules scattered throughout both lung fields.

\textbf{January 28, 2024:} Multiple Biopsies of bladder (CT-guided):
Histology confirmed urachal adenocarcinoma. Molecular panel sequencing
revealed mutations: KRAS p.G12V, BCORL p.R1332*, TP53 p.H214fs\emph{7,
CDKN2C p.L65F, MAP3K1 p.T949\_E950insT, MYCN p.E47fs}8, CTNNA1
p.K577\_L578\textgreater TKL, JAK1 p.I597M, FANCL p.T367fs*12+, PIK3CA
amplification (n\textgreater6), MYC amplification (n\textgreater6),
MYCL1 amplification (n\textgreater6), SOX2 amplification
(n\textgreater6), MUTYH amplification (n\textgreater6).

\textbf{February 1, 2024:} Initiated chemotherapy with Cisplatin + 5-FU.

\textbf{April 23, 2024:} CT scan abdomen/pelvis: Mass located at the
bladder dome, now measuring approximately 4.5 cm in diameter. Increased
evidence of local invasion into surrounding structures. Several enlarged
local lymph nodes are noted, with the largest lymph node located near
the pelvic sidewall, now measuring 2.0 cm. Chest CT scan: Increased
number and size of metastatic lesions in both lungs. The largest
metastasis is located peripherally in the left lung and now measures
approximately 4.0 cm in diameter. Numerous other metastatic nodules,
with some showing an increase in size, are scattered throughout both
lung fields.

\textbf{April 25, 2024:} Detailed assessment of health status confirmed
adequate organ function. Routine labs within normal limits: ANC
4,500/mcL, platelet count 250,000/mcL, total bilirubin 0.8 mg/dL,
AST/ALT within normal limits, creatinine 0.8 mg/dL, hemoglobin 14.0
g/dL, serum albumin 4.0 g/dL, lipase and amylase within normal limits.
Serum HCG test negative.~

\textbf{April 28, 2024:} Tumor board review recommended considering
eligibility for clinical trials due to limited response to standard and
investigational therapies. Patient in good clinical condition, willing
to participate in a trial.

~

\begin{center}
\textbf{===== Patient 2.2.1 =====}
\end{center}
\textbf{Patient Information\\
}Name: Sarah Mitchell\\
Born: June 12, 1998\\
Address: 8765 Pine Street, Springfield, IL

~

\textbf{Overview of Tumor Diagnosis and Therapy}

\textbf{Tumor Diagnosis\\
}Diagnosis: Stage IV urachal adenocarcinoma; m+ (LYM, PULMONARY)\\
Initial Detection: January 18, 2024, following persistent hematuria and
abdominal discomfort\\
Biopsy Date: January 28, 2024\\
Molecular Profile: KRAS p.G12V, BCORL p.R1332*, TP53 p.H214fs\emph{7,
CDKN2C p.L65F, MAP3K1 p.T949\_E950insT, MYCN p.E47fs}8, CTNNA1
p.K577\_L578\textgreater TKL, JAK1 p.I597M, FANCL p.T367fs*12+, PIK3CA
amplification (n\textgreater6), MYC amplification (n\textgreater6),
MYCL1 amplification (n\textgreater6), SOX2 amplification
(n\textgreater6), MUTYH amplification (n\textgreater6)

~

\textbf{Therapy Overview\\
}Chemotherapy: Began February 1 - April 22, 2024 (Cisplatin + 5-FU)

~

\textbf{Comorbidities}

Seasonal Allergies

\colorbox{orange}{Recurrent gastrointestinal bleedings due to tumor infiltration of the rectum,}\\
\colorbox{orange}{requiring transfusions}

~

\textbf{Medication}

Cetirizine 10mg as needed

~

\textbf{Performance Status:} ECOG Performance Status 1

~

\textbf{Chronological Medical Findings:}

\textbf{January 18, 2024:} Presented with hematuria and abdominal
discomfort. Abdominal ultrasound revealed a mass in the bladder dome.

\textbf{January 22, 2024:} CT scan abdomen/pelvis: Mass located at the
bladder dome, measuring approximately 3.0 cm in diameter. Evidence of
local invasion into surrounding structures + several enlarged local
lymph nodes are noted, with the largest lymph node located near the
pelvic sidewall, measuring 1.5 cm.

Chest CT scan: Multiple metastatic lesions are present in both lungs.
The largest metastasis is located peripherally in the left lung and
measures approximately 3.1 cm in diameter. Other smaller metastatic
nodules scattered throughout both lung fields.

\textbf{January 28, 2024:} Multiple Biopsies of bladder (CT-guided):
Histology confirmed urachal adenocarcinoma. Molecular panel sequencing
revealed mutations: KRAS p.G12V, BCORL p.R1332*, TP53 p.H214fs\emph{7,
CDKN2C p.L65F, MAP3K1 p.T949\_E950insT, MYCN p.E47fs}8, CTNNA1
p.K577\_L578\textgreater TKL, JAK1 p.I597M, FANCL p.T367fs*12+, PIK3CA
amplification (n\textgreater6), MYC amplification (n\textgreater6),
MYCL1 amplification (n\textgreater6), SOX2 amplification
(n\textgreater6), MUTYH amplification (n\textgreater6).

\textbf{February 1, 2024:} Initiated chemotherapy with Cisplatin + 5-FU.

\textbf{April 23, 2024:} CT scan abdomen/pelvis: Mass located at the
bladder dome, now measuring approximately 4.5 cm in diameter. \colorbox{orange}{Rectal
tumor invasion.} Increased evidence of local invasion into surrounding
structures. Several enlarged local lymph nodes are noted, with the
largest lymph node located near the pelvic sidewall, now measuring 2.0
cm. Chest CT scan: Increased number and size of metastatic lesions in
both lungs. The largest metastasis is located peripherally in the left
lung and now measures approximately 4.0 cm in diameter. Numerous other
metastatic nodules, with some showing an increase in size, are scattered
throughout both lung fields.

\textbf{April 25, 2024:} Detailed assessment of health status confirmed
adequate organ function. Routine labs within normal limits: ANC
4,500/mcL, platelet count 250,000/mcL, total bilirubin 0.8 mg/dL,
AST/ALT within normal limits, creatinine 0.8 mg/dL, \colorbox{orange}{hemoglobin 8.4 g/dL}\\ \colorbox{orange}{(due to recurrent GI bleedings)}, serum albumin 4.0 g/dL, lipase and amylase within normal limits. Serum HCG test negative.~

\textbf{April 28, 2024:} Tumor board review recommended considering
eligibility for clinical trials due to limited response to standard and
investigational therapies. Patient in good clinical condition, willing
to participate in a trial.

~
\begin{center}
\textbf{===== Patient 3.1 =====}~
\end{center}
\textbf{Patient Information\\
}Name: Thomas Meyer\\
Born: January 12, 1966\\
Address: Schlossallee 1, Dresden, Germany

~

\textbf{Overview of Tumor Diagnosis and Therapy}

\textbf{Tumor Diagnosis\\
}Diagnosis: Masaoka-Koga Stage IVb thymic adenocarcinoma (metastases to
the lungs, liver and spines T6, T9, L3)\\
Initial Detection: March 15, 2023, following persistent chest pain and
cough, shortness-of-breath

Biopsy Date: March 28, 2023\\
Molecular Profile: Germline: BRCA2 p.K3326* (1N); Tumor: SMAD4 p.C363R
(1N), TP53 p.305fs (2N\_LOH), CDKN1B p.K100N (2N\_LOH), ATM p.E1666*
(4N), MAP3K8 p.H236Y (1N), TRAF1 p.R70H (2N), HDAC2 p.R409* (1N),
TMEM111-TDRD3 fusion, PRKDC-CDH17 fusion, EXT1-MAGI2 fusion;
Overexpressed genes: ERBB2, ERBB3, PDGFRB, TGFA, EGF, FGFR3, MET.

~

\textbf{Therapy Overview\\
}Initial Treatment:\\
Chemotherapy: Began April 5, 2023, with combination of Doxorubicin,
Cisplatin, Vincristine and Cyclophosphamide (ADOC). Partial response
after the initial two chemotherapy cycles completed by June, 2023.
Continued chemotherapy until November 2023 (progressive disease).\\
Subsequent Treatment: Second-line treatment with Carboplatin plus
Paclitaxel starting Nov 24, 2023. Current Status: Disease progression as
of May 2024, with new metastatic lesions in the lungs.

\textbf{Comorbidities}

\begin{quote}
Former Smoker 25 py

Hypertension Stage 2

Type 2 Diabetes Mellitus

Hyperlipidemia

Gastroesophageal Reflux Disease (GERD)

H/o cholecystectomy 2011
\end{quote}

\textbf{Medication}

\begin{quote}
Losartan 50mg once daily

HCT 12.5mg once daily

Metformin 1000mg once daily

Atorvastatin 40mg once daily

Omeprazole 20mg once daily

XGEVA~
\end{quote}

~

\textbf{Performance Status:} ECOG Performance Status 1

\textbf{Chronological Medical Findings:}

\textbf{March 15, 2023:} Presented with persistent chest pain and cough
and SOB.~

\textbf{March 20, 2023:} CT scan of the chest: Mass in the anterior
mediastinum measuring approximately 6.0 cm with evidence of local
invasion into surrounding structures. Multiple pulmonary nodules
suggestive of metastasis.~

\textbf{March 28, 2023:} CT-guided biopsy of mediastinal mass. Histology
confirmed thymic adenocarcinoma. Whole exome sequencing revealed
germline mutation BRCA2 p.K3326*, tumor mutations: SMAD4 p.C363R, TP53
p.305fs, CDKN1B p.K100N, ATM p.E1666*, MAP3K8 p.H236Y, TRAF1 p.R70H,
HDAC2 p.R409*, TMEM111-TDRD3 fusion, PRKDC-CDH17 fusion, EXT1-MAGI2
fusion. Overexpressed genes: ERBB2, ERBB3, PDGFRB, TGFA, EGF, FGFR3,
MET.

\textbf{April 5, 2023:} Initiated chemotherapy with Doxorubicin,
Cisplatin, Vincristine and Cyclophosphamide (ADOC). Patient in
sophisticated conditions, first therapy today.

\textbf{June 20, 2023:} Follow-up CT scan showed partial response with a
decrease in the size of the primary tumor and pulmonary nodules.
Continued chemotherapy regimen.

\textbf{November 15, 2023:} Follow-up imaging CT chest/abdomen: disease
progression with new metastatic lesions. Multiple hepatic lesions, with
the largest lesion in segment VIII measuring 4.5 cm, and another lesion
in segment II measuring 3.0 cm. Bone scan indicates metastatic
involvement of the spine, with lesions identified at T6, T9, and L3
vertebrae. Additional findings include new pulmonary nodules and further
enlargement of the primary mass in the anterior mediastinum.

\textbf{November 24, 2023:} Started second-line therapy with Carboplatin
plus Paclitaxel.~

\textbf{March 10, 2024:} Follow-up CT scan: PD. Progression of disease
with increased size of liver metastases and new bone lesions.

\textbf{May 20, 2024:} Tumor board review recommended considering
eligibility for clinical trials due to limited response to standard and
investigational therapies.

\textbf{June 1, 2024:} Detailed assessment of health status confirmed
adequate organ function. All routine labs within normal limits.
\\

~
\begin{center}
\textbf{===== Patient 3.1.1 =====}
\end{center}
\textbf{Patient Information\\
}Name: Thomas Meyer\\
Born: January 12, 1966\\
Address: Schlossallee 1, Dresden, Germany

~

\textbf{Overview of Tumor Diagnosis and Therapy}

\textbf{Tumor Diagnosis\\
}Diagnosis: Masaoka-Koga Stage IVb thymic adenocarcinoma (metastases to
the lungs, liver and spines T6, T9, L3)\\
Initial Detection: March 15, 2023, following persistent chest pain and
cough, shortness-of-breath

Biopsy Date: March 28, 2023\\
Molecular Profile: Germline: BRCA2 p.K3326* (1N); Tumor: SMAD4 p.C363R
(1N), TP53 p.305fs (2N\_LOH), CDKN1B p.K100N (2N\_LOH), ATM p.E1666*
(4N), MAP3K8 p.H236Y (1N), TRAF1 p.R70H (2N), HDAC2 p.R409* (1N),
TMEM111-TDRD3 fusion, PRKDC-CDH17 fusion, EXT1-MAGI2 fusion;
Overexpressed genes: ERBB2, ERBB3, PDGFRB, TGFA, EGF, FGFR3, MET.

~

\textbf{Therapy Overview\\
}Initial Treatment:\\
Chemotherapy: Began April 5, 2023, with combination of Doxorubicin,
Cisplatin, Vincristine and Cyclophosphamide (ADOC). Partial response
after the initial two chemotherapy cycles completed by June, 2023.
Continued chemotherapy until November 2023 (progressive disease).\\
Subsequent Treatment: Second-line treatment with Carboplatin plus
Paclitaxel starting Nov 24, 2023. Current Status: Disease progression as
of May 2024, with new metastatic lesions in the lungs.

\textbf{Comorbidities}

\begin{quote}
Former Smoker 25 py

\colorbox{orange}{Interstitial Lung disease (ILD)}

Hypertension Stage 2

Type 2 Diabetes Mellitus

Hyperlipidemia

Gastroesophageal Reflux Disease (GERD)

H/o cholecystectomy 2011
\end{quote}

\textbf{Medication}

~~~~~~~~ \colorbox{orange}{Prednisone 10mg 1x}
\begin{quote}
Losartan 50mg once daily

HCT 12.5mg once daily

Metformin 1000mg 1x/d

Atorvastatin 40mg once daily

Omeprazole 20mg daily

XGEVA q4w
\end{quote}

~

\textbf{Performance Status:} ECOG Performance Status 1

\textbf{Chronological Medical Findings:}

\textbf{March 15, 2023:} Presented with persistent chest pain and cough
and SOB.~

\textbf{March 20, 2023:} CT scan of the chest: Mass in the anterior
mediastinum measuring approximately 6.0 cm with evidence of local
invasion into surrounding structures. Multiple pulmonary nodules
suggestive of metastasis. \colorbox{orange}{Evidence of known Interstitial Lung Disease (ILD) with diffuse interstitial}\\
\colorbox{orange}{markings and fibrosis.}

\textbf{March 28, 2023:} CT-guided biopsy of mediastinal mass. Histology
confirmed thymic adenocarcinoma. Whole exome sequencing revealed
germline mutation BRCA2 p.K3326*, tumor mutations: SMAD4 p.C363R, TP53
p.305fs, CDKN1B p.K100N, ATM p.E1666*, MAP3K8 p.H236Y, TRAF1 p.R70H,
HDAC2 p.R409*, TMEM111-TDRD3 fusion, PRKDC-CDH17 fusion, EXT1-MAGI2
fusion. Overexpressed genes: ERBB2, ERBB3, PDGFRB, TGFA, EGF, FGFR3,
MET.

\textbf{April 5, 2023:} Initiated chemotherapy with Doxorubicin,
Cisplatin, Vincristine and Cyclophosphamide (ADOC). Patient in
sophisticated conditions, first therapy today.

\textbf{June 20, 2023:} Follow-up CT scan showed partial response with a
decrease in the size of the primary tumor and pulmonary nodules.
Continued chemotherapy regimen.

\textbf{November 15, 2023:} Follow-up imaging CT chest/abdomen: disease
progression with new metastatic lesions. Multiple hepatic lesions, with
the largest lesion in segment VIII measuring 4.5 cm, and another lesion
in segment II measuring 3.0 cm. Bone scan indicates metastatic
involvement of the spine, with lesions identified at T6, T9, and L3
vertebrae. Additional findings include new pulmonary nodules and further
enlargement of the primary mass in the anterior mediastinum. \colorbox{orange}{Signs of
known ILD, stable.}

\textbf{November 24, 2023:} Started second-line therapy with Carboplatin
plus Paclitaxel.~

\textbf{March 10, 2024:} Follow-up CT scan: PD. Progression of disease
with increased size of liver metastases and new bone lesions.

\textbf{May 20, 2024:} Tumor board review recommended considering
eligibility for clinical trials due to limited response to standard and
investigational therapies.

\textbf{June 1, 2024:} Detailed assessment of health status confirmed
adequate organ function. All routine labs within normal limits.

~
\begin{center}
\textbf{===== Patient 3.1.2 =====}~~
\end{center}
\textbf{Patient Information\\
}Name: Thomas Meyer\\
Born: January 12, 1966\\
Address: Schlossallee 1, Dresden, Germany

~

\textbf{Overview of Tumor Diagnosis and Therapy}

\textbf{Tumor Diagnosis\\
}Diagnosis: Masaoka-Koga Stage IVb thymic adenocarcinoma (metastases to
the lungs, liver and spines T6, T9, L3)\\
Initial Detection: March 15, 2023, following persistent chest pain and
cough, shortness-of-breath

Biopsy Date: March 28, 2023\\
Molecular Profile: Germline: BRCA2 p.K3326* (1N); Tumor: SMAD4 p.C363R
(1N), TP53 p.305fs (2N\_LOH), CDKN1B p.K100N (2N\_LOH), ATM p.E1666*
(4N), MAP3K8 p.H236Y (1N), TRAF1 p.R70H (2N), HDAC2 p.R409* (1N),
TMEM111-TDRD3 fusion, PRKDC-CDH17 fusion, EXT1-MAGI2 fusion;
Overexpressed genes: ERBB2, ERBB3, PDGFRB, TGFA, EGF, FGFR3, MET.

~

\textbf{Therapy Overview\\
}Initial Treatment:\\
Chemotherapy: Began April 5, 2023, with combination of Doxorubicin,
Cisplatin, Vincristine and Cyclophosphamide (ADOC). Partial response
after the initial two chemotherapy cycles completed by June, 2023.
Continued chemotherapy until November 2023 (progressive disease).\\
Subsequent Treatment: Second-line treatment with Carboplatin plus
Paclitaxel starting Nov 24, 2023. Current Status: Disease progression as
of May 2024, with new metastatic lesions in the lungs.

\textbf{Comorbidities}

\colorbox{orange}{Coronary Artery Disease (CAD), status post percutaneous coronary
intervention}

\begin{quote}
\colorbox{orange}{(PCI) with stent placement in 2018}

\colorbox{orange}{Interstitial Lung disease (ILD)}

Hypertension Stage 2

Type 2 Diabetes Mellitus

Hyperlipidemia

Gastroesophageal Reflux Disease (GERD)

H/o cholecystectomy 2011

Former Smoker 25 py
\end{quote}

~

\textbf{Medication}

\begin{quote}
\colorbox{orange}{Aspirin 100 1-0-0}\\
\colorbox{orange}{Clopidogrel 75mg 1-0-0}\\
\colorbox{orange}{Prednisone 10mg 1x}
\end{quote}
\begin{quote}
Losartan 50mg once daily

HCT 12.5mg once daily

Metformin 1000mg 1x/d

Atorvastatin 40mg once daily

Omeprazole 20mg daily

XGEVA q4w
\end{quote}

~

\textbf{Performance Status:} ECOG Performance Status 1

\textbf{Chronological Medical Findings:}

\textbf{March 15, 2023:} Presented with persistent chest pain and cough
and SOB.~

\textbf{March 20, 2023:} CT scan of the chest: Mass in the anterior
mediastinum measuring approximately 6.0 cm with evidence of local
invasion into surrounding structures. Multiple pulmonary nodules
suggestive of metastasis. \colorbox{orange}{Evidence of known Interstitial Lung Disease (ILD) with diffuse interstitial}\\
\colorbox{orange}{markings and fibrosis.}

\textbf{March 28, 2023:} CT-guided biopsy of mediastinal mass. Histology
confirmed thymic adenocarcinoma. Whole exome sequencing revealed
germline mutation BRCA2 p.K3326*, tumor mutations: SMAD4 p.C363R, TP53
p.305fs, CDKN1B p.K100N, ATM p.E1666*, MAP3K8 p.H236Y, TRAF1 p.R70H,
HDAC2 p.R409*, TMEM111-TDRD3 fusion, PRKDC-CDH17 fusion, EXT1-MAGI2
fusion. Overexpressed genes: ERBB2, ERBB3, PDGFRB, TGFA, EGF, FGFR3,
MET.

\textbf{April 5, 2023:} Initiated chemotherapy with Doxorubicin,
Cisplatin, Vincristine and Cyclophosphamide (ADOC). Patient in
sophisticated conditions, first therapy today.

\textbf{June 20, 2023:} Follow-up CT scan showed partial response with a
decrease in the size of the primary tumor and pulmonary nodules.
Continued chemotherapy regimen.

\textbf{November 15, 2023:} Follow-up imaging CT chest/abdomen: disease
progression with new metastatic lesions. Multiple hepatic lesions, with
the largest lesion in segment VIII measuring 4.5 cm, and another lesion
in segment II measuring 3.0 cm. Bone scan indicates metastatic
involvement of the spine, with lesions identified at T6, T9, and L3
vertebrae. Additional findings include new pulmonary nodules and further
enlargement of the primary mass in the anterior mediastinum. \colorbox{orange}{Signs of
known ILD, stable.}

\textbf{November 24, 2023:} Started second-line therapy with Carboplatin
plus Paclitaxel.~

\textbf{March 10, 2024:} Follow-up CT scan: PD. Progression of disease
with increased size of liver metastases and new bone lesions.

\textbf{May 20, 2024:} Tumor board review recommended considering
eligibility for clinical trials due to limited response to standard and
investigational therapies.

\textbf{June 1, 2024:} Detailed assessment of health status confirmed
adequate organ function. All routine labs within normal limits. Patient
claims newly intermittent chest pain -\textgreater{} next week
appointment at in-house cardiology department.

\begin{center}
\textbf{===== Patient 3.2 =====}~~
\end{center}

\textbf{Patient Information\\
}Name: Tim Müller\\
Born: January 03, 1966\\
Address: Parkallee 10, Dresden, Germany

\textbf{Overview of Tumor Diagnosis and Therapy}

\textbf{Tumor Diagnosis\\
}Diagnosis: Masaoka-Koga Stage IVb thymic adenocarcinoma (metastases to
the lungs, liver and spines T6, T9, L3)\\
Initial Detection: March 15, 2023, following persistent chest pain and
cough, shortness-of-breath

Biopsy Date: March 28, 2023\\
Molecular Profile: Germline: \colorbox{orange}{del BRCA2 mutation}; Tumor: SMAD4 p.C363R
(1N), TP53 p.305fs (2N\_LOH), CDKN1B p.K100N (2N\_LOH), ATM p.E1666*
(4N), MAP3K8 p.H236Y (1N), TRAF1 p.R70H (2N), HDAC2 p.R409* (1N),
TMEM111-TDRD3 fusion, PRKDC-CDH17 fusion, EXT1-MAGI2 fusion;
Overexpressed genes: ERBB2, ERBB3, PDGFRB, TGFA, EGF, FGFR3, MET.

~

\textbf{Therapy Overview\\
}Initial Treatment:\\
Chemotherapy: Began April 5, 2023, with combination of Doxorubicin,
Cisplatin, Vincristine and Cyclophosphamide (ADOC). Partial response
after the initial two chemotherapy cycles completed by June, 2023.
Continued chemotherapy until November 2023 (progressive disease).\\
Subsequent Treatment: Second-line treatment with Carboplatin plus
Paclitaxel starting Nov 24, 2023. Current Status: Disease progression as
of May 2024, with new metastatic lesions in the lungs.

\textbf{Comorbidities}

\begin{quote}
Former Smoker 25 py

Hypertension Stage 2

Type 2 Diabetes Mellitus

Hyperlipidemia

Gastroesophageal Reflux Disease (GERD)

H/o cholecystectomy 2011
\end{quote}

\textbf{Medication}

\begin{quote}
Losartan 50mg once daily

HCT 12.5mg once daily

Metformin 1000mg once daily

Atorvastatin 40mg once daily

Omeprazole 20mg once daily

XGEVA~
\end{quote}

~

\textbf{Performance Status:} ECOG Performance Status 1

\textbf{Chronological Medical Findings:}

\textbf{March 15, 2023:} Presented with persistent chest pain and cough
and SOB.~

\textbf{March 20, 2023:} CT scan of the chest: Mass in the anterior
mediastinum measuring approximately 6.0 cm with evidence of local
invasion into surrounding structures. Multiple pulmonary nodules
suggestive of metastasis.~

\textbf{March 28, 2023:} CT-guided biopsy of mediastinal mass. Histology
confirmed thymic adenocarcinoma. Whole exome sequencing revealed:
\colorbox{orange}{germline BRCA2 mutation (del)}, tumor mutations: SMAD4 p.C363R, TP53
p.305fs, CDKN1B p.K100N, ATM p.E1666*, MAP3K8 p.H236Y, TRAF1 p.R70H,
HDAC2 p.R409*, TMEM111-TDRD3 fusion, PRKDC-CDH17 fusion, EXT1-MAGI2
fusion. Overexpressed genes: ERBB2, ERBB3, PDGFRB, TGFA, EGF, FGFR3,
MET.

\textbf{April 5, 2023:} Initiated chemotherapy with Doxorubicin,
Cisplatin, Vincristine and Cyclophosphamide (ADOC). Patient in
sophisticated conditions, first therapy today.

\textbf{June 20, 2023:} Follow-up CT scan showed partial response with a
decrease in the size of the primary tumor and pulmonary nodules.
Continued chemotherapy regimen.

\textbf{November 15, 2023:} Follow-up imaging CT chest/abdomen: disease
progression with new metastatic lesions. Multiple hepatic lesions, with
the largest lesion in segment VIII measuring 4.5 cm, and another lesion
in segment II measuring 3.0 cm. Bone scan indicates metastatic
involvement of the spine, with lesions identified at T6, T9, and L3
vertebrae. Additional findings include new pulmonary nodules and further
enlargement of the primary mass in the anterior mediastinum.

\textbf{November 24, 2023:} Started second-line therapy with Carboplatin
plus Paclitaxel.~

\textbf{March 8, 2024:} End of chemotherapy.

\textbf{March 10, 2024:} Follow-up CT scan: PD. Progression of disease
with increased size of liver metastases and new bone lesions.

\textbf{March 20, 2024:} Tumor board review recommended considering
eligibility for clinical trials due to limited response to standard and
investigational therapies.

\textbf{April 5, 2024:} Detailed assessment of health status confirmed
adequate organ function. All routine labs within normal limits.

~
\begin{center}
\textbf{===== Patient 4.1 =====}~~~
\end{center}
\textbf{Patient Information}

Name: David Gärtner

Born: March 22, 1965

Address: Cologne, Domstrasse 1, Germany

~

\textbf{Overview of Tumor Diagnosis and Therapy}

\textbf{Tumor Diagnosis}

\textbf{Diagnosis}: UICC Stage IV oropharyngeal carcinoma (Metastatic:
LYM, HEP, OSS)

Initial Detection: February 10, 2023, following persistent sore throat
and difficulty swallowing

Biopsy Date: February 20, 2023

Molecular Profile: PIK3CA p.E545K (AF 25\%), MAPK1 p.E322K (AF 10\%),
FGFR3 p.D786N (AF 30\%)

~

\textbf{Therapy Overview}

Initial Treatment:

Radiochemotherapy: Began March 1, 2023, with a regimen of Cisplatin (200
mg/m2) paired with local radiotherapy (70G). Partial response noted
after the initial radiochemotherapy completed by June 15, 2023.
Follow-up CT scan shows disease progression in September 2024.

Subsequent Treatment: Began September 15, immunotherapy with Nivolumab
(240mg/2weeks).

~

\textbf{Current Status:} Disease progression as of March 2024, with new
metastatic lesions identified. ECOG 1.

~

\textbf{Comorbidities}

Active Smoker 50 py

Hypertension Stage 1

Hyperlipidemia

Peripheral artery disease Fontaine 2a

Diverticular disease CDD 3a

~

\textbf{Medication}

Lisinopril 20mg 1-0-0

Simvastatin 20mg 0-0-0-1

XGEVA, Vitamin D

~

\textbf{Chronological Medical Findings:}

\textbf{February 10, 2023:} Presented with persistent sore throat and
difficulty swallowing. CT scan of the neck revealed a suspicious mass in
the oropharynx.

\textbf{February 15, 2023:} CT scan of the neck: Mass in the oropharynx
measuring approximately 4.5 cm with evidence of local invasion into
surrounding structures. Multiple enlarged cervical lymph nodes are
noted, with the largest measuring approximately 2.2 cm in the right
level II region. These nodes exhibit round morphology and loss of the
fatty hilum, characteristics suggestive of metastatic involvement.
Additional enlarged lymph nodes are present in the levels III and IV on
the right side.

\textbf{February 18, 2023:} Staging CT (chest and abdomen): No signs of
distant metastasis.

\textbf{February 20, 2023:} Biopsy of the oropharyngeal mass performed.
Histology confirmed oropharyngeal carcinoma. Molecular panel sequencing
revealed mutations: PIK3CA p.E545K (AF 25\%), MAPK1 p.E322K (AF 10\%),
FGFR3 p.D786N (AF 30\%). Tumor purity was 60\%.

March 1, 2024: Initiated radiochemotherapy with Cisplatin and
5-Fluorouracil alongside local radiotherapy (70G).

\textbf{June 15, 2023:} Follow-up CT scan showed partial response with a
decrease in the size of the primary tumor and cervical lymph nodes.

\textbf{September 10, 2023:} Follow-up imaging CT Neck/Chest/Abdomen:
Disease progression (PD). Several new hypodense lesions identified in
the liver: The largest lesion located in segment VI, measuring
approximately 3.1 cm in diameter. Smaller lesions scattered throughout
the right~ hepatic lobe. Multiple new pulmonary nodules are detected in
both lungs. The largest nodule located in the right lower lobe,
measuring approximately 1.5 cm in diameter. Additional smaller nodules
are distributed throughout the bilateral lung fields. No evidence of
pleural effusion or pneumothorax. The oropharyngeal mass remains
present, with no significant change in size compared to the previous
scan. The previously noted enlarged cervical lymph nodes remain
prominent, with no significant interval change in size or number.

\textbf{September 15, 2023:} Began immunotherapy with Nivolumab.

\textbf{December 18, 2023:} Follow-up CT scan Neck/Chest/abdomen: Stable
disease.

\textbf{December-February 2023:} Continuation Nivolumab.

\textbf{February 20, 2024:} Follow-up CT scan Neck/Chest/Abdomen:
Progression of disease. Enlargement of multiple known hypodense lesions
in the liver, with the largest now measuring 4.5 cm in segment VI
(previously 3.1cm). New lytic lesions in the pelvis. Previously noted
pulmonary nodules remain stable with no significant interval change.
Stable primary tumor and cervical lymphadenopathy.

\textbf{March 3, 2024:} Tumor board review recommended considering
eligibility for clinical trials due to limited response to standard and
investigational therapies.

\textbf{March 10, 2024:} Detailed assessment of health status confirmed
adequate organ function. Routine labs within normal limits: ANC
4,500/mcL, platelet count 250,000/mcL, total bilirubin 0.8 mg/dL,
AST/ALT within normal limits, creatinine 0.8 mg/dL, hemoglobin 14.0
g/dL, serum albumin 4.0 g/dL, lipase and amylase within normal limits.
Patient in good clinical condition.

\begin{center}
\textbf{===== Patient 4.1.1 =====}~~~
\end{center}
\textbf{Patient Information}

Name: David Gärtner

Born: March 22, 1965

Address: Cologne, Domstrasse 1, Germany

~

\textbf{Overview of Tumor Diagnosis and Therapy}

\textbf{Tumor Diagnosis}

\textbf{Diagnosis}: UICC Stage IV oropharyngeal carcinoma (Metastatic:
LYM, HEP, OSS)

Initial Detection: February 10, 2023, following persistent sore throat
and difficulty swallowing

Biopsy Date: February 20, 2023

Molecular Profile: PIK3CA p.E545K (AF 25\%), MAPK1 p.E322K (AF 10\%),
FGFR3 p.D786N (AF 30\%)

~

\textbf{Therapy Overview}

Initial Treatment:

Radiochemotherapy: Began March 1, 2023, with a regimen of Cisplatin (200
mg/m2) paired with local radiotherapy (70G). Partial response noted
after the initial radiochemotherapy completed by June 15, 2023.
Follow-up CT scan shows disease progression in September 2024.

Subsequent Treatment: Began September 15, immunotherapy with Nivolumab
(240mg/2weeks).

~

\textbf{Current Status:} Disease progression as of March 2024, with new
metastatic lesions identified. ECOG 1.

~

\textbf{Comorbidities}

Active Smoker 50 py

Hypertension Stage 1

Hyperlipidemia

Peripheral artery disease Fontaine 2a

Diverticular disease CDD 3a

~

\textbf{Medication}

Lisinopril 20mg 1-0-0

Simvastatin 20mg 0-0-0-1

XGEVA, Vitamin D

~

\textbf{Chronological Medical Findings:}

\textbf{February 10, 2023:} Presented with persistent sore throat and
difficulty swallowing. CT scan of the neck revealed a suspicious mass in
the oropharynx.

\textbf{February 15, 2023:} CT scan of the neck: Mass in the oropharynx
measuring approximately 4.5 cm with evidence of local invasion into
surrounding structures. Multiple enlarged cervical lymph nodes are
noted, with the largest measuring approximately 2.2 cm in the right
level II region. These nodes exhibit round morphology and loss of the
fatty hilum, characteristics suggestive of metastatic involvement.
Additional enlarged lymph nodes are present in the levels III and IV on
the right side.

\textbf{February 18, 2023:} Staging CT (chest and abdomen): No signs of
distant metastasis.

\textbf{February 20, 2023:} Biopsy of the oropharyngeal mass performed.
Histology confirmed oropharyngeal carcinoma. Molecular panel sequencing
revealed mutations: PIK3CA p.E545K (AF 25\%), MAPK1 p.E322K (AF 10\%),
FGFR3 p.D786N (AF 30\%). Tumor purity was 60\%.

March 1, 2024: Initiated radiochemotherapy with Cisplatin and
5-Fluorouracil alongside local radiotherapy (70G).

\textbf{June 15, 2023:} Follow-up CT scan showed partial response with a
decrease in the size of the primary tumor and cervical lymph nodes.

\textbf{September 10, 2023:} Follow-up imaging CT Neck/Chest/Abdomen:
Disease progression (PD). Several new hypodense lesions identified in
the liver: The largest lesion located in segment VI, measuring
approximately 3.1 cm in diameter. Smaller lesions scattered throughout
the right~ hepatic lobe. Multiple new pulmonary nodules are detected in
both lungs. The largest nodule located in the right lower lobe,
measuring approximately 1.5 cm in diameter. Additional smaller nodules
are distributed throughout the bilateral lung fields. No evidence of
pleural effusion or pneumothorax. The oropharyngeal mass remains
present, with no significant change in size compared to the previous
scan. The previously noted enlarged cervical lymph nodes remain
prominent, with no significant interval change in size or number.

\textbf{September 15, 2023:} Began immunotherapy with Nivolumab.

\textbf{December 18, 2023:} Follow-up CT scan Neck/Chest/abdomen: Stable
disease.

\textbf{December-February 2023:} Continuation Nivolumab.

\textbf{February 20, 2024:} Follow-up CT scan Neck/Chest/Abdomen:
Progression of disease. Enlargement of multiple known hypodense lesions
in the liver, with the largest now measuring 4.5 cm in segment VI
(previously 3.1cm). New lytic lesions in the pelvis. Previously noted
pulmonary nodules remain stable with no significant interval change.
Stable primary tumor and cervical lymphadenopathy.

\textbf{March 3, 2024:} Tumor board review recommended considering
eligibility for clinical trials due to limited response to standard and
investigational therapies.

\colorbox{orange}{\textbf{March 10, 2024:} Visit in the Emergency department with fever for 3 days, }\\
\colorbox{orange}{shortness of breath + cough + severe headaches.}\\
\colorbox{orange}{Routine labs: ANC 15,000/mcL, platelet count 200,000/mcL,}\\
\colorbox{orange}{total bilirubin 1.2mg/dL, AST/ALT 1.5 x ULN, creatinine 1.1 mg/dL,}\\
\colorbox{orange}{hemoglobin 12.0 g/dL, serum albumin 3.5 g/dL, leukocytes 18,000/mcL,}\\ 
\colorbox{orange}{CRP 23 mg/dL.}\\ 
\colorbox{orange}{Chest X-ray and CT scan confirmed pneumonia.}\\
\colorbox{orange}{Hospitalized for further evaluation and treatment.}\\
\colorbox{orange}{Blood and sputum cultures were taken and are pending.}\\
\colorbox{orange}{Patient started on IV antibiotics: Ceftriaxone and Azithromycin.}

~
\begin{center}
\textbf{===== Patient 4.1.2 =====}~~~
\end{center}
\textbf{Patient Information}

Name: David Gärtner

Born: March 22, 1965

Address: Cologne, Domstrasse 1, Germany

~

\textbf{Overview of Tumor Diagnosis and Therapy}

\textbf{Tumor Diagnosis}

\textbf{Diagnosis}: UICC Stage IV oropharyngeal carcinoma (Metastatic:
LYM, HEP, OSS)

Initial Detection: February 10, 2023, following persistent sore throat
and difficulty swallowing

Biopsy Date: February 20, 2023

Molecular Profile: PIK3CA p.E545K (AF 25\%), MAPK1 p.E322K (AF 10\%),
FGFR3 p.D786N (AF 30\%)

~

\textbf{Therapy Overview}

Initial Treatment:

Radiochemotherapy: Began March 1, 2023, with a regimen of Cisplatin (200
mg/m2) paired with local radiotherapy (70G). Partial response noted
after the initial radiochemotherapy completed by June 15, 2023.
Follow-up CT scan shows disease progression in September 2024.

Subsequent Treatment: Began September 15, immunotherapy with Nivolumab
(240mg/2weeks).

~

\textbf{Current Status:} Disease progression as of March 2024, with new
metastatic lesions identified. ECOG 1.

~

\textbf{Comorbidities}

Active Smoker 50 py

Hypertension Stage 1

Hyperlipidemia

Peripheral artery disease Fontaine 2a

Diverticular disease CDD 3a

\colorbox{orange}{Epilepsy, Focal Onset Impaired Awareness Seizures}\\
\colorbox{orange}{NYHA Class II Heart Failure}\\
\colorbox{orange}{COPD, GOLD Stage 3 (Severe)}\\
~

\textbf{Medication}

\colorbox{orange}{Levetiracetam 500mg 1-0-1 (for epilepsy)}

\colorbox{orange}{Metoprolol Succinate 50mg 1-0-0 (for heart failure)}

\colorbox{orange}{Tiotropium 18mcg 1-0-0 (for COPD)}

\colorbox{orange}{Salbutamol Inhaler 100mcg as needed (for COPD)}

Lisinopril 20mg 1-0-0

Simvastatin 20mg 0-0-0-1

XGEVA, Vitamin D

~

\textbf{Chronological Medical Findings:}

\textbf{February 10, 2023:} Presented with persistent sore throat and
difficulty swallowing. CT scan of the neck revealed a suspicious mass in
the oropharynx.

\textbf{February 15, 2023:} CT scan of the neck: Mass in the oropharynx
measuring approximately 4.5 cm with evidence of local invasion into
surrounding structures. Multiple enlarged cervical lymph nodes are
noted, with the largest measuring approximately 2.2 cm in the right
level II region. These nodes exhibit round morphology and loss of the
fatty hilum, characteristics suggestive of metastatic involvement.
Additional enlarged lymph nodes are present in the levels III and IV on
the right side.

\textbf{February 18, 2023:} Staging CT (chest and abdomen): No signs of
distant metastasis.

\textbf{February 20, 2023:} Biopsy of the oropharyngeal mass performed.
Histology confirmed oropharyngeal carcinoma. Molecular panel sequencing
revealed mutations: PIK3CA p.E545K (AF 25\%), MAPK1 p.E322K (AF 10\%),
FGFR3 p.D786N (AF 30\%). Tumor purity was 60\%.

March 1, 2024: Initiated radiochemotherapy with Cisplatin and
5-Fluorouracil alongside local radiotherapy (70G).

\textbf{June 15, 2023:} Follow-up CT scan showed partial response with a
decrease in the size of the primary tumor and cervical lymph nodes.

\textbf{September 10, 2023:} Follow-up imaging CT Neck/Chest/Abdomen:
Disease progression (PD). Several new hypodense lesions identified in
the liver: The largest lesion located in segment VI, measuring
approximately 3.1 cm in diameter. Smaller lesions scattered throughout
the right~ hepatic lobe. Multiple new pulmonary nodules are detected in
both lungs. The largest nodule located in the right lower lobe,
measuring approximately 1.5 cm in diameter. Additional smaller nodules
are distributed throughout the bilateral lung fields. No evidence of
pleural effusion or pneumothorax. The oropharyngeal mass remains
present, with no significant change in size compared to the previous
scan. The previously noted enlarged cervical lymph nodes remain
prominent, with no significant interval change in size or number.

\textbf{September 15, 2023:} Began immunotherapy with Nivolumab.

\textbf{December 18, 2023:} Follow-up CT scan Neck/Chest/abdomen: Stable
disease.

\textbf{December-February 2023:} Continuation Nivolumab.

\textbf{February 20, 2024:} Follow-up CT scan Neck/Chest/Abdomen:
Progression of disease. Enlargement of multiple known hypodense lesions
in the liver, with the largest now measuring 4.5 cm in segment VI
(previously 3.1cm). New lytic lesions in the pelvis. Previously noted
pulmonary nodules remain stable with no significant interval change.
Stable primary tumor and cervical lymphadenopathy.

\textbf{March 3, 2024:} Tumor board review recommended considering
eligibility for clinical trials due to limited response to standard and
investigational therapies.

\textbf{March 10, 2024:} Detailed assessment of health status confirmed
adequate organ function. Routine labs within normal limits: ANC
4,500/mcL, platelet count 250,000/mcL, total bilirubin 0.8 mg/dL,
AST/ALT within normal limits, creatinine 0.8 mg/dL, hemoglobin 14.0
g/dL, serum albumin 4.0 g/dL, lipase and amylase within normal limits.
Patient in good clinical condition.

~
\begin{center}
\textbf{===== Patient 4.1.3 =====}~~~
\end{center}
\textbf{Patient Information}

Name: David Gärtner

Born: March 22, 1965

Address: Cologne, Domstrasse 1, Germany

~

\textbf{Overview of Tumor Diagnosis and Therapy}

\textbf{Tumor Diagnosis}

\textbf{Diagnosis}: UICC Stage IV oropharyngeal carcinoma (Metastatic:
LYM, HEP, OSS)

Initial Detection: February 10, 2023, following persistent sore throat
and difficulty swallowing

Biopsy Date: February 20, 2023

Molecular Profile: PIK3CA p.E545K (AF 25\%), MAPK1 p.E322K (AF 10\%),
FGFR3 p.D786N (AF 30\%)

~

\textbf{Therapy Overview}

Initial Treatment:

Radiochemotherapy: Began March 1, 2023, with a regimen of Cisplatin (200
mg/m2) paired with local radiotherapy (70G). Partial response noted
after the initial radiochemotherapy completed by June 15, 2023.
Follow-up CT scan shows disease progression in September 2024.

Subsequent Treatment: Began September 15, immunotherapy with Nivolumab
(240mg/2weeks).

~

\textbf{Current Status:} Disease progression as of March 2024, with new
metastatic lesions identified. ECOG 1.

~

\textbf{Comorbidities}

\colorbox{orange}{Active hepatitis C virus (HCV) infection (HCV antibody +),}\\
\colorbox{orange}{HCV RNA elevated (March 10, 2024)}

Active Smoker 50 py

\colorbox{orange}{Alcoholic (6 bottles of beer / day)

Regular marijuana use (up to 10 joints per day)}

~

Hypertension Stage 1

Hyperlipidemia

Peripheral artery disease Fontaine 2a

Diverticular disease CDD 3a

~

\textbf{Medication}

\colorbox{orange}{Sofosbuvir/Velpatasvir 400mg/100mg 1-0-0}

Lisinopril 20mg 1-0-0

Simvastatin 20mg 0-0-0-1

XGEVA, Vitamin D

~

\textbf{Chronological Medical Findings:}

\textbf{February 10, 2023:} Presented with persistent sore throat and
difficulty swallowing. CT scan of the neck revealed a suspicious mass in
the oropharynx.

\textbf{February 15, 2023:} CT scan of the neck: Mass in the oropharynx
measuring approximately 4.5 cm with evidence of local invasion into
surrounding structures. Multiple enlarged cervical lymph nodes are
noted, with the largest measuring approximately 2.2 cm in the right
level II region. These nodes exhibit round morphology and loss of the
fatty hilum, characteristics suggestive of metastatic involvement.
Additional enlarged lymph nodes are present in the levels III and IV on
the right side.

\textbf{February 18, 2023:} Staging CT (chest and abdomen): No signs of
distant metastasis.

\textbf{February 20, 2023:} Biopsy of the oropharyngeal mass performed.
Histology confirmed oropharyngeal carcinoma. Molecular panel sequencing
revealed mutations: PIK3CA p.E545K (AF 25\%), MAPK1 p.E322K (AF 10\%),
FGFR3 p.D786N (AF 30\%). Tumor purity was 60\%.

March 1, 2024: Initiated radiochemotherapy with Cisplatin and
5-Fluorouracil alongside local radiotherapy (70G).

\textbf{June 15, 2023:} Follow-up CT scan showed partial response with a
decrease in the size of the primary tumor and cervical lymph nodes.

\textbf{September 10, 2023:} Follow-up imaging CT Neck/Chest/Abdomen:
Disease progression (PD). Several new hypodense lesions identified in
the liver: The largest lesion located in segment VI, measuring
approximately 3.1 cm in diameter. Smaller lesions scattered throughout
the right~ hepatic lobe. Multiple new pulmonary nodules are detected in
both lungs. The largest nodule located in the right lower lobe,
measuring approximately 1.5 cm in diameter. Additional smaller nodules
are distributed throughout the bilateral lung fields. No evidence of
pleural effusion or pneumothorax. The oropharyngeal mass remains
present, with no significant change in size compared to the previous
scan. The previously noted enlarged cervical lymph nodes remain
prominent, with no significant interval change in size or number.

\textbf{September 15, 2023:} Began immunotherapy with Nivolumab.

\textbf{December 18, 2023:} Follow-up CT scan Neck/Chest/abdomen: Stable
disease.

\textbf{December-February 2023:} Continuation Nivolumab.

\textbf{February 20, 2024:} Follow-up CT scan Neck/Chest/Abdomen:
Progression of disease. Enlargement of multiple known hypodense lesions
in the liver, with the largest now measuring 4.5 cm in segment VI
(previously 3.1cm). New lytic lesions in the pelvis. Previously noted
pulmonary nodules remain stable with no significant interval change.
Stable primary tumor and cervical lymphadenopathy.

\textbf{March 3, 2024:} Tumor board review recommended considering
eligibility for clinical trials due to limited response to standard and
investigational therapies.

\textbf{March 10, 2024:} Detailed assessment of health status confirmed
adequate organ function. Routine labs within normal limits: ANC
4,500/mcL, platelet count 250,000/mcL, total bilirubin 0.8 mg/dL,
AST/ALT within normal limits, creatinine 0.8 mg/dL, hemoglobin 14.0
g/dL, serum albumin 4.0 g/dL, lipase and amylase within normal limits.
\colorbox{orange}{HCV RNA at 2,500,000 IU/mL.} Patient in good clinical condition.

~
\begin{center}
\textbf{===== Patient 5.1 =====}~~~~
\end{center}
\textbf{Patient Information}

Name: Lisa Müller\\
Born: April 12, 1960\\
Address: Hamburg, Hafenstrasse 3, Germany

~

\textbf{Overview of Tumor Diagnosis and Therapy}

\textbf{Tumor Diagnosis\\
}Diagnosis: Stage IV lung adenocarcinoma (M+: HEP, LYM, BONE, ADRENAL)\\
Initial Detection: January November 21, 2023, following persistent cough
and weight loss\\
Biopsy Date: November 28, 2023

Molecular Profile: EGFR p.E746\_A750del (AF 43\%), TP53 p.A138\_Q144del
(AF 37\%), MET Amplification FISH positive; Tumor Purity: 30\%; Tumor
Mutational Burden (TMB): 3.8 Mut/MB

~

\textbf{Therapy Overview\\
}Initial Treatment:\\
Chemotherapy: Began February December 2023, with a regimen of
Pembrolizumab, Carboplatin and Pemetrexed. Partial response after the
initial chemotherapy cycle completed by May 1, 2024. Continued
chemotherapy until May 2024 (progressive disease).

\textbf{Comorbidities}

\begin{quote}
Hyperlipidemia~

Osteoarthritis

Psoriasis vulgaris

H/o cholecystectomy 2007

45py
\end{quote}

~

\textbf{Medication}

\begin{quote}
Atorvastatin 40mg once daily

Hydrocortisone cream~

Ibuprofen 400mg as needed

XGEVA

Novalgin 500 2-2-2-2
\end{quote}

~

\textbf{Performance Status:} ECOG Performance Status 1

~

\textbf{Chronological Medical Findings:}

\textbf{November 8, 2023:} Presented with persistent cough and weight
loss (-6kg/3 mo) at her primary care physician. 1 week antibiotic
treatment for suspected airway infection without clinical improvements,
chest x-ray revealed tumorous lesion in the left lung.

\textbf{November 21, 2023:} CT scan of the chest: Mass in the left upper
lobe measuring approximately 5.0 cm with evidence of local invasion into
surrounding structures, including the left main bronchus and adjacent
vascular structures. Enlarged mediastinal lymph nodes, particularly in
the subcarinal and right paratracheal regions, with the largest node
measuring 1.8 cm. Additional moderate pleural effusion on the left side.
Multiple liver lesions suggestive of metastasis, with the largest lesion
in segment VIII measuring 3.5 cm and another lesion in segment IVa
measuring 2.2 cm. Adrenal metastasis on the left side. Bone metastases
in C3, T3,4,7.

\textbf{November 28, 2023:} CT guided tumor biopsy: Histology confirmed
lung adenocarcinoma. Molecular panel sequencing revealed mutations: EGFR
p.E746\_A750del (AF 43\%), TP53 p.A138\_Q144del (AF 37\%), MET
Amplification FISH positive. Tumor purity was 30\%. Tumor Mutational
Burden (TMB) was 3.8 Mut/MB.

\textbf{December 5, 2023:} Initiated chemotherapy with Carboplatin and
Pemetrexed + immunotherapy with Pembrolizumab.

\textbf{March 10, 2024:} Follow-up CT scan: Partial Response. Continued
chemotherapy regimen.

\textbf{March - May 2024:} Continued therapy with Carbo/Pem +
Pembrolizumab.

\textbf{May 07, 2024:} CT-scan Chest/Abdomen: Significant disease
progression. Mass in the left upper lobe has increased to approximately
6.5 cm with further invasion into the left main bronchus and adjacent
vascular structures. Enlarged mediastinal lymph nodes are now more
prominent, especially in the subcarinal and right paratracheal regions,
with the largest node now measuring 2.5 cm. Progressive pleural effusion
left \textgreater{} right.

Abdomen and Pelvis: Multiple liver lesions, with the largest in segment
VIII now measuring 4.5 cm and another in segment IVa measuring 3.0 cm.
New metastatic lesions observed in segments V and VI. The adrenal
metastasis on the left side has increased in size to 2.5 cm.

Bone Metastases: Increased metastatic involvement with new lesions
identified in the spine, including C2, T5, and L1, in addition to the
previously noted C3, T3, T4, and T7.

\textbf{May 10, 2024:} Tumor board review recommended considering
eligibility for clinical trials due to limited response to standard and
investigational therapies.

\textbf{May 15, 2024:} Detailed assessment of health status confirmed
adequate organ function. Routine labs within normal limits: ANC
4,800/mcL, platelet count 220,000/mcL, total bilirubin 0.9 mg/dL,
AST/ALT within normal limits, creatinine 0.9 mg/dL, hemoglobin 13.5
g/dL, serum albumin 4.2 g/dL, lipase and amylase within normal limits.

~

\begin{center}
\textbf{===== Patient 5.1.1 =====}~~~~
\end{center}
\textbf{Patient Information}

Name: Lisa Müller\\
Born: April 12, 1960\\
Address: Hamburg, Hafenstrasse 3, Germany

~

\textbf{Overview of Tumor Diagnosis and Therapy}

\textbf{Tumor Diagnosis\\
}Diagnosis: Stage IV lung adenocarcinoma (M+: HEP, LYM, BONE, ADRENAL)\\
Initial Detection: January November 21, 2023, following persistent cough
and weight loss\\
Biopsy Date: November 28, 2023

Molecular Profile: EGFR p.E746\_A750del (AF 43\%), TP53 p.A138\_Q144del
(AF 37\%), MET Amplification FISH positive; Tumor Purity: 30\%; Tumor
Mutational Burden (TMB): 3.8 Mut/MB

~

\textbf{Therapy Overview\\
}Initial Treatment:\\
Chemotherapy: Began February December 2023, with a regimen of
Pembrolizumab, Carboplatin and Pemetrexed. Partial response after the
initial chemotherapy cycle completed by May 1, 2024. Continued
chemotherapy until May 2024 (progressive disease).

\textbf{Comorbidities}

\begin{quote}
Hyperlipidemia~

Osteoarthritis

Psoriasis vulgaris

H/o cholecystectomy 2007

45py
\end{quote}

~

\textbf{Medication}

\begin{quote}
Atorvastatin 40mg once daily

Hydrocortisone cream~

Ibuprofen 400mg as needed

XGEVA

Novalgin 500 2-2-2-2

\colorbox{orange}{Prednisone 40mg daily}
\end{quote}

~

\textbf{Performance Status:} \colorbox{orange}{ECOG Performance Status 2}

~

\textbf{Chronological Medical Findings:}

\textbf{November 8, 2023:} Presented with persistent cough and weight
loss (-6kg/3 mo) at her primary care physician. 1 week antibiotic
treatment for suspected airway infection without clinical improvements,
chest x-ray revealed tumorous lesion in the left lung.

\textbf{November 21, 2023:} CT scan of the chest: Mass in the left upper
lobe measuring approximately 5.0 cm with evidence of local invasion into
surrounding structures, including the left main bronchus and adjacent
vascular structures. Enlarged mediastinal lymph nodes, particularly in
the subcarinal and right paratracheal regions, with the largest node
measuring 1.8 cm. Additional moderate pleural effusion on the left side.
Multiple liver lesions suggestive of metastasis, with the largest lesion
in segment VIII measuring 3.5 cm and another lesion in segment IVa
measuring 2.2 cm. Adrenal metastasis on the left side. Bone metastases
in C3, T3,4,7.

\textbf{November 28, 2023:} CT guided tumor biopsy: Histology confirmed
lung adenocarcinoma. Molecular panel sequencing revealed mutations: EGFR
p.E746\_A750del (AF 43\%), TP53 p.A138\_Q144del (AF 37\%), MET
Amplification FISH positive. Tumor purity was 30\%. Tumor Mutational
Burden (TMB) was 3.8 Mut/MB.

\textbf{December 5, 2023:} Initiated chemotherapy with Carboplatin and
Pemetrexed + immunotherapy with Pembrolizumab.

\textbf{March 10, 2024:} Follow-up CT scan: Partial Response. Continued
chemotherapy regimen.

\textbf{March - May 2024:} Continued therapy with Carbo/Pem +
Pembrolizumab.

\textbf{May 07, 2024:} CT-scan Chest/Abdomen: Significant disease
progression. Mass in the left upper lobe has increased to approximately
6.5 cm with further invasion into the left main bronchus and adjacent
vascular structures. Enlarged mediastinal lymph nodes are now more
prominent, especially in the subcarinal and right paratracheal regions,
with the largest node now measuring 2.5 cm. Progressive pleural effusion
left \textgreater{} right.
\colorbox{orange}{Additionally, there are diffuse ground-glass
opacities and reticular}\\
\colorbox{orange}{markings throughout both lungs, suspicious for immune mediated Pneumonitis.}

Abdomen and Pelvis: Multiple liver lesions, with the largest in segment
VIII now measuring 4.5 cm and another in segment IVa measuring 3.0 cm.
New metastatic lesions observed in segments V and VI. The adrenal
metastasis on the left side has increased in size to 2.5 cm.

Bone Metastases: Increased metastatic involvement with new lesions
identified in the spine, including C2, T5, and L1, in addition to the
previously noted C3, T3, T4, and T7.

\colorbox{orange}{\textbf{May 08, 2024:} Started on Prednisone 40 mg daily because of Lung findings.}\\
\colorbox{orange}{Follow-up CT scan and pulmonary function tests scheduled.}\\
\colorbox{orange}{Patient advised on potential side effects and the need for regular
monitoring.}

\textbf{May 10, 2024:} Tumor board review recommended considering
eligibility for clinical trials due to limited response to standard and
investigational therapies.

\textbf{May 15, 2024:} Detailed assessment of health status confirmed
adequate organ function. Routine labs within normal limits: ANC
4,800/mcL, platelet count 220,000/mcL, total bilirubin 0.9 mg/dL,
AST/ALT within normal limits, creatinine 0.9 mg/dL, hemoglobin 13.5
g/dL, serum albumin 4.2 g/dL, lipase and amylase within normal limits.

~
\begin{center}
\textbf{===== Patient 6.1 =====}~~~~~
\end{center}
\textbf{Patient Information}

Name: Ehrich, Wolfgang

born: 18.08.1968

Address: Kurfürstendamm 1, Berlin, Germany

~

\textbf{Overview of Tumor Diagnosis and Therapy}

\textbf{Tumor Diagnosis}

\begin{quote}
Diagnosis: UICC Stage IVA, M1a (contralateral metastases, malignant
pleural effusions),~

KRAS G12C mutant non-small cell lung cancer (NSCLC)
\end{quote}

Initial Detection: March 22, 2023, following symptoms of persistent
cough and weight loss

Biopsy Date: April 15, 2023, squamous cell lung cancer, PDL1 3\%

\begin{quote}
Molecular Profile: Molecular alterations: KRAS p.G12C (AF 18\%), KRAS
p.G12C (AF~

18\%), KEAP1 p.L276F (AF 45\%), STK11 p.K83Tfs*13 (AF 38\%).
\end{quote}

~

\textbf{Therapy Overview}

Combined Immuno-chemotherapy: Began May 5, 2023, with Cisplatin,

\begin{quote}
Pemetrexed and Pembrolizumab, partial response noted after cycle
completion by August 10, 2023, continuation of therapy until october
2023 (progressive disease)
\end{quote}

\textbf{Current Status}

\begin{quote}
Health Condition: Stable with an ECOG performance status of 1
\end{quote}

~

\textbf{Comorbidities}

\begin{quote}
Former Smoker: 40 py

Hypertension Stage I

COPD GOLD 2

Type 2 Diabetes Mellitus

Hyperlipidemia
\end{quote}

~

\textbf{Medication}

~~~~~~~~ Lisinopril 20mg 1-0-0

~~~~~~~~ Metformin 1000mg 1-0-1

~~~~~~~~ Atorvastatin 40mg 0-0-0-1

~~~~~~~~ Tiotropium (Inhaler) on demand

~~~~~~~~ Novalgin 500mg 1--1-1-1~

Apixaban 5mg 1-0-1

\begin{quote}
~~
\end{quote}

\textbf{Chronological Medical Findings:}

\begin{quote}
March 22, 2023: Experienced persistent cough and weight loss. Chest
X-ray and CT scan revealed a mass in the left lung. Referred to
oncologist. CT-Angiography: Tumor Size: Approximately 4.5 cm in
diameter. At least 2 contralateral metastases. Bronchial Obstruction:
Partial obstruction of the left main bronchus leading to atelectasis of
the left upper lobe. Suspicion of mediastinal lymph node metastases. No
evidence of pulmonary artery embolism. Thrombus in the left atrium at
the transition to the auricle. Emphysematous and fibrotic changes in the
lung parenchyma. Urgent suspicion of a tumor-atelectasis complex in the
left upper lobe of the lung. Mucus present in the lower lobe bronchi on
the left. Lymph Nodes: Enlarged, FDG-positive lymph nodes in the
mediastinum, particularly in regions 4R and infracarinal.

April 15, 2023: Lung biopsy via bronchoscopy: Endobronchial tumor
manifestation in the distal left main bronchus extending to the upper
lobe. Acute and chronic atrophic tracheobronchitis. Collapsed bronchial
system in the affected area. Biopsy taken. Diagnosed with squamous
non-small cell lung cancer (NSCLC), molecular diagnostics: KRAS G12C
mutant.

April 27, 2023: Ventilation: Moderate obstruction, no restriction.
Increased airway resistance and slight hyperinflation. Tiffeneau index
(FEV1/FVC) at 42.34\%, z-score -3.32. FEV1: 0.93 L (42\% predicted),
z-score -2.89.Total lung capacity (TLC): 5.86 L (103\% predicted),
z-score 0.22. Forced vital capacity (FVC): 2.19 L, z-score -1.5.
Residual volume (RV): 3.67 L, z-score 2.44. RV/TLC: 62.68\%, z-score
-1.18.

May 5, 2023: Initiated on platinum-based immunochemotherapy regimen
(Cisplatin, Pemetrexed, Pembrolizumab).

August 10, 2023: Completed initial therapy cycle. Partial response as
per CT chest / abdomen +PET CT:~ Moderate reduction in tumor size to
approximately 4.2 cm. Contralateral metastases still present, but no new
lesions. Partial bronchial obstruction persists with ongoing atelectasis
in the left upper lobe. Mediastinal lymph nodes remain enlarged and
FDG-positive, although with reduced metabolic activity. Thrombus in the
left atrium remains unchanged. Emphysematous and fibrotic changes are
stable. Overall, mild response observed with no significant progression,
as per RECIST stable disease.

August-October: Continued chemotherapy with Cisplatin/Pemetrexed and
Pembrolizumab.

October 13, 2023: Follow-up CT (chest + abdomen): SD / Progressive
Disease. New nodule in the right lung (1cm). Slight increase in the size
of previously noted FDG-positive lymph nodes in the mediastinum. No
additional metastatic lesions were detected. The patient has continued
to tolerate the current treatment regimen well, with no significant
adverse effects reported.

October 17, 2023: Tumor board: SD. Continuation of therapy.

October 25, 2023: Continuation of Cisplatin (dose reduced), Pemetrexed
and Pembrolizumab.

January 12, 2024: Follow-up CT scan abdomen and chest, FDG-PET-CT:
Progressive Disease with three new metastases in the right lung and
additional enlarged FDG-positive lymph nodes in the mediastinum. MRI
scan of the brain conducted; no evidence of metastatic disease.
Incidental findings included mild age-related cerebral atrophy and
scattered white matter hyperintensities consistent with chronic
microvascular ischemic changes.

March 17, 2024: Tumorboard recommends considering clinical trial options
due to limited response to standard therapies.

April 20, 2024: Detailed assessment of health status. ECOG performance
status 1. All routine labs, including liver and renal function tests,
within normal limits.
\end{quote}

~

\begin{center}
\textbf{===== Patient 6.1.1 =====}~~~~~
\end{center}
\textbf{Patient Information}

Name: Ehrich, Wolfgang

born: 18.08.1968

Address: Kurfürstendamm 1, Berlin, Germany

~

\textbf{Overview of Tumor Diagnosis and Therapy}

\textbf{Tumor Diagnosis}

\begin{quote}
Diagnosis: UICC Stage IVA, M1a (contralateral metastases, malignant
pleural effusions),~

KRAS G12C mutant non-small cell lung cancer (NSCLC)
\end{quote}

Initial Detection: March 22, 2023, following symptoms of persistent
cough and weight loss

Biopsy Date: April 15, 2023, squamous cell lung cancer, PDL1 3\%

\begin{quote}
Molecular Profile: Molecular alterations: KRAS p.G12C (AF 18\%), KRAS
p.G12C (AF~

18\%), KEAP1 p.L276F (AF 45\%), STK11 p.K83Tfs*13 (AF 38\%).
\end{quote}

~

\textbf{Therapy Overview}

Combined Immuno-chemotherapy: Began May 5, 2023, with Cisplatin,

\begin{quote}
Pemetrexed and Pembrolizumab, partial response noted after cycle
completion by August 10, 2023, continuation of therapy until october
2023 (progressive disease)
\end{quote}

\textbf{Current Status}

\begin{quote}
Health Condition: Stable with an ECOG performance status of 1
\end{quote}

~

\textbf{Comorbidities}

\begin{quote}
\colorbox{orange}{Chronic heart failure (NYHA Class III),}\\
\colorbox{orange}{reduced ejection fraction (HFrEF) of 35\%}

\colorbox{orange}{Post Myocardial Infarction (2021), 2 coronary stents}

Former Smoker: 40 py

Hypertension Stage I

COPD GOLD 2

Type 2 Diabetes Mellitus

Hyperlipidemia
\end{quote}

~

\textbf{Medication}

~~~~~~~~ Lisinopril 20mg 1-0-0

~~~~~~~~ Metformin 1000mg 1-0-1

\colorbox{orange}{ASS 100mg 1-0-0}\\

\colorbox{orange}{Carvedilol 12.5mg 1-0-1}\\

\colorbox{orange}{Furosemide 40mg 1-0-1}\\

\colorbox{orange}{Apixaban 5mg 1-0-1}\\

Atorvastatin 40mg 0-0-0-1

~~~~~~~~ Tiotropium (Inhaler) on demand

Novalgin 500mg 1-1-1-1~

\begin{quote}
~~
\end{quote}

\textbf{Chronological Medical Findings:}

\begin{quote}
March 22, 2023: Experienced persistent cough and weight loss. Chest
X-ray and CT scan revealed a mass in the left lung. Referred to
oncologist. CT-Angiography: Tumor Size: Approximately 4.5 cm in
diameter. At least 2 contralateral metastases. Bronchial Obstruction:
Partial obstruction of the left main bronchus leading to atelectasis of
the left upper lobe. Suspicion of mediastinal lymph node metastases. No
evidence of pulmonary artery embolism. Thrombus in the left atrium at
the transition to the auricle. Emphysematous and fibrotic changes in the
lung parenchyma. Urgent suspicion of a tumor-atelectasis complex in the
left upper lobe of the lung. Mucus present in the lower lobe bronchi on
the left. Lymph Nodes: Enlarged, FDG-positive lymph nodes in the
mediastinum, particularly in regions 4R and infracarinal.

April 15, 2023: Lung biopsy via bronchoscopy: Endobronchial tumor
manifestation in the distal left main bronchus extending to the upper
lobe. Acute and chronic atrophic tracheobronchitis. Collapsed bronchial
system in the affected area. Biopsy taken. Diagnosed with squamous
non-small cell lung cancer (NSCLC), molecular diagnostics: KRAS G12C
mutant.

April 27, 2023: Ventilation: Moderate obstruction, no restriction.
Increased airway resistance and slight hyperinflation. Tiffeneau index
(FEV1/FVC) at 42.34\%, z-score -3.32. FEV1: 0.93 L (42\% predicted),
z-score -2.89.Total lung capacity (TLC): 5.86 L (103\% predicted),
z-score 0.22. Forced vital capacity (FVC): 2.19 L, z-score -1.5.
Residual volume (RV): 3.67 L, z-score 2.44. RV/TLC: 62.68\%, z-score
-1.18.

May 5, 2023: Initiated on platinum-based immunochemotherapy regimen
(Cisplatin, Pemetrexed, Pembrolizumab).

August 10, 2023: Completed initial therapy cycle. Partial response as
per CT chest / abdomen +PET CT:~ Moderate reduction in tumor size to
approximately 4.2 cm. Contralateral metastases still present, but no new
lesions. Partial bronchial obstruction persists with ongoing atelectasis
in the left upper lobe. Mediastinal lymph nodes remain enlarged and
FDG-positive, although with reduced metabolic activity. Thrombus in the
left atrium remains unchanged. Emphysematous and fibrotic changes are
stable. Overall, mild response observed with no significant progression,
as per RECIST stable disease.

August-October: Continued chemotherapy with Cisplatin/Pemetrexed and
Pembrolizumab.

October 13, 2023: Follow-up CT (chest + abdomen): SD / Progressive
Disease. New nodule in the right lung (1cm). Slight increase in the size
of previously noted FDG-positive lymph nodes in the mediastinum. No
additional metastatic lesions were detected. The patient has continued
to tolerate the current treatment regimen well, with no significant
adverse effects reported.

October 17, 2023: Tumor board: SD. Continuation of therapy.

October 25, 2023: Continuation of Cisplatin (dose reduced), Pemetrexed
and Pembrolizumab.

January 12, 2024: Follow-up CT scan abdomen and chest, FDG-PET-CT:
Progressive Disease with three new metastases in the right lung and
additional enlarged FDG-positive lymph nodes in the mediastinum. \colorbox{orange}{MRI scan of the brain revealed multiple metastases,} \\
\colorbox{orange}{specifically three lesions in the left hemisphere:}\\
\colorbox{orange}{one in the left frontal lobe, one in the left parietal lobe,}\\
\colorbox{orange}{and one in the left occipital lobe.} Incidental
findings included mild age-related cerebral atrophy and scattered white
matter hyperintensities consistent with chronic microvascular ischemic
changes.

March 17, 2024: Tumorboard recommends considering clinical trial options
due to limited response to standard therapies.

April 20, 2024: Detailed assessment of health status. Patient currently
in ECOG performance status 1. All routine labs, including liver and
renal function tests, within normal limits.
\end{quote}

~
\begin{center}
\textbf{===== Patient 6.1.2 =====}~~~~~~
\end{center}
\textbf{Patient Information}

Name: Ehrich, Wolfgang

born: 18.08.1968

Address: Kurfürstendamm 1, Berlin, Germany

~

\textbf{Overview of Tumor Diagnosis and Therapy}

\textbf{Tumor Diagnosis}

\begin{quote}
Diagnosis: UICC Stage IVA, M1a (contralateral metastases, malignant
pleural effusions),~

KRAS G12C mutant non-small cell lung cancer (NSCLC)
\end{quote}

Initial Detection: March 22, 2023, following symptoms of persistent
cough and weight loss

Biopsy Date: April 15, 2023, squamous cell lung cancer, PDL1 3\%

\begin{quote}
Molecular Profile: Molecular alterations: KRAS p.G12C (AF 18\%), KRAS
p.G12C (AF~

18\%), KEAP1 p.L276F (AF 45\%), STK11 p.K83Tfs*13 (AF 38\%).
\end{quote}

~

\textbf{Therapy Overview}

Combined Immuno-chemotherapy: Began May 5, 2023, with Cisplatin,

\begin{quote}
Pemetrexed and Pembrolizumab, partial response noted after cycle
completion by August 10, 2023, continuation of therapy until october
2023 (progressive disease)
\end{quote}

\textbf{Current Status}

\begin{quote}
Health Condition: Stable with an ECOG performance status of 1
\end{quote}

~

\textbf{Comorbidities}

\begin{quote}
\colorbox{orange}{Chronic heart failure (NYHA Class III), reduced ejection fraction}\\
\colorbox{orange}{(HFrEF) of 35\%}

Post Myocardial Infarction (2021), 2 coronary stents

Former Smoker: 40 py

Hypertension Stage I

COPD GOLD 2

Type 2 Diabetes Mellitus

Hyperlipidemia
\end{quote}

~

\textbf{Medication}

~~~~~~~~ Lisinopril 20mg 1-0-0

~~~~~~~~ Metformin 1000mg 1-0-1

\colorbox{orange}{ASS 100mg 1-0-0}\\

\colorbox{orange}{Carvedilol 12.5mg 1-0-1}\\

\colorbox{orange}{Furosemide 40mg 1-0-1}\\

\colorbox{orange}{Apixaban 5mg 1-0-1}\\

Atorvastatin 40mg 0-0-0-1

~~~~~~~~ Tiotropium (Inhaler) on demand

Novalgin 500mg 1-1-1-1~

\begin{quote}
~~
\end{quote}

\textbf{Chronological Medical Findings:}

\begin{quote}
March 22, 2023: Experienced persistent cough and weight loss. Chest
X-ray and CT scan revealed a mass in the left lung. Referred to
oncologist. CT-Angiography: Tumor Size: Approximately 4.5 cm in
diameter. At least 2 contralateral metastases. Bronchial Obstruction:
Partial obstruction of the left main bronchus leading to atelectasis of
the left upper lobe. Suspicion of mediastinal lymph node metastases. No
evidence of pulmonary artery embolism. Thrombus in the left atrium at
the transition to the auricle. Emphysematous and fibrotic changes in the
lung parenchyma. Urgent suspicion of a tumor-atelectasis complex in the
left upper lobe of the lung. Mucus present in the lower lobe bronchi on
the left. Lymph Nodes: Enlarged, FDG-positive lymph nodes in the
mediastinum, particularly in regions 4R and infracarinal.

April 15, 2023: Lung biopsy via bronchoscopy: Endobronchial tumor
manifestation in the distal left main bronchus extending to the upper
lobe. Acute and chronic atrophic tracheobronchitis. Collapsed bronchial
system in the affected area. Biopsy taken. Diagnosed with squamous
non-small cell lung cancer (NSCLC), molecular diagnostics: KRAS G12C
mutant.

April 27, 2023: Ventilation: Moderate obstruction, no restriction.
Increased airway resistance and slight hyperinflation. Tiffeneau index
(FEV1/FVC) at 42.34\%, z-score -3.32. FEV1: 0.93 L (42\% predicted),
z-score -2.89.Total lung capacity (TLC): 5.86 L (103\% predicted),
z-score 0.22. Forced vital capacity (FVC): 2.19 L, z-score -1.5.
Residual volume (RV): 3.67 L, z-score 2.44. RV/TLC: 62.68\%, z-score
-1.18.

May 5, 2023: Initiated on platinum-based immunochemotherapy regimen
(Cisplatin, Pemetrexed, Pembrolizumab).

August 10, 2023: Completed initial therapy cycle. Partial response as
per CT chest / abdomen +PET CT:~ Moderate reduction in tumor size to
approximately 4.2 cm. Contralateral metastases still present, but no new
lesions. Partial bronchial obstruction persists with ongoing atelectasis
in the left upper lobe. Mediastinal lymph nodes remain enlarged and
FDG-positive, although with reduced metabolic activity. Thrombus in the
left atrium remains unchanged. Emphysematous and fibrotic changes are
stable. Overall, mild response observed with no significant progression,
as per RECIST stable disease.

August-October: Continued chemotherapy with Cisplatin/Pemetrexed and
Pembrolizumab.

October 13, 2023: Follow-up CT (chest + abdomen): SD / Progressive
Disease. New nodule in the right lung (1cm). Slight increase in the size
of previously noted FDG-positive lymph nodes in the mediastinum. No
additional metastatic lesions were detected. The patient has continued
to tolerate the current treatment regimen well, with no significant
adverse effects reported.

October 17, 2023: Tumor board: SD. Continuation of therapy.

October 25, 2023: Continuation of Cisplatin (dose reduced), Pemetrexed
and Pembrolizumab.

January 12, 2024: Follow-up CT scan abdomen and chest, FDG-PET-CT:
Progressive Disease with three new metastases in the right lung and
additional enlarged FDG-positive lymph nodes in the mediastinum.\\
\colorbox{orange}{MRI scan of the brain revealed multiple metastases,}\\
\colorbox{orange}{specifically three lesions in the left hemisphere:}\\
\colorbox{orange}{one in the left frontal lobe,}\\
\colorbox{orange}{one in the left parietal lobe, and one in the left occipital lobe.} Incidental
findings included mild age-related cerebral atrophy and scattered white
matter hyperintensities consistent with chronic microvascular ischemic
changes.

March 17, 2024: Tumorboard recommends considering clinical trial options
due to limited response to standard therapies.

April 20, 2024: Detailed assessment of health status.\\
\colorbox{orange}{Patient currently
in ECOG performance status 2.}\\
\colorbox{orange}{Routine labs: GOT 103 U/L, GPT 112 U/L, Creatinine 2.3 mg/dL}
\end{quote}

~
\begin{center}
\textbf{===== Patient 6.2 =====}~~~~~~~
\end{center}
\textbf{Patient Information}

Name: Ehrich, Wolfgang

born: 18.08.1968

Address: Kurfürstendamm 1, Berlin, Germany

~

\textbf{Overview of Tumor Diagnosis and Therapy}

\textbf{Tumor Diagnosis}

\begin{quote}
Diagnosis: UICC Stage IVA, M1a (contralateral metastases, malignant
pleural effusions),~

KRAS G12C mutant non-small cell lung cancer (NSCLC)
\end{quote}

Initial Detection: March 22, 2023, following symptoms of persistent
cough and weight loss

Biopsy Date: April 15, 2023, squamous cell lung cancer, PDL1 3\%

\begin{quote}
Molecular Profile: Molecular alterations: KRAS p.G12C (AF 18\%), KRAS
p.G12C (AF~

18\%), KEAP1 p.L276F (AF 45\%), STK11 p.K83Tfs*13 (AF 38\%).
\end{quote}

~

\textbf{Therapy Overview}

Combined Immuno-chemotherapy: Began May 5, 2023, with Cisplatin,

\begin{quote}
Pemetrexed and Pembrolizumab, partial response noted after cycle
completion by August 10, 2023, continuation of therapy until october
2023 (progressive disease)
\end{quote}

\textbf{Current Status}

\begin{quote}
Health Condition: Stable with an ECOG performance status of 1
\end{quote}

~

\textbf{Comorbidities}

\begin{quote}
Former Smoker: 40 py

Hypertension Stage I

COPD GOLD 2

Type 2 Diabetes Mellitus

Hyperlipidemia
\end{quote}

~

\textbf{Medication}

~~~~~~~~ Lisinopril 20mg 1-0-0

~~~~~~~~ Metformin 1000mg 1-0-1

~~~~~~~~ Atorvastatin 40mg 0-0-0-1

~~~~~~~~ Tiotropium (Inhaler) on demand

~~~~~~~~ Novalgin 500mg 1--1-1-1~

Apixaban 5mg 1-0-1

\begin{quote}
~~
\end{quote}

\textbf{Chronological Medical Findings:}

\begin{quote}
March 22, 2023: Experienced persistent cough and weight loss. Chest
X-ray and CT scan revealed a mass in the left lung. Referred to
oncologist. CT-Angiography: Tumor Size: Approximately 4.5 cm in
diameter. At least 2 contralateral metastases. Bronchial Obstruction:
Partial obstruction of the left main bronchus leading to atelectasis of
the left upper lobe. Suspicion of mediastinal lymph node metastases. No
evidence of pulmonary artery embolism. Thrombus in the left atrium at
the transition to the auricle. Emphysematous and fibrotic changes in the
lung parenchyma. Urgent suspicion of a tumor-atelectasis complex in the
left upper lobe of the lung. Mucus present in the lower lobe bronchi on
the left. Lymph Nodes: Enlarged, FDG-positive lymph nodes in the
mediastinum, particularly in regions 4R and infracarinal.

April 15, 2023: Lung biopsy via bronchoscopy: Endobronchial tumor
manifestation in the distal left main bronchus extending to the upper
lobe. Acute and chronic atrophic tracheobronchitis. Collapsed bronchial
system in the affected area. Biopsy taken. Diagnosed with squamous
non-small cell lung cancer (NSCLC), molecular diagnostics: KRAS G12C
mutant.

April 27, 2023: Ventilation: Moderate obstruction, no restriction.
Increased airway resistance and slight hyperinflation. Tiffeneau index
(FEV1/FVC) at 42.34\%, z-score -3.32. FEV1: 0.93 L (42\% predicted),
z-score -2.89.Total lung capacity (TLC): 5.86 L (103\% predicted),
z-score 0.22. Forced vital capacity (FVC): 2.19 L, z-score -1.5.
Residual volume (RV): 3.67 L, z-score 2.44. RV/TLC: 62.68\%, z-score
-1.18.

May 5, 2023: Initiated on platinum-based immunochemotherapy regimen
(Cisplatin, Pemetrexed, Pembrolizumab).

August 10, 2023: Completed initial therapy cycle. Partial response as
per CT chest / abdomen +PET CT:~ Moderate reduction in tumor size to
approximately 4.2 cm. Contralateral metastases still present, but no new
lesions. Partial bronchial obstruction persists with ongoing atelectasis
in the left upper lobe. Mediastinal lymph nodes remain enlarged and
FDG-positive, although with reduced metabolic activity. Thrombus in the
left atrium remains unchanged. Emphysematous and fibrotic changes are
stable. Overall, mild response observed with no significant progression,
as per RECIST stable disease.

August-October: Continued chemotherapy with Cisplatin/Pemetrexed and
Pembrolizumab.

October 13, 2023: Follow-up CT (chest + abdomen): SD / Progressive
Disease. New nodule in the right lung (1cm). Slight increase in the size
of previously noted FDG-positive lymph nodes in the mediastinum. No
additional metastatic lesions were detected. The patient has continued
to tolerate the current treatment regimen well, with no significant
adverse effects reported.

October 17, 2023: Tumor board: SD. Continuation of therapy.

October 25, 2023: Continuation of Cisplatin (dose reduced), Pemetrexed
and Pembrolizumab.

January 12, 2024: Follow-up CT scan abdomen and chest, FDG-PET-CT:
Progressive Disease with three new metastases in the right lung and
additional enlarged FDG-positive lymph nodes in the mediastinum. MRI
scan of the brain conducted; no evidence of metastatic disease.
Incidental findings included mild age-related cerebral atrophy and
scattered white matter hyperintensities consistent with chronic
microvascular ischemic changes.

March 17, 2024: Tumorboard recommends considering clinical trial options
due to limited response to standard therapies.

April 20, 2024: Detailed assessment of health status. ECOG performance
status 1. All routine labs, including liver and renal function tests,
within normal limits.
\end{quote}

~

\begin{center}
\textbf{===== Patient 6.2.1 =====}~~~~~~~~
\end{center}
\textbf{Patient Information}

Name: Ehrich, Wolfgang

born: 18.08.1968

Address: Kurfürstendamm 1, Berlin, Germany

~

\textbf{Overview of Tumor Diagnosis and Therapy}

\textbf{Tumor Diagnosis}

\begin{quote}
Diagnosis: \colorbox{orange}{UICC Stage IVB, M1c (brain metastases)},

KRAS G12C mutant non-small cell lung cancer (NSCLC)
\end{quote}

Initial Detection: March 22, 2023, following symptoms of persistent
cough and weight loss

Biopsy Date: April 15, 2023, squamous cell lung cancer, PDL1 3\%

\begin{quote}
Molecular Profile: Molecular alterations: KRAS p.G12C (AF 18\%), KRAS
p.G12C (AF~

18\%), KEAP1 p.L276F (AF 45\%), STK11 p.K83Tfs*13 (AF 38\%).
\end{quote}

~

\textbf{Therapy Overview}

Combined Immuno-chemotherapy: Began May 5, 2023, with Cisplatin,

\begin{quote}
Pemetrexed and Pembrolizumab, partial response noted after cycle
completion by August 10, 2023, continuation of therapy until october
2023 (progressive disease)
\end{quote}

\textbf{Current Status}

\begin{quote}
Health Condition: Stable with an ECOG performance status of 1
\end{quote}

~

\textbf{Comorbidities}

\begin{quote}
Former Smoker: 40 py

Hypertension Stage I

COPD GOLD 2

Type 2 Diabetes Mellitus

Hyperlipidemia
\end{quote}

~

\textbf{Medication}

~~~~~~~~ Lisinopril 20mg 1-0-0

~~~~~~~~ Metformin 1000mg 1-0-1

~~~~~~~~ Atorvastatin 40mg 0-0-0-1

~~~~~~~~ Tiotropium (Inhaler) on demand

~~~~~~~~ Novalgin 500mg 1--1-1-1~

Apixaban 5mg 1-0-1

\begin{quote}
~~
\end{quote}

\textbf{Chronological Medical Findings:}

\begin{quote}
March 22, 2023: Experienced persistent cough and weight loss. Chest
X-ray and CT scan revealed a mass in the left lung. Referred to
oncologist. CT-Angiography: Tumor Size: Approximately 4.5 cm in
diameter. At least 2 contralateral metastases. Bronchial Obstruction:
Partial obstruction of the left main bronchus leading to atelectasis of
the left upper lobe. Suspicion of mediastinal lymph node metastases. No
evidence of pulmonary artery embolism. Thrombus in the left atrium at
the transition to the auricle. Emphysematous and fibrotic changes in the
lung parenchyma. Urgent suspicion of a tumor-atelectasis complex in the
left upper lobe of the lung. Mucus present in the lower lobe bronchi on
the left. Lymph Nodes: Enlarged, FDG-positive lymph nodes in the
mediastinum, particularly in regions 4R and infracarinal.

April 15, 2023: Lung biopsy via bronchoscopy: Endobronchial tumor
manifestation in the distal left main bronchus extending to the upper
lobe. Acute and chronic atrophic tracheobronchitis. Collapsed bronchial
system in the affected area. Biopsy taken. Diagnosed with squamous
non-small cell lung cancer (NSCLC), molecular diagnostics: KRAS G12C
mutant.

April 27, 2023: Ventilation: Moderate obstruction, no restriction.
Increased airway resistance and slight hyperinflation. Tiffeneau index
(FEV1/FVC) at 42.34\%, z-score -3.32. FEV1: 0.93 L (42\% predicted),
z-score -2.89.Total lung capacity (TLC): 5.86 L (103\% predicted),
z-score 0.22. Forced vital capacity (FVC): 2.19 L, z-score -1.5.
Residual volume (RV): 3.67 L, z-score 2.44. RV/TLC: 62.68\%, z-score
-1.18.

May 5, 2023: Initiated on platinum-based immunochemotherapy regimen
(Cisplatin, Pemetrexed, Pembrolizumab).

August 10, 2023: Completed initial therapy cycle. Partial response as
per CT chest / abdomen +PET CT:~ Moderate reduction in tumor size to
approximately 4.2 cm. Contralateral metastases still present, but no new
lesions. Partial bronchial obstruction persists with ongoing atelectasis
in the left upper lobe. Mediastinal lymph nodes remain enlarged and
FDG-positive, although with reduced metabolic activity. Thrombus in the
left atrium remains unchanged. Emphysematous and fibrotic changes are
stable. Overall, mild response observed with no significant progression,
as per RECIST stable disease.

August-October: Continued chemotherapy with Cisplatin/Pemetrexed and
Pembrolizumab.

October 13, 2023: Follow-up CT (chest + abdomen): SD / Progressive
Disease. New nodule in the right lung (1cm). Slight increase in the size
of previously noted FDG-positive lymph nodes in the mediastinum. No
additional metastatic lesions were detected. The patient has continued
to tolerate the current treatment regimen well, with no significant
adverse effects reported.

October 17, 2023: Tumor board: SD. Continuation of therapy.

October 25, 2023: Continuation of Cisplatin (dose reduced), Pemetrexed
and Pembrolizumab.

January 12, 2024: Follow-up CT scan abdomen and chest, FDG-PET-CT:
Progressive Disease with three new metastases in the right lung and
additional enlarged FDG-positive lymph nodes in the mediastinum. \colorbox{orange}{MRI scan of the brain: MRI scan of the brain revealed multiple} \\
\colorbox{orange}{metastases, specifically three lesions in the left hemisphere:}\\
\colorbox{orange}{one in the left frontal lobe,}\\
\colorbox{orange}{one in the left parietal lobe, and one in the left occipital lobe.}

\colorbox{orange}{January 16, 2024: Begin Sotorasib (Lumakras) 960 per day}

March 17, 2024: Tumorboard recommends considering clinical trial options
due to limited response to standard therapies.

April 20, 2024: Detailed assessment of health status. ECOG performance
status 1. All routine labs, including liver and renal function tests,
within normal limits.
\end{quote}

~
\begin{center}
\textbf{===== Patient 6.3 =====}~~~~~~~~
\end{center}
\textbf{Patient Information}

Name: Ehrich, Wolfgang

born: 18.08.1968

Address: Italy

~

\textbf{Overview of Tumor Diagnosis and Therapy}

\textbf{Tumor Diagnosis}

\begin{quote}
Diagnosis: UICC Stage IVA, M1a (contralateral metastases),~

KRAS G12C mutant non-small cell lung cancer (NSCLC) / adenocarcinoma of
the~

lung

Initial Detection: March 22, 2024, following symptoms of persistent
cough and weight loss (-5kg)
\end{quote}

Biopsy Date: April 15, 2024, adenocarcinoma, PDL1 3\%

\begin{quote}
Molecular Profile: Molecular alterations: KRAS p.G12C (AF 18\%), KRAS
p.G12C (AF~

18\%), KEAP1 p.L276F (AF 45\%), STK11 p.K83Tfs*13 (AF 38\%).
\end{quote}

~

\textbf{Therapy Overview}

\begin{quote}
None.
\end{quote}

\textbf{Current Status}

\begin{quote}
Health Condition: Stable with an ECOG performance status of 1
\end{quote}

\textbf{Allergies:} None.

\textbf{Comorbidities}

\begin{quote}
Former Smoker: 40 py

Hypertension Stage I

COPD GOLD 2

Type 2 Diabetes Mellitus

Hyperlipidemia
\end{quote}

~

\textbf{Medication}

~~~~~~~~ Lisinopril 20mg 1-0-0

~~~~~~~~ Metformin 1000mg 1-0-1

~~~~~~~~ Atorvastatin 40mg 0-0-0-1

~~~~~~~~ Tiotropium (Inhaler) on demand

~~~~~~~~ Novalgin 500mg 1--1-1-1~

Apixaban 5mg 1-0-1

\begin{quote}
~~
\end{quote}

\textbf{Chronological Medical Findings:}

\begin{quote}
\textbf{March 22, 2024:} Experienced persistent cough and weight loss.
Chest X-ray and CT scan revealed a mass in the left lung. Referred to
oncologist. CT-Angiography: Tumor Size: Approximately 4.5 cm in
diameter. At least 2 contralateral metastases. Bronchial Obstruction:
Partial obstruction of the left main bronchus leading to atelectasis of
the left upper lobe. Suspicion of mediastinal lymph node metastases. No
evidence of pulmonary artery embolism. Thrombus in the left atrium at
the transition to the auricle. Emphysematous and fibrotic changes in the
lung parenchyma. Urgent suspicion of a tumor-atelectasis complex in the
left upper lobe of the lung. Mucus present in the lower lobe bronchi on
the left. Lymph Nodes: Enlarged, FDG-positive lymph nodes in the
mediastinum, particularly in regions 4R and infracarinal.

\textbf{April 15, 2024:} Lung biopsy via bronchoscopy: Endobronchial
tumor manifestation in the distal left main bronchus extending to the
upper lobe. Acute and chronic atrophic tracheobronchitis. Collapsed
bronchial system in the affected area. Biopsy taken. Diagnosed with
non-small cell lung cancer (NSCLC) (adenocarcinoma), molecular
diagnostics: KRAS G12C mutant.

\textbf{April 27, 2024:} Ventilation: Moderate obstruction, no
restriction. Increased airway resistance and slight hyperinflation.
Tiffeneau index (FEV1/FVC) at 42.34\%, z-score -3.32. FEV1: 0.93 L (42\%
predicted), z-score -2.89.Total lung capacity (TLC): 5.86 L (103\%
predicted), z-score 0.22. Forced vital capacity (FVC): 2.19 L, z-score
-1.5. Residual volume (RV): 3.67 L, z-score 2.44. RV/TLC: 62.68\%,
z-score -1.18.

\textbf{April 20, 2024:} Detailed assessment of health status. ECOG
performance status 1. All routine labs, including liver and renal
function tests, within normal limits. Discussion in tumor board
conference: palliative systemic treatment or clinical trial enrollment.
\end{quote}

\begin{center}
\textbf{===== Patient 6.3.1 =====}~~~~~~~~
\end{center}
\textbf{Patient Information}

Name: Ehrich, Wolfgang

born: 18.08.1968

Address: Italy

~

\textbf{Overview of Tumor Diagnosis and Therapy}

\textbf{Tumor Diagnosis}

\begin{quote}
Diagnosis: UICC Stage IVA, M1a (contralateral metastases, \colorbox{orange}{pleural
effusions}),~

KRAS G12C mutant non-small cell lung cancer (NSCLC) / adenocarcinoma of
the~

lung

Initial Detection: March 22, 2024, following symptoms of persistent
cough and weight loss (-5kg)
\end{quote}

Biopsy Date: April 15, 2024, adenocarcinoma, PDL1 3\%

\begin{quote}
Molecular Profile: Molecular alterations: KRAS p.G12C (AF 18\%), KRAS
p.G12C (AF~

18\%), KEAP1 p.L276F (AF 45\%), STK11 p.K83Tfs*13 (AF 38\%).
\end{quote}

~

\textbf{Therapy Overview}

\begin{quote}
None.
\end{quote}

\textbf{Current Status}

\begin{quote}
Health Condition: Stable with an ECOG performance status of 1
\end{quote}

\textbf{Allergies:} None.

\textbf{Comorbidities}

\begin{quote}
Former Smoker: 40 py

Hypertension Stage I

COPD GOLD 2

Type 2 Diabetes Mellitus

Hyperlipidemia
\end{quote}

~

\textbf{Medication}

~~~~~~~~ Lisinopril 20mg 1-0-0

~~~~~~~~ Metformin 1000mg 1-0-1

~~~~~~~~ Atorvastatin 40mg 0-0-0-1

~~~~~~~~ Tiotropium (Inhaler) on demand

~~~~~~~~ Novalgin 500mg 1--1-1-1~

Apixaban 5mg 1-0-1

\begin{quote}
~~
\end{quote}

\textbf{Chronological Medical Findings:}

\begin{quote}
\textbf{March 22, 2024:} Experienced persistent cough and weight loss.
Chest X-ray and CT scan revealed a mass in the left lung. Referred to
oncologist. CT-Angiography: Tumor Size: Approximately 4.5 cm in
diameter. At least 2 contralateral metastases. Bronchial Obstruction:
Partial obstruction of the left main bronchus leading to atelectasis of
the left upper lobe. Suspicion of mediastinal lymph node metastases. No
evidence of pulmonary artery embolism. Thrombus in the left atrium at
the transition to the auricle. Emphysematous and fibrotic changes in the
lung parenchyma. Urgent suspicion of a tumor-atelectasis complex in the
left upper lobe of the lung. Mucus present in the lower lobe bronchi on
the left. Lymph Nodes: Enlarged, FDG-positive lymph nodes in the
mediastinum, particularly in regions 4R and infracarinal. \colorbox{orange}{Bilateral
pleural effusions, additionally mild pericardial effusion.}

\textbf{April 15, 2024:} Lung biopsy via bronchoscopy: Endobronchial
tumor manifestation in the distal left main bronchus extending to the
upper lobe. Acute and chronic atrophic tracheobronchitis. Collapsed
bronchial system in the affected area. Biopsy taken. Diagnosed with
non-small cell lung cancer (NSCLC) (adenocarcinoma), molecular
diagnostics: KRAS G12C mutant.

\textbf{April 27, 2024:} Ventilation: Moderate obstruction, no
restriction. Increased airway resistance and slight hyperinflation.
Tiffeneau index (FEV1/FVC) at 42.34\%, z-score -3.32. FEV1: 0.93 L (42\%
predicted), z-score -2.89.Total lung capacity (TLC): 5.86 L (103\%
predicted), z-score 0.22. Forced vital capacity (FVC): 2.19 L, z-score
-1.5. Residual volume (RV): 3.67 L, z-score 2.44. RV/TLC: 62.68\%,
z-score -1.18.

\textbf{April 20, 2024:} Detailed assessment of health status.\\
\colorbox{orange}{ECOG performance status 2}. All routine labs, including liver and renal
function tests, within normal limits. Discussion in tumor board
conference: palliative systemic treatment or clinical trial enrollment.
\end{quote}

~~~~~~

\begin{center}
\textbf{===== Patient 6.4 =====}~~~~~~~~
\end{center}
\textbf{Patient Information}

Name: Ehrich, Wolfgang

born: 18.08.1968

Address: Kurfürstendamm 1, Berlin, Germany

~

\textbf{Overview of Tumor Diagnosis and Therapy}

\textbf{Tumor Diagnosis}

\begin{quote}
Diagnosis: UICC Stage IVA, M1a (contralateral metastases, malignant
pleural effusions),~

KRAS G12C mutant non-small cell lung cancer (NSCLC)
\end{quote}

Initial Detection: March 22, 2023, following symptoms of persistent
cough and weight loss

Biopsy Date: April 15, 2023, squamous cell lung cancer, PDL1 3\%

\begin{quote}
Molecular Profile: Molecular alterations: KRAS p.G12C (AF 18\%), KRAS
p.G12C (AF~

18\%), KEAP1 p.L276F (AF 45\%), STK11 p.K83Tfs*13 (AF 38\%).
\end{quote}

~

\textbf{Therapy Overview}

Combined Immuno-chemotherapy: Began May 5, 2023, with Cisplatin,

\begin{quote}
Pemetrexed and Pembrolizumab, partial response noted after cycle
completion by August 10, 2023, continuation of therapy until october
2023 (progressive disease)
\end{quote}

\textbf{Current Status}

\begin{quote}
Health Condition: Stable with an ECOG performance status of 1
\end{quote}

~

\textbf{Comorbidities}

\begin{quote}
Former Smoker: 40 py

Hypertension Stage I

COPD GOLD 2

Type 2 Diabetes Mellitus

Hyperlipidemia
\end{quote}

~

\textbf{Medication}

~~~~~~~~ Lisinopril 20mg 1-0-0

~~~~~~~~ Metformin 1000mg 1-0-1

~~~~~~~~ Atorvastatin 40mg 0-0-0-1

~~~~~~~~ Tiotropium (Inhaler) on demand

~~~~~~~~ Novalgin 500mg 1--1-1-1~

Apixaban 5mg 1-0-1

\begin{quote}
~~
\end{quote}

\textbf{Chronological Medical Findings:}

\begin{quote}
March 22, 2023: Experienced persistent cough and weight loss. Chest
X-ray and CT scan revealed a mass in the left lung. Referred to
oncologist. CT-Angiography: Tumor Size: Approximately 4.5 cm in
diameter. At least 2 contralateral metastases. Bronchial Obstruction:
Partial obstruction of the left main bronchus leading to atelectasis of
the left upper lobe. Suspicion of mediastinal lymph node metastases. No
evidence of pulmonary artery embolism. Thrombus in the left atrium at
the transition to the auricle. Emphysematous and fibrotic changes in the
lung parenchyma. Urgent suspicion of a tumor-atelectasis complex in the
left upper lobe of the lung. Mucus present in the lower lobe bronchi on
the left. Lymph Nodes: Enlarged, FDG-positive lymph nodes in the
mediastinum, particularly in regions 4R and infracarinal.

April 15, 2023: Lung biopsy via bronchoscopy: Endobronchial tumor
manifestation in the distal left main bronchus extending to the upper
lobe. Acute and chronic atrophic tracheobronchitis. Collapsed bronchial
system in the affected area. Biopsy taken. Diagnosed with squamous
non-small cell lung cancer (NSCLC), molecular diagnostics: KRAS G12C
mutant.

April 27, 2023: Ventilation: Moderate obstruction, no restriction.
Increased airway resistance and slight hyperinflation. Tiffeneau index
(FEV1/FVC) at 42.34\%, z-score -3.32. FEV1: 0.93 L (42\% predicted),
z-score -2.89.Total lung capacity (TLC): 5.86 L (103\% predicted),
z-score 0.22. Forced vital capacity (FVC): 2.19 L, z-score -1.5.
Residual volume (RV): 3.67 L, z-score 2.44. RV/TLC: 62.68\%, z-score
-1.18.

May 5, 2023: Initiated on platinum-based immunochemotherapy regimen
(Cisplatin, Pemetrexed, Pembrolizumab).

August 10, 2023: Completed initial therapy cycle. Partial response as
per CT chest / abdomen +PET CT:~ Moderate reduction in tumor size to
approximately 4.2 cm. Contralateral metastases still present, but no new
lesions. Partial bronchial obstruction persists with ongoing atelectasis
in the left upper lobe. Mediastinal lymph nodes remain enlarged and
FDG-positive, although with reduced metabolic activity. Thrombus in the
left atrium remains unchanged. Emphysematous and fibrotic changes are
stable. Overall, mild response observed with no significant progression,
as per RECIST stable disease.

August-October: Continued chemotherapy with Cisplatin/Pemetrexed and
Pembrolizumab.

October 13, 2023: Follow-up CT (chest + abdomen): SD / Progressive
Disease. New nodule in the right lung (1cm). Slight increase in the size
of previously noted FDG-positive lymph nodes in the mediastinum. No
additional metastatic lesions were detected. The patient has continued
to tolerate the current treatment regimen well, with no significant
adverse effects reported.

October 17, 2023: Tumor board: SD. Continuation of therapy.

October 25, 2023: Continuation of Cisplatin (dose reduced), Pemetrexed
and Pembrolizumab.

January 12, 2024: Follow-up CT scan abdomen and chest, FDG-PET-CT:
Progressive Disease with three new metastases in the right lung and
additional enlarged FDG-positive lymph nodes in the mediastinum. Primary
tumor 5.1 cm in diameter. MRI scan of the brain conducted; no evidence
of metastatic disease. Incidental findings included mild age-related
cerebral atrophy and scattered white matter hyperintensities consistent
with chronic microvascular ischemic changes.

March 17, 2024: Tumorboard recommends considering clinical trial options
due to limited response to standard therapies.

April 20, 2024: Detailed assessment of health status. ECOG performance
status 1. All routine labs, including liver and renal function tests,
within normal limits.
\end{quote}

~~~~~~~~
\begin{center}
\textbf{===== Patient 7.1 =====}~~~~~~~~
\end{center}
\textbf{Patient Information\\
}Name: Jessica Smith\\
Born: August 10, 1982\\
Address: Hamburg, Germany

~

\textbf{Overview of Tumor Diagnosis and Therapy}

\textbf{Tumor Diagnosis\\
}Diagnosis: UICC Stage IV metastatic malignant melanoma (hepatic, M.
pectoralis major)~

Initial Detection: January 5, 2024, following a rapidly growing mole and
enlarged lymph nodes\\
Biopsy Date: January 15, 2024\\
Molecular Profile: Tumor Purity: 80\%; Tumor Mutational Burden (TMB):
12.8 Mut/Mb; NF1 p.I1605fs (AF 39\%), TP53 c.672+1G\textgreater A (AF
50\%), RB1 p.Q846* (AF 20\%), TERT p.R859Q (AF 41\%)

\textbf{Therapy Overview\\
}Initial Treatment: None so far.\\
\strut \\
\textbf{Health Condition:} ECOG 1

~

\textbf{Comorbidities}

\begin{quote}
Former smoker 10 py

Hypertension Stage 1

Mild Asthma

H/o appendectomy 2014
\end{quote}

\textbf{Medication}

\begin{quote}
Amlodipine 10mg 1-0-0

Albuterol inhaler as needed
\end{quote}

~

\textbf{Chronological Medical Findings:}

\textbf{January 5, 2024:} Presented with a rapidly growing mole on the
left arm and enlarged lymph nodes in the axillary region.

\textbf{January 10, 2024:} CT scan of the chest and abdomen: Solid tumor
in the left axillary region measuring approximately 3.5 cm with evidence
of local invasion into surrounding soft tissues and possibly the
pectoralis major muscle. Demonstrates irregular borders and
heterogeneous enhancement. Multiple hypodense lesions noted throughout
the liver, suggestive of metastatic disease. The largest lesion is
located in segment VIII, measuring approximately 2.8 cm in diameter.
Additional smaller lesions scattered in both hepatic lobes.

\textbf{January 15, 2024:} Biopsy of the left axillary mass performed.
Histology confirmed melanoma. Molecular panel sequencing: NF1 p.I1605fs
(AF 39\%), TP53 c.672+1G\textgreater A (AF 50\%), RB1 p.Q846* (AF 20\%),
TERT p.R859Q (AF 41\%). Tumor purity 80\%. Tumor Mutational Burden (TMB)
12.8 Mut/Mb.

\textbf{January 16, 2024:} Detailed assessment of health status
confirmed adequate organ function. Routine labs within normal limits:
ANC 5,300/mcL, platelet count 140,000/mcL, total bilirubin 1.1 mg/dL,
AST/ALT within range, creatinine 1.1 mg/dL, hemoglobin 10.5 g/dL, serum
albumin 3.4 g/dL, lipase and amylase within normal limits.

~~~~~~~~
\begin{center}
\textbf{===== Patient 7.1.1 =====}~~~~~~~~
\end{center}
\textbf{Patient Information\\
}Name: Jessica Smith\\
Born: August 10, 1982\\
Address: Hamburg, Germany

~

\textbf{Overview of Tumor Diagnosis and Therapy}

\textbf{Tumor Diagnosis\\
}Diagnosis: UICC Stage IV metastatic malignant melanoma (HEP, M.
pectoralis major)~

Initial Detection: January 5, 2024, following a rapidly growing mole and
enlarged lymph nodes\\
Biopsy Date: January 15, 2024\\
Molecular Profile: Tumor Purity: 80\%; Tumor Mutational Burden (TMB):
12.8 Mut/Mb; NF1 p.I1605fs (AF 39\%), TP53 c.672+1G\textgreater A (AF
50\%), RB1 p.Q846* (AF 20\%), TERT p.R859Q (AF 41\%)

\textbf{Therapy Overview\\
}Initial Treatment:\\
\colorbox{orange}{Immunotherapy: Began February 1, 2024, with Nivolumab and Ipilimumab,}\\
\colorbox{orange}{partial response noted}\\
\colorbox{orange}{after the initial treatment cycle completed by
May 15, 2024.}\\
\colorbox{orange}{Continued Nivolumab maintenance until December 2024
(progressive disease).}\\
\strut \\
\colorbox{orange}{\textbf{Current Status:} Disease progression as of December 2024,}\\
\colorbox{orange}{with new metastatic lesions identified.}\\
\textbf{Health Condition:} ECOG 1

~

\textbf{Comorbidities}

\begin{quote}
Former smoker 10 py

Hypertension Stage 1

Mild Asthma

H/o appendectomy 2014
\end{quote}

\textbf{Medication}

\begin{quote}
Amlodipine 10mg 1-0-0

Albuterol inhaler as needed
\end{quote}

~

\textbf{Chronological Medical Findings:}

\textbf{January 5, 2023:} Presented with a rapidly growing mole on the
left arm and enlarged lymph nodes in the axillary region.

\textbf{January 10, 2023:} CT scan of the chest and abdomen: Solid tumor
in the left axillary region measuring approximately 3.5 cm with evidence
of local invasion into surrounding soft tissues and possibly the
pectoralis major muscle. Demonstrates irregular borders and
heterogeneous enhancement. Multiple hypodense lesions noted throughout
the liver, suggestive of metastatic disease. The largest lesion is
located in segment VIII, measuring approximately 2.8 cm in diameter.
Additional smaller lesions scattered in both hepatic lobes.

\textbf{January 15, 2023:} Biopsy of the left axillary mass performed.
Histology confirmed melanoma. Molecular panel sequencing: NF1 p.I1605fs
(AF 39\%), TP53 c.672+1G\textgreater A (AF 50\%), RB1 p.Q846* (AF 20\%),
TERT p.R859Q (AF 41\%). Tumor purity 80\%. Tumor Mutational Burden (TMB)
12.8 Mut/Mb.

\colorbox{orange}{\textbf{February 1, 2023:} Initiated combined immunotherapy with
Nivolumab and}\\
\colorbox{orange}{Ipilimumab.}

\colorbox{orange}{\textbf{May 5, 2023:} CT scan showed partial response with a decrease in the size of}\\
\colorbox{orange}{the primary tumor}\\
\colorbox{orange}{and axillary lymph nodes. Partial response also regarding liver mets.}\\
\colorbox{orange}{Continued maintenance therapy with Nivolumab.}

\colorbox{orange}{\textbf{September 15, 2023:} Follow-up imaging: SD.}

\colorbox{orange}{\textbf{September - December 2023:} Continuation of Nivolumab.}~

\colorbox{orange}{\textbf{December 18, 2023:} Follow-up CT scan: Disease progression with new}\\
\colorbox{orange}{metastatic lesions in the liver and bones.}\\
\colorbox{orange}{Multiple enlarged lymph nodes persistent in the left axillary region,}\\
\colorbox{orange}{consistent with known metastatic melanoma.}\\
\colorbox{orange}{No significant change in size or number compared to
the previous scan.}\\
\colorbox{orange}{Liver demonstrates multiple hypodense lesions
throughout both hepatic lobes.}\\
\colorbox{orange}{The large lesion located in segment VIII
now measures approximately}\\
\colorbox{orange}{5.0 cm in diameter.}\\
\colorbox{orange}{Previously noted lesions have increased in size, with the largest lesion in segment }\\
\colorbox{orange}{IVa now measuring 4.2 cm (previously 3.1 cm).}\\
\colorbox{orange}{New lytic lesions are identified in the thoracic spine,}\\ \colorbox{orange}{specifically at T5 and T8 vertebral bodies, suggestive of metastatic disease.}

\colorbox{orange}{\textbf{December 21, 2023:} Bone scan confirmed multiple metastatic
lesions in the}\\
\colorbox{orange}{thoracic spine.}

\colorbox{orange}{\textbf{January 4, 2024:} Tumor board review recommended considering eligibility for }\\
\colorbox{orange}{clinical trials due to limited response to standard and investigational therapies.}

\textbf{January 16, 2024:} Detailed assessment of health status
confirmed adequate organ function. Routine labs within normal limits:
ANC 5,300/mcL, platelet count 140,000/mcL, total bilirubin 1.1 mg/dL,
AST/ALT within range, creatinine 1.1 mg/dL, hemoglobin 10.5 g/dL, serum
albumin 3.4 g/dL, lipase and amylase within normal limits.

~~~~~~~~
\begin{center}
\textbf{===== Patient 7.1.2 =====}~~~~~~~~
\end{center}
\textbf{Patient Information\\
}Name: Jessica Smith\\
Born: August 10, 1982\\
Address: Hamburg, Germany

~

\textbf{Overview of Tumor Diagnosis and Therapy}

\textbf{Tumor Diagnosis\\
}Diagnosis: UICC Stage IV metastatic malignant melanoma (hepatic, M.
pectoralis major, Bone)~

Initial Detection: January 5, 2024, following a rapidly growing mole and
enlarged lymph nodes\\
Biopsy Date: January 15, 2024\\
Molecular Profile: Tumor Purity: 80\%; Tumor Mutational Burden (TMB):
12.8 Mut/Mb; NF1 p.I1605fs (AF 39\%), TP53 c.672+1G\textgreater A (AF
50\%), RB1 p.Q846* (AF 20\%), TERT p.R859Q (AF 41\%)

\textbf{Therapy Overview\\
}Initial Treatment: None so far.\\
\strut \\
\textbf{Health Condition:} ECOG 1

~

\textbf{Comorbidities}

\begin{quote}
Former smoker 10 py

Hypertension Stage 1

Mild Asthma

H/o appendectomy 2014

\colorbox{orange}{Systemic Lupus Erythematosus (SLE) diagnosed in 2022,}\\
\colorbox{orange}{presenting with joint pain, fatigue, and a malar rash}
\end{quote}

\textbf{Medication}

\begin{quote}
\colorbox{orange}{Hydroxychloroquine 200 mg, once daily

Prednisone 5 mg, once daily}

Amlodipine 10mg 1-0-0

Albuterol inhaler as needed
\end{quote}

~

\textbf{Chronological Medical Findings:}

\textbf{January 5, 2024:} Presented with a rapidly growing mole on the
left arm and enlarged lymph nodes in the axillary region.

\textbf{January 10, 2024:} CT scan of the chest and abdomen: Solid tumor
in the left axillary region measuring approximately 3.5 cm with evidence
of local invasion into surrounding soft tissues and possibly the
pectoralis major muscle. Demonstrates irregular borders and
heterogeneous enhancement. Multiple hypodense lesions noted throughout
the liver, suggestive of metastatic disease. The largest lesion is
located in segment VIII, measuring approximately 2.8 cm in diameter.
Additional smaller lesions scattered in both hepatic lobes.

\textbf{January 15, 2024:} Biopsy of the left axillary mass performed.
Histology confirmed melanoma. Molecular panel sequencing: NF1 p.I1605fs
(AF 39\%), TP53 c.672+1G\textgreater A (AF 50\%), RB1 p.Q846* (AF 20\%),
TERT p.R859Q (AF 41\%). Tumor purity 80\%. Tumor Mutational Burden (TMB)
12.8 Mut/Mb.

\textbf{January 16, 2024:} Detailed assessment of health status
confirmed adequate organ function. Routine labs within normal limits:
ANC 5,300/mcL, platelet count 140,000/mcL, total bilirubin 1.1 mg/dL,
AST/ALT within range, creatinine 1.1 mg/dL, hemoglobin 10.5 g/dL, serum
albumin 3.4 g/dL, lipase and amylase within normal limits.

~
\begin{center}
\textbf{===== Patient 7.1.3 =====}~~~~~~~~
\end{center}
\textbf{Patient Information\\
}Name: Jessica Smith\\
Born: August 10, 1982\\
Address: Hamburg, Germany

~

\textbf{Overview of Tumor Diagnosis and Therapy}

\textbf{Tumor Diagnosis\\
}Diagnosis: UICC Stage IV metastatic malignant melanoma (hepatic, M.
pectoralis major, brain)~

Initial Detection: January 5, 2024, following a rapidly growing mole and
enlarged lymph nodes\\
Biopsy Date: January 15, 2024\\
Molecular Profile: Tumor Purity: 80\%; Tumor Mutational Burden (TMB):
12.8 Mut/Mb; NF1 p.I1605fs (AF 39\%), TP53 c.672+1G\textgreater A (AF
50\%), RB1 p.Q846* (AF 20\%), TERT p.R859Q (AF 41\%)

\textbf{Therapy Overview\\
}Initial Treatment: None so far.\\
\strut \\
\textbf{Health Condition:} ECOG 1

~

\textbf{Comorbidities}

\begin{quote}
Former smoker 10 py

Hypertension Stage 1

Mild Asthma

H/o appendectomy 2014

\colorbox{orange}{Systemic Lupus Erythematosus (SLE) diagnosed in 2022,}\\
\colorbox{orange}{presenting with joint pain, fatigue, and a malar rash}
\end{quote}

\textbf{Medication}

\begin{quote}
\colorbox{orange}{Hydroxychloroquine 200 mg, once daily

Prednisone 5 mg, once daily}

Amlodipine 10mg 1-0-0

Albuterol inhaler as needed
\end{quote}

~

\textbf{Chronological Medical Findings:}

\textbf{January 5, 2024:} Presented with a rapidly growing mole on the
left arm and enlarged lymph nodes in the axillary region.

\textbf{January 10, 2024:} CT scan of the chest and abdomen: Solid tumor
in the left axillary region measuring approximately 3.5 cm with evidence
of local invasion into surrounding soft tissues and possibly the
pectoralis major muscle. Demonstrates irregular borders and
heterogeneous enhancement. Multiple hypodense lesions noted throughout
the liver, suggestive of metastatic disease. The largest lesion is
located in segment VIII, measuring approximately 2.8 cm in diameter.
Additional smaller lesions scattered in both hepatic lobes.

\colorbox{orange}{cMRI: Imaging reveals five small brain metastases:}\\
\colorbox{orange}{A 1.2 cm lesion in the right frontal lobe.}\\
\colorbox{orange}{A 0.8 cm lesion in the left parietal lobe. A 0.6
cm lesion in the right occipital lobe.}\\
\colorbox{orange}{A 0.7 cm lesion in the left cerebellum. A 0.9 cm lesion in the right temporal lobe.}\\
\colorbox{orange}{All lesions with heterogeneous enhancement and associated with surrounding}\\
\colorbox{orange}{vasogenic edema.}\\
\colorbox{orange}{No evidence of midline shift or significant mass effect at this
time.}~

\textbf{January 15, 2024:} Biopsy of the left axillary mass performed.
Histology confirmed melanoma. Molecular panel sequencing: NF1 p.I1605fs
(AF 39\%), TP53 c.672+1G\textgreater A (AF 50\%), RB1 p.Q846* (AF 20\%),
TERT p.R859Q (AF 41\%). Tumor purity 80\%. Tumor Mutational Burden (TMB)
12.8 Mut/Mb.

\textbf{January 16, 2024:} Detailed assessment of health status
confirmed adequate organ function. Routine labs within normal limits:
ANC 5,300/mcL, platelet count 140,000/mcL, total bilirubin 1.1 mg/dL,
AST/ALT within range, creatinine 1.1 mg/dL, hemoglobin 10.5 g/dL, serum
albumin 3.4 g/dL, lipase and amylase within normal limits.

~~~~~~~~~
\begin{center}
\textbf{===== Patient 7.2 =====}~~~~~~~~~
\end{center}
\textbf{Patient Information}

Name: Jessica Smith

Born: August 10, 1982

Address: Hamburg, Germany

~

\textbf{Overview of Tumor Diagnosis and Therapy}

\textbf{Tumor Diagnosis}

\textbf{Diagnosis}: UICC Stage IV metastatic malignant melanoma
(hepatic, M. pectoralis major)

Initial Detection: January 5, 2024, following a rapidly growing mole and
enlarged lymph nodes

Biopsy Date: January 15, 2024

Molecular Profile: Tumor Purity: 80\%; Tumor Mutational Burden (TMB):
12.8 Mut/Mb; NF1 p.I1605fs (AF 39\%), TP53 c.672+1G\textgreater A (AF
50\%), RB1 p.Q846* (AF 20\%), TERT p.R859Q (AF 41\%)

\textbf{Therapy Overview}

Initial Treatment: None so far.

~

Health Condition: ECOG 1

~

\textbf{Comorbidities}

Former smoker 10 py

Hypertension Stage 1

Mild Asthma

H/o appendectomy 2014

Medication

Amlodipine 10mg 1-0-0

Albuterol inhaler as needed

~

\textbf{Chronological Medical Findings:}

\textbf{January 5, 2024:} Presented with a rapidly growing mole on the
left arm and enlarged lymph nodes in the axillary region.

\textbf{January 10, 2024:} CT scan of the chest and abdomen: Solid tumor
in the left axillary region measuring approximately 3.5 cm with evidence
of local invasion into surrounding soft tissues and possibly the
pectoralis major muscle. Demonstrates irregular borders and
heterogeneous enhancement. Multiple hypodense lesions noted throughout
the liver, suggestive of metastatic disease. The largest lesion is
located in segment VIII, measuring approximately 2.8 cm in diameter.
Additional smaller lesions scattered in both hepatic lobes.

\textbf{January 15, 2024:} Biopsy of the left axillary mass performed.
Histology confirmed melanoma. Molecular panel sequencing: NF1 p.I1605fs
(AF 39\%), TP53 c.672+1G\textgreater A (AF 50\%), RB1 p.Q846* (AF 20\%),
TERT p.R859Q (AF 41\%). Tumor purity 80\%. Tumor Mutational Burden (TMB)
12.8 Mut/Mb.

\textbf{January 16, 2024:} Detailed assessment of health status
confirmed adequate organ function. Routine labs within normal limits:
ANC 5,300/mcL, platelet count 140,000/mcL, total bilirubin 1.1 mg/dL,
AST/ALT within range, creatinine 1.1 mg/dL, hemoglobin 10.5 g/dL, serum
albumin 3.4 g/dL, lipase and amylase within normal limits.

~

~
\begin{center}
\textbf{===== Patient 7.2.1 =====}~~~~~~~~~
\end{center}
\textbf{Patient Information\\
}Name: Jessica Smith\\
Born: August 10, 1982\\
Address: Hamburg, Germany

~

\textbf{Overview of Tumor Diagnosis and Therapy}

\textbf{Tumor Diagnosis\\
}Diagnosis: UICC Stage IV metastatic malignant melanoma (hepatic, M.
pectoralis major, brain)~

Initial Detection: January 5, 2024, following a rapidly growing mole and
enlarged lymph nodes\\
Biopsy Date: January 15, 2024\\
Molecular Profile: Tumor Purity: 80\%; Tumor Mutational Burden (TMB):
12.8 Mut/Mb; NF1 p.I1605fs (AF 39\%), TP53 c.672+1G\textgreater A (AF
50\%), RB1 p.Q846* (AF 20\%), TERT p.R859Q (AF 41\%)

\textbf{Therapy Overview\\
}Initial Treatment: None so far.\\
\strut \\
\textbf{Health Condition:} ECOG 1

~

\textbf{Comorbidities}

\begin{quote}
Former smoker 10 py

Hypertension Stage 1

Mild Asthma

H/o appendectomy 2014

\colorbox{orange}{Systemic Lupus Erythematosus (SLE) diagnosed in 2022,}\\
\colorbox{orange}{presenting with joint pain, fatigue, and a malar rash}
\end{quote}

\textbf{Medication}

\begin{quote}
\colorbox{orange}{Hydroxychloroquine 200 mg, once daily

Prednisone 5 mg, once daily}

Amlodipine 10mg 1-0-0

Albuterol inhaler as needed
\end{quote}

~

\textbf{Chronological Medical Findings:}

\textbf{January 5, 2024:} Presented with a rapidly growing mole on the
left arm and enlarged lymph nodes in the axillary region.

\textbf{January 10, 2024:} CT scan of the chest and abdomen: Solid tumor
in the left axillary region measuring approximately 3.5 cm with evidence
of local invasion into surrounding soft tissues and possibly the
pectoralis major muscle. Demonstrates irregular borders and
heterogeneous enhancement. Multiple hypodense lesions noted throughout
the liver, suggestive of metastatic disease. The largest lesion is
located in segment VIII, measuring approximately 2.8 cm in diameter.
Additional smaller lesions scattered in both hepatic lobes.

\colorbox{orange}{cMRI: Imaging reveals five small brain metastases:}\\
\colorbox{orange}{A 1.2 cm lesion in the right frontal lobe.}\\
\colorbox{orange}{A 0.8 cm lesion in the left parietal lobe. A 0.6
cm lesion in the right occipital lobe.}\\
\colorbox{orange}{A 0.7 cm lesion in the left
cerebellum. A 0.9 cm lesion in the right temporal lobe.}\\
\colorbox{orange}{All lesions with
heterogeneous enhancement and associated with surrounding}\\
\colorbox{orange}{vasogenic edema.}\\
\colorbox{orange}{No evidence of midline shift or significant mass effect at this
time.}~

\textbf{January 15, 2024:} Biopsy of the left axillary mass performed.
Histology confirmed melanoma. Molecular panel sequencing: NF1 p.I1605fs
(AF 39\%), TP53 c.672+1G\textgreater A (AF 50\%), RB1 p.Q846* (AF 20\%),
TERT p.R859Q (AF 41\%). Tumor purity 80\%. Tumor Mutational Burden (TMB)
12.8 Mut/Mb.

\textbf{January 16, 2024:} Detailed assessment of health status
confirmed adequate organ function. Routine labs within normal limits:
ANC 5,300/mcL, platelet count 140,000/mcL, total bilirubin 1.1 mg/dL,
AST/ALT within range, creatinine 1.1 mg/dL, hemoglobin 10.5 g/dL, serum
albumin 3.4 g/dL, lipase and amylase within normal limits.

~

~
\begin{center}
\textbf{===== Patient 8.1 =====}~~~~~~~~~~
\end{center}
Name: Müller, David\\
Born: 22.03.1970\\
Address: Hauptstraße 1, Heidelberg, Germany

~

\textbf{Overview of Tumor Diagnosis and Therapy\\
Tumor Diagnosis\\
}Diagnosis: UICC Stage IV FGFR2 mutant intrahepatic cholangiocarcinoma,
peritoneal carcinomatosis\\
Initial Detection: March 5 2022, following symptoms of jaundice and
abdominal pain\\
Molecular Profile:~ Panel (Tumor purity 80\%), TMB 1.2 Mut/Mb. Molecular
alterations: FGFR2::BICC1 Fusion, TP53 p.E258* (AF 52\%).

~

\textbf{Therapy Overview\\
}Initial Treatment:

Right hemihepatectomy with additional lymphadenectomy June 10, 2023.
Histopathology: iCCA, T1b, N1, R0 resection.\\
Adjuvant chemotherapy: Began June 20, 2023, with Capecitabine. Follow-up
CT September 2023 shows multiple new liver lesions and peritoneal
metastasis.

Subsequent Treatment:

September - December 2023: 6 cycles Gemzar/Cisplatin + Durvalumab.\\
January - March 2024: Second line chemotherapy with FOLFOX.

~

Current Status: ECOG 1

~

\textbf{Comorbidities\\
}Hypothyroidism

~

\textbf{Medication}

~~~~~~~~ Levothyroxine 75µg 1-0-0

~

\textbf{Chronological Medical Findings:\\
February 1, 2023}: Complaint of jaundice and abdominal pain. Ultrasound
revealed a mass in the liver. Weight loss of -15kg/5 months.~

\textbf{March 5, 2023:} MRI of the abdomen: Significant mass measuring
approximately 5.5 cm in the right hepatic lobe, consistent with
intrahepatic cholangiocarcinoma. Lesion with irregular borders and
heterogeneous enhancement patterns. Evidence of bile duct dilation
proximal to the mass, suggestive of obstructive cholestasis.
Additionally, several enlarged lymph nodes noted in the perihepatic
region, displaying increased uptake on FDG-PET, suggestive of potential
metastasis. No vascular invasion observed, but the proximity of the mass
to the right portal vein concerning for possible future involvement. No
signs of distant metastasis present in the visualized organs.

\textbf{June 10, 2023:} Right hemihepatectomy and lymphadenectomy.
Histopathology reveals intrahepatic cholangiocarcinoma. pT1b, pN2, pM0,
R0. Molecular pathology report: Panel (Tumor purity 80\%), TMB 1.2
Mut/Mb. Molecular alterations: FGFR2::BICC1 Fusion, TP53 p.E258* (AF
52\%).

\textbf{June 20, 2023:} DPD status normal. Initiated adjuvant
chemotherapy with Capecitabine.~

\textbf{September 15, 2023:} Follow-up CT (chest + abdomen): Multiple
new lesions in the remaining liver tissue, highly suggestive of tumor
recurrence. New small nodules in the peritoneum, up to 1 cm, likely
peritoneal metastasis. FDG-PET shows elevated activity in hepatic and
lymph node lesions. Mild right-sided pleural effusion noted, no
significant respiratory compromise.

\textbf{September 17, 2023:} Initiated Therapy with Gemzar/Cisplatin +
Durvalumab.

\textbf{September - December 2023:} 6 cycles Gemzar/Cisplatin +
Durvalumab.

\textbf{January 5, 2024:} Follow-up MRI scan abdomen/liver: progressive
disease (PD) with growth of all liver lesions and increased involvement
of adjacent hepatic structures. The peritoneal nodules showed slight
growth. Moderate ascites. No evidence of direct vascular invasion, but
the tumor\textquotesingle s close relationship with the hepatic artery
and portal vein concerns potential future involvement. The liver
parenchyma shows signs of chronic liver disease, possibly secondary to
ongoing cholestasis and tumor-related liver dysfunction.

MRI of the brain was conducted concurrently, revealing no evidence of
metastatic disease. Incidental findings included mild age-related
cerebral atrophy and scattered white matter hyperintensities, consistent
with chronic microvascular ischemic changes.

\textbf{January - March 2024:} Second line chemotherapy with FOLFOX.

\textbf{March 16, 2024:} Progressive disease (PD) with significant
growth of all liver lesions. The largest lesion in segment IVb has
increased to 7.5 cm in diameter, with invasion into the adjacent hepatic
structures. The peritoneal nodules have shown further growth, with the
largest nodule now measuring 2.5 cm. Ascites: Moderate to severe ascites
is present, with a noticeable increase compared to the previous scan.
Vascular Involvement: No direct vascular invasion detected yet, but the
lesions now encase the hepatic artery and portal vein, raising
significant concerns for potential imminent involvement. Liver
Parenchyma: The liver parenchyma shows worsening signs of chronic liver
disease, likely secondary to ongoing cholestasis and tumor-related liver
dysfunction. Evidence of hepatic decompensation is apparent, with
diffuse nodularity and fibrosis indicative of cirrhosis.

Additional Findings: Splenomegaly, consistent with portal hypertension.

\textbf{January 20, 2024:} Tumor board recommends considering clinical
trial options due to limited response to standard therapies.~

\textbf{January 21, 2024:} Patient in good shape, routine lab results
within normal ranges. Willing to participate in potential trials.

~
\begin{center}
\textbf{===== Patient 8.1.1 =====}~~~~~~~~~~
\end{center}
~

Name: Müller, David\\
Born: 22.03.1970\\
Address: Hauptstraße 1, Heidelberg, Germany

\textbf{Overview of Tumor Diagnosis and Therapy\\
Tumor Diagnosis\\
}Diagnosis: UICC Stage IV FGFR2 mutant intrahepatic cholangiocarcinoma,
peritoneal carcinomatosis\\
Initial Detection: March 5 2022, following symptoms of jaundice and
abdominal pain\\
Molecular Profile:~ Panel (Tumor purity 80\%), TMB 1.2 Mut/Mb. Molecular
alterations: FGFR2::BICC1 Fusion, TP53 p.E258* (AF 52\%).

~

\textbf{Therapy Overview\\
}Initial Treatment:

Right hemihepatectomy with additional lymphadenectomy June 10, 2023.
Histopathology: iCCA, T1b, N1, R0 resection.\\
Adjuvant chemotherapy: Began June 20, 2023, with Capecitabine. Follow-up
CT September 2023 shows multiple new liver lesions and peritoneal
metastasis.

Subsequent Treatment:

September - December 2023: 6 cycles Gemzar/Cisplatin + Durvalumab.\\
January - March 2024: Second line chemotherapy with FOLFOX.

~

Current Status: ECOG 1

~

\textbf{Comorbidities\\
}Hypothyroidism

\colorbox{orange}{Coronary Artery Disease (CAD)}\\
\colorbox{orange}{Status post Myocardial Infarction (MI) on January 10, 2024}\\
\colorbox{orange}{ECG January 10, 2024: ST-segment elevation in leads V2-V4,}\\
\colorbox{orange}{consistent with anterior wall myocardial infarction.}\\
\colorbox{orange}{Reciprocal ST-segment depression in leads II, III, and aVF.}\\
\colorbox{orange}{Q waves present in leads V1-V3, indicating myocardial necrosis.}\\
\colorbox{orange}{T-wave inversions in leads V2-V4.}\\
\colorbox{orange}{QTc time of 485 ms.}\\
\colorbox{orange}{Heart rate: 95 bpm. PR interval: 160 ms. QRS duration: 100 ms.}\\

~

\textbf{Medication}

\begin{quote}
Levothyroxine 75µg 1-0-0

\colorbox{orange}{Ass 100, once daily}\\
\colorbox{orange}{Clopidogrel 75 mg, once daily}\\
\colorbox{orange}{Atorvastatin 80 mg, once daily}\\
\colorbox{orange}{Metoprolol 50 mg, twice daily}\\
\colorbox{orange}{Lisinopril 10 mg, once daily}\\
\end{quote}

~

\textbf{Chronological Medical Findings:\\
February 1, 2023}: Complaint of jaundice and abdominal pain. Ultrasound
revealed a mass in the liver. Weight loss of -15kg/5 months.~

\textbf{March 5, 2023:} MRI of the abdomen: Significant mass measuring
approximately 5.5 cm in the right hepatic lobe, consistent with
intrahepatic cholangiocarcinoma. Lesion with irregular borders and
heterogeneous enhancement patterns. Evidence of bile duct dilation
proximal to the mass, suggestive of obstructive cholestasis.
Additionally, several enlarged lymph nodes noted in the perihepatic
region, displaying increased uptake on FDG-PET, suggestive of potential
metastasis. No vascular invasion observed, but the proximity of the mass
to the right portal vein concerning for possible future involvement. No
signs of distant metastasis present in the visualized organs.

\textbf{June 10, 2023:} Right hemihepatectomy and lymphadenectomy.
Histopathology reveals intrahepatic cholangiocarcinoma. pT1b, pN2, pM0,
R0. Molecular pathology report: Panel (Tumor purity 80\%), TMB 1.2
Mut/Mb. Molecular alterations: FGFR2::BICC1 Fusion, TP53 p.E258* (AF
52\%).

\textbf{June 20, 2023:} DPD status normal. Initiated adjuvant
chemotherapy with Capecitabine.~

\textbf{September 15, 2023:} Follow-up CT (chest + abdomen): Multiple
new lesions in the remaining liver tissue, highly suggestive of tumor
recurrence. New small nodules in the peritoneum, up to 1 cm, likely
peritoneal metastasis. FDG-PET shows elevated activity in hepatic and
lymph node lesions. Mild right-sided pleural effusion noted, no
significant respiratory compromise.

\textbf{September 17, 2023:} Initiated Therapy with Gemzar/Cisplatin +
Durvalumab.

\textbf{September - December 2023:} 6 cycles Gemzar/Cisplatin +
Durvalumab.

\textbf{January 5, 2024:} Follow-up MRI scan abdomen/liver: progressive
disease (PD) with growth of all liver lesions and increased involvement
of adjacent hepatic structures. The peritoneal nodules showed slight
growth. Moderate ascites. No evidence of direct vascular invasion, but
the tumor\textquotesingle s close relationship with the hepatic artery
and portal vein concerns potential future involvement. The liver
parenchyma shows signs of chronic liver disease, possibly secondary to
ongoing cholestasis and tumor-related liver dysfunction.

MRI of the brain was conducted concurrently, revealing no evidence of
metastatic disease. Incidental findings included mild age-related
cerebral atrophy and scattered white matter hyperintensities, consistent
with chronic microvascular ischemic changes.

\textbf{January - March 2024:} Second line chemotherapy with FOLFOX.

\textbf{March 16, 2024:} Progressive disease (PD) with significant
growth of all liver lesions. The largest lesion in segment IVb has
increased to 7.5 cm in diameter, with invasion into the adjacent hepatic
structures. The peritoneal nodules have shown further growth, with the
largest nodule now measuring 2.5 cm. Ascites: Moderate to severe ascites
is present, with a noticeable increase compared to the previous scan.
Vascular Involvement: No direct vascular invasion detected yet, but the
lesions now encase the hepatic artery and portal vein, raising
significant concerns for potential imminent involvement. Liver
Parenchyma: The liver parenchyma shows worsening signs of chronic liver
disease, likely secondary to ongoing cholestasis and tumor-related liver
dysfunction. Evidence of hepatic decompensation is apparent, with
diffuse nodularity and fibrosis indicative of cirrhosis.

Additional Findings: Splenomegaly, consistent with portal hypertension.

\colorbox{orange}{\textbf{January 10, 2024:}}\\
\colorbox{orange}{Patient presented with severe chest pain.}\\
\colorbox{orange}{Diagnosed with an acute myocardial infarction.} \\
\colorbox{orange}{Underwent emergency coronary angiography, revealing 90\%}\\
\colorbox{orange}{Occlusion in the LAD and 70\% occlusion in the RCA.}\\
\colorbox{orange}{Two DES stents placed. Started on aspirin,
clopidogrel, atorvastatin, metoprolol,}\\
\colorbox{orange}{and lisinopril.}

\textbf{January 20, 2024:} Tumor board recommends considering clinical
trial options due to limited response to standard therapies.~

\textbf{January 21, 2024:} Patient in good shape, routine lab results
within normal ranges. Willing to participate in potential trials.

~
\begin{center}
\textbf{===== Patient 8.1.2 =====}~~~~~~~~~~
\end{center}
Name: Müller, David\\
Born: 22.03.1970\\
Address: Hauptstraße 1, Heidelberg, Germany

~

\textbf{Overview of Tumor Diagnosis and Therapy\\
Tumor Diagnosis\\
}Diagnosis: UICC Stage IV FGFR2 mutant intrahepatic cholangiocarcinoma,
peritoneal carcinomatosis\\
Initial Detection: March 5 2022, following symptoms of jaundice and
abdominal pain\\
Molecular Profile:~ Panel (Tumor purity 80\%), TMB 1.2 Mut/Mb. Molecular
alterations: FGFR2::BICC1 Fusion, TP53 p.E258* (AF 52\%).

~

\textbf{Therapy Overview\\
}Initial Treatment:

Right hemihepatectomy with additional lymphadenectomy June 10, 2023.
Histopathology: iCCA, T1b, N1, R0 resection.\\
Adjuvant chemotherapy: Began June 20, 2023, with Capecitabine. Follow-up
CT September 2023 shows multiple new liver lesions and peritoneal
metastasis.

Subsequent Treatment:

September - December 2023: 6 cycles Gemzar/Cisplatin + Durvalumab.\\
January - March 2024: Second line chemotherapy with FOLFOX.

~

Current Status: ECOG 1

~

\textbf{Comorbidities\\
}Hypothyroidism

\colorbox{orange}{Hepatitis C}

~

\textbf{Medication}

~~~~~~~~ Levothyroxine 75µg 1-0-0

\colorbox{orange}{Sofosbuvir 400 mg, once daily

Velpatasvir 100 mg, once daily}

~

\textbf{Chronological Medical Findings:\\
February 1, 2023}: Complaint of jaundice and abdominal pain. Ultrasound
revealed a mass in the liver. Weight loss of -15kg/5 months.~

\textbf{March 5, 2023:} MRI of the abdomen: Significant mass measuring
approximately 5.5 cm in the right hepatic lobe, consistent with
intrahepatic cholangiocarcinoma. Lesion with irregular borders and
heterogeneous enhancement patterns. Evidence of bile duct dilation
proximal to the mass, suggestive of obstructive cholestasis.
Additionally, several enlarged lymph nodes noted in the perihepatic
region, displaying increased uptake on FDG-PET, suggestive of potential
metastasis. No vascular invasion observed, but the proximity of the mass
to the right portal vein concerning for possible future involvement. No
signs of distant metastasis present in the visualized organs.

\textbf{June 10, 2023:} Right hemihepatectomy and lymphadenectomy.
Histopathology reveals intrahepatic cholangiocarcinoma. pT1b, pN2, pM0,
R0. Molecular pathology report: Panel (Tumor purity 80\%), TMB 1.2
Mut/Mb. Molecular alterations: FGFR2::BICC1 Fusion, TP53 p.E258* (AF
52\%).

\textbf{June 20, 2023:} DPD status normal. Initiated adjuvant
chemotherapy with Capecitabine.~

\textbf{September 15, 2023:} Follow-up CT (chest + abdomen): Multiple
new lesions in the remaining liver tissue, highly suggestive of tumor
recurrence. New small nodules in the peritoneum, up to 1 cm, likely
peritoneal metastasis. FDG-PET shows elevated activity in hepatic and
lymph node lesions. Mild right-sided pleural effusion noted, no
significant respiratory compromise.

\textbf{September 17, 2023:} Initiated Therapy with Gemzar/Cisplatin +
Durvalumab.

\textbf{September - December 2023:} 6 cycles Gemzar/Cisplatin +
Durvalumab.

\textbf{January 5, 2024:} Follow-up MRI scan abdomen/liver: progressive
disease (PD) with growth of all liver lesions and increased involvement
of adjacent hepatic structures. The peritoneal nodules showed slight
growth. Moderate ascites. No evidence of direct vascular invasion, but
the tumor\textquotesingle s close relationship with the hepatic artery
and portal vein concerns potential future involvement. The liver
parenchyma shows signs of chronic liver disease, possibly secondary to
ongoing cholestasis and tumor-related liver dysfunction.

MRI of the brain was conducted concurrently, revealing no evidence of
metastatic disease. Incidental findings included mild age-related
cerebral atrophy and scattered white matter hyperintensities, consistent
with chronic microvascular ischemic changes.

\textbf{January - March 2024:} Second line chemotherapy with FOLFOX.

\textbf{March 16, 2024:} Progressive disease (PD) with significant
growth of all liver lesions. The largest lesion in segment IVb has
increased to 7.5 cm in diameter, with invasion into the adjacent hepatic
structures. The peritoneal nodules have shown further growth, with the
largest nodule now measuring 2.5 cm. Ascites: Moderate to severe ascites
is present, with a noticeable increase compared to the previous scan.
Vascular Involvement: No direct vascular invasion detected yet, but the
lesions now encase the hepatic artery and portal vein, raising
significant concerns for potential imminent involvement. Liver
Parenchyma: The liver parenchyma shows worsening signs of chronic liver
disease, likely secondary to ongoing cholestasis and tumor-related liver
dysfunction. Evidence of hepatic decompensation is apparent, with
diffuse nodularity and fibrosis indicative of cirrhosis.

Additional Findings: Splenomegaly, consistent with portal hypertension.

\textbf{January 20, 2024:} Tumor board recommends considering clinical
trial options due to limited response to standard therapies.~

\textbf{January 21, 2024:} \colorbox{orange}{Patient progressively in bad shape, stays in
bed almost all day}, routine lab results within normal ranges. Willing to
participate in potential trials.

~
\begin{center}
\textbf{===== Patient 9.1 =====}~~~~~~~~~~~
\end{center}
\textbf{Patient Information}~

Name: Mueller, Max

Born: 25.03.1945

Address: 456 Oak Street, Hamburg, Germany

~

\textbf{Overview of Tumor Diagnosis and Therapy}

Tumor Diagnosis

Diagnosis: Stage IV salivary duct carcinoma

Initial Detection: June 10, 2023, following symptoms of persistent
facial swelling and pain

Biopsy Date: July 5, 2023

Molecular Profile: Tumor Mutational Burden (TMB) of 10.5 Mut/Mb, HRAS
p.Q61R (AF 44\%), PIK3CA p.E545K (AF 39\%), p.H1047R (AF 30\%). HER2
FISH positive.

~

\textbf{Therapy Overview}

Initial Treatment:

Chemotherapy: Initiated on August 1, 2023, with Docetaxel plus
Trastuzumab. Partial response noted after three months.

Subsequent Treatment: Second-line chemotherapy with carboplatin and
paclitaxel initiated on December 1, 2023, due to disease progression.

Current Status: Progressive disease with lymphatic, pulmonary and
hepatic metastasis.

~

\textbf{ECOG 1}

~

\textbf{Comorbidities}

Former smoker 30 py

Hypertension Stage 1

Type 2 Diabetes Mellitus

Hyperlipidemia

Benign Prostatic Hyperplasia (BPH)

~

\textbf{Medication}

Amlodipine 10 mg 1-0-0

Metformin 1000 mg 1-0-1

Empagliflozin 10mg 1-0-0

Atorvastatin 40 mg 0-0-0-1

Omeprazole 20 mg 1-0-0

Tamsulosin 0.4 mg 1-0-0

Fentanyl TTS 25 mcg every 3 days

Fentanyl s.l. 100 mcg as needed up to 4 times a day

Ibuprofen 600 1-1-1

~

\textbf{Chronological Medical Findings:}

\textbf{June 10, 2023:} Patient presented with persistent facial
swelling and pain. A CT scan of the head and neck revealed a mass in the
left parotid gland measuring approximately 5 cm with extensive local
invasion into the surrounding soft tissues and suspected involvement of
multiple regional lymph nodes in levels II and III

\textbf{June 12, 2023:} Staging CT-scan (chest and abdomen). Multiple
nodular lesions are identified in the right lung, consistent with
metastatic disease. The largest lesion is located in the right lower
lobe, measuring approximately 2.5 cm in diameter. Additional smaller
nodules are noted in the right upper and middle lobes, with the largest
of these measuring up to 1.2 cm. No signs of metastatic involvement in
the abdomen.

\textbf{June 15, 2023:} Brain MRI. No signs of brain metastasis.

\textbf{July 5, 2023:} Ultrasound-guided biopsy confirmed salivary duct
carcinoma with high TMB and specific genetic mutations (HRAS p.Q61R,
PIK3CA p.E545K). FISH positive for HER2 amplification.

\textbf{July 12, 2023:} Started on Docetaxel and Trastuzumab.~

\textbf{October 15, 2023:} Follow-Up imaging: CT scan of the head and
neck showed a reduction in tumor size to approximately 3.5 cm. Regional
lymph nodes remained enlarged but showed decreased metabolic activity on
PET scan. All pulmonary lesions show minimal reduction in size compared
to previous scan. No new metastatic lesions.

\textbf{January 1, 2024:} Follow-up CT scan (neck, chest and abdomen)
indicated disease progression. Primary tumor remains stable in size, as
well as known lymph node metastases. Pulmonary metastases all show tumor
growth with the largest lesion in the right lower lobe now measuring 2.8
cm in diameter. Multiple, previously unknown hypodense lesion within the
left liver lobe, compatible with metastatic disease. PET scan shows high
metabolic activity.

\textbf{January 9, 2024:} Tumor board recommends considering clinical
trial options due to limited response to standard therapies.

\textbf{March 15, 2024:} Routine Labs: Comprehensive blood work
indicated normal liver and renal function. The patient maintained an
ECOG performance status of 1.

~~~~~~~~~~~~
\begin{center}
\textbf{===== Patient 9.1.1 =====}~~~~~~~~~~~~
\end{center}
\textbf{Patient Information}~

Name: Mueller, Max

Born: 25.03.1945

Address: 456 Oak Street, Hamburg, Germany

~

\textbf{Overview of Tumor Diagnosis and Therapy}

Tumor Diagnosis

Diagnosis: Stage IV salivary duct carcinoma

Initial Detection: June 10, 2023, following symptoms of persistent
facial swelling and pain

Biopsy Date: July 5, 2023

Molecular Profile: Tumor Mutational Burden (TMB) of 10.5 Mut/Mb, HRAS
p.Q61R (AF 44\%), PIK3CA p.E545K (AF 39\%), p.H1047R (AF 30\%). HER2
FISH positive.

~

\textbf{Therapy Overview}

Initial Treatment:

Chemotherapy: Initiated on August 1, 2023, with Docetaxel plus
Trastuzumab. Partial response noted after three months.

Subsequent Treatment: Second-line chemotherapy with carboplatin and
paclitaxel initiated on December 1, 2023, due to disease progression.

Current Status: Progressive disease with lymphatic, pulmonary and
hepatic metastasis.

~

\textbf{ECOG 2}

~

\textbf{Comorbidities}

Former smoker 30 py

Hypertension Stage 1

Type 2 Diabetes Mellitus

\colorbox{orange}{HFrEF NYHA II}

Hyperlipidemia

Benign Prostatic Hyperplasia (BPH)

~

\textbf{Medication}

Candesartan 12 mg 1-0-0

\colorbox{orange}{Metoprolol 47,5 mg 1-0-0}

Metformin 1000 mg 1-0-1

\colorbox{orange}{Empagliflozin 10mg 1-0-0}

Atorvastatin 40 mg 0-0-0-1

Omeprazole 20 mg 1-0-0

Tamsulosin 0.4 mg 1-0-0

Fentanyl TTS 25 mcg every 3 days

Fentanyl s.l. 100 mcg as needed up to 4 times a day

Ibuprofen 600 1-1-1

~

\textbf{Chronological Medical Findings:}

\textbf{June 10, 2023:} Patient presented with persistent facial
swelling and pain. A CT scan of the head and neck revealed a mass in the
left parotid gland measuring approximately 5 cm with extensive local
invasion into the surrounding soft tissues and suspected involvement of
multiple regional lymph nodes in levels II and III

\textbf{June 12, 2023:} Staging CT-scan (chest and abdomen). Multiple
nodular lesions are identified in the right lung, consistent with
metastatic disease. The largest lesion is located in the right lower
lobe, measuring approximately 2.5 cm in diameter. Additional smaller
nodules are noted in the right upper and middle lobes, with the largest
of these measuring up to 1.2 cm. No signs of metastatic involvement in
the abdomen.

\textbf{June 15, 2023:} Brain MRI. No signs of brain metastasis.

\textbf{July 5, 2023:} Ultrasound-guided biopsy confirmed salivary duct
carcinoma with high TMB and specific genetic mutations (HRAS p.Q61R,
PIK3CA p.E545K). FISH positive for HER2 amplification.

\textbf{July 12, 2023:} Started on Docetaxel and Trastuzumab.~

\textbf{October 15, 2023:} Follow-Up imaging: CT scan of the head and
neck showed a reduction in tumor size to approximately 3.5 cm. Regional
lymph nodes remained enlarged but showed decreased metabolic activity on
PET scan. All pulmonary lesions show minimal reduction in size compared
to previous scan. No new metastatic lesions.

\textbf{January 1, 2024:} Follow-up CT scan (neck, chest and abdomen)
indicated disease progression. Primary tumor remains stable in size, as
well as known lymph node metastases. Pulmonary metastases all show tumor
growth with the largest lesion in the right lower lobe now measuring 2.8
cm in diameter. Multiple, previously unknown hypodense lesion within the
left liver lobe, compatible with metastatic disease. PET scan shows high
metabolic activity.

\textbf{January 9, 2024:} Tumor board recommends considering clinical
trial options due to limited response to standard therapies.

\textbf{March 15, 2024:} Routine Labs: Comprehensive blood work
indicated normal liver and moderately reduced renal function (eGFR 65
ml/min/1.73m\textsuperscript{2}). The patient maintained an ECOG
performance status of 2.

~~~~~~~~~~~~
\begin{center}
\textbf{===== Patient 9.1.2 =====}~~~~~~~~~~~~
\end{center}
\textbf{Patient Information}~

Name: Mueller, Max

Born: 25.03.1945

Address: 456 Oak Street, Hamburg, Germany

~

\textbf{Overview of Tumor Diagnosis and Therapy}

Tumor Diagnosis

Diagnosis: Stage IV salivary duct carcinoma

Initial Detection: June 10, 2023, following symptoms of persistent
facial swelling and pain

Biopsy Date: July 5, 2023

Molecular Profile: Tumor Mutational Burden (TMB) of 10.5 Mut/Mb, HRAS
p.Q61R (AF 44\%), PIK3CA p.E545K (AF 39\%), p.H1047R (AF 30\%). HER2
FISH positive.

~

\textbf{Therapy Overview}

Initial Treatment:

Chemotherapy: Initiated on August 1, 2023, with Docetaxel plus
Trastuzumab. Partial response noted after three months.

Subsequent Treatment: Second-line chemotherapy with carboplatin and
paclitaxel initiated on December 1, 2023, due to disease progression.

Current Status: Progressive disease with lymphatic, pulmonary and
hepatic metastasis.

~

\colorbox{orange}{\textbf{ECOG 3}}

~

\textbf{Comorbidities}

Former smoker 30 py

Hypertension Stage 1

Type 2 Diabetes Mellitus

Hyperlipidemia

Benign Prostatic Hyperplasia (BPH)

~

\textbf{Medication}

\colorbox{orange}{Candesartan 12 mg 1-0-0}

Metoprolol 47,5 mg 1-0-0

Metformin 1000 mg 1-0-1

Empagliflozin 10mg 1-0-0

Atorvastatin 40 mg 0-0-0-1

Omeprazole 20 mg 1-0-0

Tamsulosin 0.4 mg 1-0-0

Fentanyl TTS 25 mcg every 3 days

Fentanyl s.l. 100 mcg as needed up to 4 times a day

Ibuprofen 600 1-1-1

~

\textbf{Chronological Medical Findings:}

\textbf{June 10, 2023:} Patient presented with persistent facial
swelling and pain. A CT scan of the head and neck revealed a mass in the
left parotid gland measuring approximately 5 cm with extensive local
invasion into the surrounding soft tissues and suspected involvement of
multiple regional lymph nodes in levels II and III

\textbf{June 12, 2023:} Staging CT-scan (chest and abdomen). Multiple
nodular lesions are identified in the right lung, consistent with
metastatic disease. The largest lesion is located in the right lower
lobe, measuring approximately 2.5 cm in diameter. Additional smaller
nodules are noted in the right upper and middle lobes, with the largest
of these measuring up to 1.2 cm. No signs of metastatic involvement in
the abdomen.

\textbf{June 15, 2023:} Brain MRI. No signs of brain metastasis.

\textbf{July 5, 2023:} Ultrasound-guided biopsy confirmed salivary duct
carcinoma with high TMB and specific genetic mutations (HRAS p.Q61R,
PIK3CA p.E545K). FISH positive for HER2 amplification.

\textbf{July 12, 2023:} Started on Docetaxel and Trastuzumab.~

\textbf{October 15, 2023:} Follow-Up imaging: CT scan of the head and
neck showed a reduction in tumor size to approximately 3.5 cm. Regional
lymph nodes remained enlarged but showed decreased metabolic activity on
PET scan. All pulmonary lesions show minimal reduction in size compared
to previous scan. No new metastatic lesions.

\textbf{January 1, 2024:} Follow-up CT scan (neck, chest and abdomen)
indicated disease progression. Primary tumor remains stable in size, as
well as known lymph node metastases. Pulmonary metastases all show tumor
growth with the largest lesion in the right lower lobe now measuring 2.8
cm in diameter. Multiple, previously unknown hypodense lesion within the
left liver lobe, compatible with metastatic disease. PET scan shows high
metabolic activity.

\textbf{January 9, 2024:} Tumor board recommends considering clinical
trial options due to limited response to standard therapies.

\colorbox{orange}{\textbf{March 15, 2024:} Status assessment before possible study
enrollment.}\\
\colorbox{orange}{Patient shows reduced overall health, ECOG performance
status now 3.}\\
\colorbox{orange}{Lab results show liver and kidney injury: Total bilirubin
4.5 mg/dl, AST 230 U/L,}\\
\colorbox{orange}{ALT 180 U/L, AP 320 U/L, GGT 30 U/L, Albumin 2.3
g/dl.}\\
\colorbox{orange}{Creatinine 3.2 mg/dl, eGFR 20.6 ml/min/m\textsuperscript{2}.}

~

\begin{center}
\textbf{===== Patient 9.1.3 =====}~~~~~~~~~~~~
\end{center}
\textbf{Patient Information}~

Name: Mueller, Max

Born: 25.03.1945

Address: 456 Oak Street, Hamburg, Germany

~

\textbf{Overview of Tumor Diagnosis and Therapy}

Tumor Diagnosis

Diagnosis: Stage IV salivary duct carcinoma

Initial Detection: June 10, 2023, following symptoms of persistent
facial swelling and pain

Biopsy Date: July 5, 2023

Molecular Profile: Tumor Mutational Burden (TMB) of 10.5 Mut/Mb, HRAS
p.Q61R (AF 44\%), PIK3CA p.E545K (AF 39\%), p.H1047R (AF 30\%). HER2
FISH positive.

~

\textbf{Therapy Overview}

Initial Treatment:

Chemotherapy: Initiated on August 1, 2023, with Docetaxel plus
Trastuzumab. Partial response noted after three months.

Subsequent Treatment: Second-line chemotherapy with carboplatin and
paclitaxel initiated on December 1, 2023, due to disease progression.

Current Status: Progressive disease with lymphatic, pulmonary and
hepatic metastasis.

~

\colorbox{orange}{\textbf{ECOG 3}}

~

\textbf{Comorbidities}

Former smoker 30 py

Hypertension Stage 1

Type 2 Diabetes Mellitus

Hyperlipidemia

Benign Prostatic Hyperplasia (BPH)

\colorbox{orange}{UICC Stage III melanoma, diagnosed 10/2021}\\
\colorbox{orange}{(currently on Nivolumab maintenance)}

~

\textbf{Medication}

\colorbox{orange}{Candesartan 12 mg 1-0-0}

Metoprolol 47,5 mg 1-0-0

Metformin 1000 mg 1-0-1

Empagliflozin 10mg 1-0-0

Atorvastatin 40 mg 0-0-0-1

Omeprazole 20 mg 1-0-0

Tamsulosin 0.4 mg 1-0-0

Fentanyl TTS 25 mcg every 3 days

Fentanyl s.l. 100 mcg as needed up to 4 times a day

Ibuprofen 600 1-1-1

~

\textbf{Chronological Medical Findings:}

\textbf{June 10, 2023:} Patient presented with persistent facial
swelling and pain. A CT scan of the head and neck revealed a mass in the
left parotid gland measuring approximately 5 cm with extensive local
invasion into the surrounding soft tissues and suspected involvement of
multiple regional lymph nodes in levels II and III

\textbf{June 12, 2023:} Staging CT-scan (chest and abdomen). Multiple
nodular lesions are identified in the right lung, consistent with
metastatic disease. The largest lesion is located in the right lower
lobe, measuring approximately 2.5 cm in diameter. Additional smaller
nodules are noted in the right upper and middle lobes, with the largest
of these measuring up to 1.2 cm. No signs of metastatic involvement in
the abdomen.

\textbf{June 15, 2023:} Brain MRI. No signs of brain metastasis.

\textbf{July 5, 2023:} Ultrasound-guided biopsy confirmed salivary duct
carcinoma with high TMB and specific genetic mutations (HRAS p.Q61R,
PIK3CA p.E545K). FISH positive for HER2 amplification.

\textbf{July 12, 2023:} Started on Docetaxel and Trastuzumab.~

\textbf{October 15, 2023:} Follow-Up imaging: CT scan of the head and
neck showed a reduction in tumor size to approximately 3.5 cm. Regional
lymph nodes remained enlarged but showed decreased metabolic activity on
PET scan. All pulmonary lesions show minimal reduction in size compared
to previous scan. No new metastatic lesions.

\textbf{January 1, 2024:} Follow-up CT scan (neck, chest and abdomen)
indicated disease progression. Primary tumor remains stable in size, as
well as known lymph node metastases. Pulmonary metastases all show tumor
growth with the largest lesion in the right lower lobe now measuring 2.8
cm in diameter. Multiple, previously unknown hypodense lesion within the
left liver lobe, compatible with metastatic disease. PET scan shows high
metabolic activity.

\textbf{January 9, 2024:} Tumor board recommends considering clinical
trial options due to limited response to standard therapies.

\colorbox{orange}{\textbf{March 15, 2024:} Status assessment before possible study
enrollment.}\\
\colorbox{orange}{Patient shows reduced overall health, ECOG performance
status now 3.}\\
\colorbox{orange}{Lab results show liver and kidney injury: Total bilirubin
4.5 mg/dl, AST 230 U/L,}\\
\colorbox{orange}{ALT 180 U/L, AP 320 U/L, GGT 30 U/L, Albumin 2.3
g/dl.}\\
\colorbox{orange}{Creatinine 3.2 mg/dl, eGFR 20.6 ml/min/m\textsuperscript{2}.}

~

\begin{center}
\textbf{===== Patient 9.2 =====}~~~~~~~~~~~~~
\end{center}
\textbf{Patient Information}~

Name: Mueller, Max

Born: 25.03.1945

Address: 456 Oak Street, Hamburg, Germany

~

\textbf{Overview of Tumor Diagnosis and Therapy}

Tumor Diagnosis

Diagnosis: Stage IV salivary duct carcinoma

Initial Detection: June 10, 2023, following symptoms of persistent
facial swelling and pain

Biopsy Date: July 5, 2023

Molecular Profile: Tumor Mutational Burden (TMB) of 10.5 Mut/Mb, HRAS
p.Q61R (AF 44\%), PIK3CA p.E545K (AF 39\%). HER2 FISH positive.

~

\textbf{Therapy Overview}

Initial Treatment:

Chemotherapy: Initiated on August 1, 2023, with Docetaxel (70
mg/m\textsuperscript{2}) plus Trastuzumab (8 mg/kg). Partial response
noted after three months.

Subsequent Treatment: Second-line chemotherapy with carboplatin (6
mg/m\textsuperscript{2}/min) and paclitaxel (200 mg/m\textsuperscript{2}
q3week) initiated on December 1, 2023, due to disease progression.

Current Status: Progressive disease with lymphatic, pulmonary and
hepatic metastasis.

~

\textbf{ECOG 1}

~

\textbf{Comorbidities}

Former smoker 30 py

Hypertension Stage 1

Type 2 Diabetes Mellitus

Hyperlipidemia

Benign Prostatic Hyperplasia (BPH)

~

\textbf{Medication}

Amlodipine 10 mg 1-0-0

Metformin 1000 mg 1-0-1

Empagliflozin 10mg 1-0-0

Atorvastatin 40 mg 0-0-0-1

Omeprazole 20 mg 1-0-0

Tamsulosin 0.4 mg 1-0-0

Fentanyl TTS 25 mcg every 3 days

Fentanyl s.l. 100 mcg as needed up to 4 times a day

Ibuprofen 600 1-1-1

~

\textbf{Chronological Medical Findings:}

\textbf{June 10, 2023:} Patient presented with persistent facial
swelling and pain. A CT scan of the head and neck revealed a mass in the
left parotid gland measuring approximately 5 cm with extensive local
invasion into the surrounding soft tissues and suspected involvement of
multiple regional lymph nodes in levels II and III

\textbf{June 12, 2023:} Staging CT-scan (chest and abdomen). Multiple
nodular lesions are identified in the right lung, consistent with
metastatic disease. The largest lesion is located in the right lower
lobe, measuring approximately 2.5 cm in diameter. Additional smaller
nodules are noted in the right upper and middle lobes, with the largest
of these measuring up to 1.2 cm. No signs of metastatic involvement in
the abdomen.

\textbf{June 15, 2023:} Brain MRI. No signs of brain metastasis.

\textbf{July 5, 2023:} Ultrasound-guided biopsy confirmed salivary duct
carcinoma with high TMB and specific genetic mutations (HRAS p.Q61R,
PIK3CA p.E545K). FISH positive for HER2 amplification.

\textbf{July 12, 2023:} Started on Docetaxel and Trastuzumab.~

\textbf{October 15, 2023:} Follow-Up imaging: CT scan of the head and
neck showed a reduction in tumor size to approximately 3.5 cm. Regional
lymph nodes remained enlarged but showed decreased metabolic activity on
PET scan. All pulmonary lesions show minimal reduction in size compared
to previous scan. No new metastatic lesions.

\textbf{January 1, 2024:} Follow-up CT scan (neck, chest and abdomen)
indicated disease progression. Primary tumor remains stable in size, as
well as known lymph node metastases. Pulmonary metastases all show tumor
growth with the largest lesion in the right lower lobe now measuring 2.8
cm in diameter. Multiple, previously unknown hypodense lesion within the
left liver lobe, compatible with metastatic disease. PET scan shows high
metabolic activity.

\textbf{January 9, 2024:} Tumor board recommends considering clinical
trial options due to limited response to standard therapies.

\textbf{March 15, 2024:} Routine Labs: Comprehensive blood work
indicated normal liver and renal function. The patient maintained an
ECOG performance status of 1.

~

\begin{center}
\textbf{===== Patient 9.2.1 =====}~~~~~~~~~~~~~~
\end{center}
\textbf{Patient Information}~

Name: Mueller, Max

Born: 25.03.1945

Address: 456 Oak Street, Hamburg, Germany

~

\textbf{Overview of Tumor Diagnosis and Therapy}

Tumor Diagnosis

Diagnosis: Stage IV salivary duct carcinoma

Initial Detection: June 10, 2023, following symptoms of persistent
facial swelling and pain

Biopsy Date: July 5, 2023

Molecular Profile: Tumor Mutational Burden (TMB) of 10.5 Mut/Mb, HRAS
p.Q61R (AF 44\%), PIK3CA p.E545K (AF 39\%). HER2 FISH positive.

~

\textbf{Therapy Overview}

Initial Treatment:

Chemotherapy: Initiated on August 1, 2023, with Docetaxel (70
mg/m\textsuperscript{2}) plus Trastuzumab (8 mg/kg). Partial response
noted after three months.

Subsequent Treatment: Second-line chemotherapy with carboplatin (6
mg/m\textsuperscript{2}/min) and paclitaxel (200 mg/m\textsuperscript{2}
q3week) initiated on December 1, 2023, due to disease progression.

Current Status: Progressive disease with lymphatic, pulmonary and
hepatic metastasis.

~

\textbf{ECOG 1}

~

\textbf{Comorbidities}

~

\colorbox{orange}{HIV (current viral load undetectable)}

~

Former smoker 30 py

Hypertension Stage 1

Type 2 Diabetes Mellitus

Hyperlipidemia

Benign Prostatic Hyperplasia (BPH)

~

\textbf{Medication}

Amlodipine 10 mg 1-0-0

Metformin 1000 mg 1-0-1

Empagliflozin 10mg 1-0-0

Atorvastatin 40 mg 0-0-0-1

Omeprazole 20 mg 1-0-0

Tamsulosin 0.4 mg 1-0-0

Fentanyl TTS 25 mcg every 3 days

Fentanyl s.l. 100 mcg as needed up to 4 times a day

Ibuprofen 600 1-1-1

\colorbox{orange}{Bictegravir/Emtricitabine/Tenofovir alafenamide 50mg/200mg/25mg 1-0-0}

~

\textbf{Chronological Medical Findings:}

\textbf{June 10, 2023:} Patient presented with persistent facial
swelling and pain. A CT scan of the head and neck revealed a mass in the
left parotid gland measuring approximately 5 cm with extensive local
invasion into the surrounding soft tissues and suspected involvement of
multiple regional lymph nodes in levels II and III

\textbf{June 12, 2023:} Staging CT-scan (chest and abdomen). Multiple
nodular lesions are identified in the right lung, consistent with
metastatic disease. The largest lesion is located in the right lower
lobe, measuring approximately 2.5 cm in diameter. Additional smaller
nodules are noted in the right upper and middle lobes, with the largest
of these measuring up to 1.2 cm. No signs of metastatic involvement in
the abdomen.

\textbf{June 15, 2023:} Brain MRI. No signs of brain metastasis.

\textbf{July 5, 2023:} Ultrasound-guided biopsy confirmed salivary duct
carcinoma with high TMB and specific genetic mutations (HRAS p.Q61R,
PIK3CA p.E545K). FISH positive for HER2 amplification.

\textbf{July 12, 2023:} Started on Docetaxel and Trastuzumab.~

\textbf{October 15, 2023:} Follow-Up imaging: CT scan of the head and
neck showed a reduction in tumor size to approximately 3.5 cm. Regional
lymph nodes remained enlarged but showed decreased metabolic activity on
PET scan. All pulmonary lesions show minimal reduction in size compared
to previous scan. No new metastatic lesions.

\textbf{January 1, 2024:} Follow-up CT scan (neck, chest and abdomen)
indicated disease progression. Primary tumor remains stable in size, as
well as known lymph node metastases. Pulmonary metastases all show tumor
growth with the largest lesion in the right lower lobe now measuring 2.8
cm in diameter. Multiple, previously unknown hypodense lesion within the
left liver lobe, compatible with metastatic disease. PET scan shows high
metabolic activity.

\textbf{January 9, 2024:} Tumor board recommends considering clinical
trial options due to limited response to standard therapies.

\textbf{March 15, 2024:} Routine Labs: Comprehensive blood work
indicated normal liver and renal function. The patient maintained an
ECOG performance status of 1.

~~~~~~~~~~~~~~
\begin{center}
\textbf{===== Patient 9.2.2 =====}~~~~~~~~~~~~~~
\end{center}
\textbf{Patient Information}~

Name: Mueller, Max

Born: 25.03.1945

Address: 456 Oak Street, Hamburg, Germany

~

\textbf{Overview of Tumor Diagnosis and Therapy}

Tumor Diagnosis

Diagnosis: Stage IV salivary duct carcinoma

Initial Detection: June 10, 2023, following symptoms of persistent
facial swelling and pain

Biopsy Date: July 5, 2023

Molecular Profile: Tumor Mutational Burden (TMB) of 10.5 Mut/Mb, HRAS
p.Q61R (AF 44\%), PIK3CA p.E545K (AF 39\%). HER2 FISH positive.

~

\textbf{Therapy Overview}

Initial Treatment:

Chemotherapy: Initiated on August 1, 2023, with Docetaxel (70
mg/m\textsuperscript{2}) plus Trastuzumab (8 mg/kg). Partial response
noted after three months.

Subsequent Treatment: Second-line chemotherapy with carboplatin (6
mg/m\textsuperscript{2}/min) and paclitaxel (200 mg/m\textsuperscript{2}
q3week) initiated on December 1, 2023, due to disease progression.

Current Status: Progressive disease with lymphatic, pulmonary and
hepatic metastasis.

~

\textbf{ECOG 1}

~

\textbf{Comorbidities}

Former smoker 30 py

Hypertension Stage 1

\colorbox{orange}{HFrEF NYHA II}

Type 2 Diabetes Mellitus

\colorbox{orange}{Generalized Epilepsy}

Hyperlipidemia

Benign Prostatic Hyperplasia (BPH)

~

\textbf{Medication}

~

\colorbox{orange}{Candesartan 12 mg 1-0-0

Metoprolol 47,5 mg 1-0-0

Valproic acid 500mg 2-0-2}

Metformin 1000 mg 1-0-1

Empagliflozin 10mg 1-0-0

Atorvastatin 40 mg 0-0-0-1

Omeprazole 20 mg 1-0-0

Tamsulosin 0.4 mg 1-0-0

Fentanyl TTS 25 mcg every 3 days

Fentanyl s.l. 100 mcg as needed up to 4 times a day

Ibuprofen 600 1-1-1

~

\textbf{Chronological Medical Findings:}

\textbf{June 10, 2023:} Patient presented with persistent facial
swelling and pain. A CT scan of the head and neck revealed a mass in the
left parotid gland measuring approximately 5 cm with extensive local
invasion into the surrounding soft tissues and suspected involvement of
multiple regional lymph nodes in levels II and III

\textbf{June 12, 2023:} Staging CT-scan (chest and abdomen). Multiple
nodular lesions are identified in the right lung, consistent with
metastatic disease. The largest lesion is located in the right lower
lobe, measuring approximately 2.5 cm in diameter. Additional smaller
nodules are noted in the right upper and middle lobes, with the largest
of these measuring up to 1.2 cm. No signs of metastatic involvement in
the abdomen.

\textbf{June 15, 2023:} Brain MRI. No signs of brain metastasis.

\textbf{July 5, 2023:} Ultrasound-guided biopsy confirmed salivary duct
carcinoma with high TMB and specific genetic mutations (HRAS p.Q61R,
PIK3CA p.E545K). FISH positive for HER2 amplification.

\textbf{July 12, 2023:} Started on Docetaxel and Trastuzumab.~

\textbf{October 15, 2023:} Follow-Up imaging: CT scan of the head and
neck showed a reduction in tumor size to approximately 3.5 cm. Regional
lymph nodes remained enlarged but showed decreased metabolic activity on
PET scan. All pulmonary lesions show minimal reduction in size compared
to previous scan. No new metastatic lesions.

\textbf{January 1, 2024:} Follow-up CT scan (neck, chest and abdomen)
indicated disease progression. Primary tumor remains stable in size, as
well as known lymph node metastases. Pulmonary metastases all show tumor
growth with the largest lesion in the right lower lobe now measuring 2.8
cm in diameter. Multiple, previously unknown hypodense lesion within the
left liver lobe, compatible with metastatic disease. PET scan shows high
metabolic activity.

\textbf{January 9, 2024:} Tumor board recommends considering clinical
trial options due to limited response to standard therapies.

\textbf{March 15, 2024:} Routine Labs: Comprehensive blood work
indicated normal liver and renal function. The patient maintained an
ECOG performance status of 1.

~

\begin{center}
\textbf{===== Patient 9.2.3 =====}~~~~~~~~~~~~~~
\end{center}
\textbf{Patient Information}~

Name: Mueller, Max

Born: 25.03.1945

Address: 456 Oak Street, Hamburg, Germany

~

\textbf{Overview of Tumor Diagnosis and Therapy}

Tumor Diagnosis

Diagnosis: Stage IV salivary duct carcinoma

Initial Detection: June 10, 2023, following symptoms of persistent
facial swelling and pain

Biopsy Date: July 5, 2023

Molecular Profile: Tumor Mutational Burden (TMB) of 10.5 Mut/Mb, HRAS
p.Q61R (AF 44\%), PIK3CA p.E545K (AF 39\%). HER2 FISH positive.

~

\textbf{Therapy Overview}

Initial Treatment:

Chemotherapy: Initiated on August 1, 2023, with Docetaxel (70
mg/m\textsuperscript{2}) plus Trastuzumab (8 mg/kg). Partial response
noted after three months.

Subsequent Treatment: Second-line chemotherapy with carboplatin (6
mg/m\textsuperscript{2}/min) and paclitaxel (200 mg/m\textsuperscript{2}
q3week) initiated on December 1, 2023, due to disease progression.

Current Status: Progressive disease with lymphatic, pulmonary, hepatic
and brain metastasis.

~

\textbf{ECOG 1}

~

\textbf{Comorbidities}

\colorbox{orange}{Alcohol dependence}

Former smoker 30 py

Hypertension Stage 1

Type 2 Diabetes Mellitus

Hyperlipidemia

Benign Prostatic Hyperplasia (BPH)

~

\textbf{Medication}

Amlodipine 10 mg 1-0-0

Metformin 1000 mg 1-0-1

Empagliflozin 10mg 1-0-0

Atorvastatin 40 mg 0-0-0-1

Omeprazole 20 mg 1-0-0

Tamsulosin 0.4 mg 1-0-0

Fentanyl TTS 25 mcg every 3 days

Fentanyl s.l. 100 mcg as needed up to 4 times a day

Ibuprofen 600 1-1-1

~

\textbf{Chronological Medical Findings:}

\textbf{June 10, 2023:} Patient presented with persistent facial
swelling and pain. A CT scan of the head and neck revealed a mass in the
left parotid gland measuring approximately 5 cm with extensive local
invasion into the surrounding soft tissues and suspected involvement of
multiple regional lymph nodes in levels II and III

\textbf{June 12, 2023:} Staging CT-scan (chest and abdomen). Multiple
nodular lesions are identified in the right lung, consistent with
metastatic disease. The largest lesion is located in the right lower
lobe, measuring approximately 2.5 cm in diameter. Additional smaller
nodules are noted in the right upper and middle lobes, with the largest
of these measuring up to 1.2 cm. No signs of metastatic involvement in
the abdomen.

\textbf{June 15, 2023:} Brain MRI. No signs of brain metastasis.

\textbf{July 5, 2023:} Ultrasound-guided biopsy confirmed salivary duct
carcinoma with high TMB and specific genetic mutations (HRAS p.Q61R,
PIK3CA p.E545K). FISH positive for HER2 amplification.

\textbf{July 12, 2023:} Started on Docetaxel and Trastuzumab.~

\textbf{October 15, 2023:} Follow-Up imaging: CT scan of the head and
neck showed a reduction in tumor size to approximately 3.5 cm. Regional
lymph nodes remained enlarged but showed decreased metabolic activity on
PET scan. All pulmonary lesions show minimal reduction in size compared
to previous scan. No new metastatic lesions.

\textbf{January 1, 2024:} Follow-up CT scan (neck, chest and abdomen)
indicated disease progression. Primary tumor remains stable in size, as
well as known lymph node metastases. Pulmonary metastases all show tumor
growth with the largest lesion in the right lower lobe now measuring 2.8
cm in diameter. Multiple, previously unknown hypodense lesion within the
left liver lobe, compatible with metastatic disease. PET scan shows high
metabolic activity.

\textbf{January 5, 2024:} \colorbox{orange}{MRI scan of the brain revealed multiple metastases,}\\
\colorbox{orange}{specifically two lesions in the left hemisphere:}\\
\colorbox{orange}{one in the left frontal lobe and one in the left occipital lobe.}\\
Incidental findings included mild age-related cerebral atrophy and scattered white
matter hyperintensities consistent with chronic microvascular ischemic
changes.

\textbf{January 9, 2024:} Tumor board recommends considering clinical
trial options due to limited response to standard therapies.

\textbf{March 15, 2024:} Routine Labs: Comprehensive blood work
indicated normal liver and renal function. The patient maintained an
ECOG performance status of 1.

~~~~~~~~~~~~~~
\begin{center}
\textbf{===== Patient 9.3 =====}~~~~~~~~~~~~~~
\end{center}
\textbf{Patient Information~}

Name: Mueller, Max

Born: 25.03.1945

Address: 456 Oak Street, Hamburg, Germany

~

\textbf{Overview of Tumor Diagnosis and Therapy}

Tumor Diagnosis

Diagnosis: Stage IV salivary duct carcinoma

Initial Detection: June 10, 2023, following symptoms of persistent
facial swelling and pain

Biopsy Date: July 5, 2023

Molecular Profile: Tumor Mutational Burden (TMB) of 10.5 Mut/Mb, HRAS
p.Q61R (AF 44\%), PIK3CA p.E545K (AF 39\%). HER2 FISH positive.

~

\textbf{Therapy Overview}

Initial Treatment:

Chemotherapy: Initiated on August 1, 2023, with Docetaxel plus
Trastuzumab. Partial response noted after three months.

Subsequent Treatment: Second-line chemotherapy with carboplatin and
paclitaxel initiated on December 1, 2023, due to disease progression.

Current Status: Progressive disease with lymphatic, pulmonary and
hepatic metastasis.

~

\textbf{ECOG 1}

~

\textbf{Comorbidities}

Former smoker 30 py

Hypertension Stage 1

Type 2 Diabetes Mellitus

Hyperlipidemia

Benign Prostatic Hyperplasia (BPH)

~

\textbf{Medication}

Amlodipine 10 mg 1-0-0

Metformin 1000 mg 1-0-1

Empagliflozin 10mg 1-0-0

Atorvastatin 40 mg 0-0-0-1

Omeprazole 20 mg 1-0-0

Tamsulosin 0.4 mg 1-0-0

Fentanyl TTS 25 mcg every 3 days

Fentanyl s.l. 100 mcg as needed up to 4 times a day

Ibuprofen 600 1-1-1

~

\textbf{Chronological Medical Findings:}

\textbf{June 10, 2023:} Patient presented with persistent facial
swelling and pain. A CT scan of the head and neck revealed a mass in the
left parotid gland measuring approximately 5 cm with extensive local
invasion into the surrounding soft tissues and suspected involvement of
multiple regional lymph nodes in levels II and III

\textbf{June 12, 2023:} Staging CT-scan (chest and abdomen). Multiple
nodular lesions are identified in the right lung, consistent with
metastatic disease. The largest lesion is located in the right lower
lobe, measuring approximately 2.5 cm in diameter. Additional smaller
nodules are noted in the right upper and middle lobes, with the largest
of these measuring up to 1.2 cm. No signs of metastatic involvement in
the abdomen.

\textbf{June 15, 2023:} Brain MRI. No signs of brain metastasis.

\textbf{July 5, 2023:} Ultrasound-guided biopsy confirmed salivary duct
carcinoma with high TMB and specific genetic mutations (HRAS p.Q61R,
PIK3CA p.E545K). FISH positive for HER2 amplification.

\textbf{July 12, 2023:} Started on Docetaxel and Trastuzumab.~

\textbf{October 15, 2023:} Follow-Up imaging: CT scan of the head and
neck showed a reduction in tumor size to approximately 3.5 cm. Regional
lymph nodes remained enlarged but showed decreased metabolic activity on
PET scan. All pulmonary lesions show minimal reduction in size compared
to previous scan. No new metastatic lesions.

\textbf{January 1, 2024:} Follow-up CT scan (neck, chest and abdomen)
indicated disease progression. Primary tumor remains stable in size, as
well as known lymph node metastases. Pulmonary metastases all show tumor
growth with the largest lesion in the right lower lobe now measuring 2.8
cm in diameter. Multiple, previously unknown hypodense lesion within the
left liver lobe, compatible with metastatic disease. PET scan shows high
metabolic activity.

\textbf{January 9, 2024:} Tumor board recommends considering clinical
trial options due to limited response to standard therapies.

\textbf{March 15, 2024:} Routine Labs: Comprehensive blood work
indicated normal liver and renal function. The patient maintained an
ECOG performance status of 1.

~

\begin{center}
\textbf{===== Patient 9.3.1 =====}~~~~~~~~~~~~~~
\end{center}
\textbf{Patient Information~}

Name: Mueller, Max

Born: 25.03.1945

Address: 456 Oak Street, Hamburg, Germany

~

\textbf{Overview of Tumor Diagnosis and Therapy}

Tumor Diagnosis

Diagnosis: Stage IV salivary duct carcinoma

Initial Detection: June 10, 2023, following symptoms of persistent
facial swelling and pain

Biopsy Date: July 5, 2023

Molecular Profile: Tumor Mutational Burden (TMB) of 10.5 Mut/Mb, HRAS
p.Q61R (AF 44\%), PIK3CA p.E545K (AF 39\%). HER2 FISH positive.

~

\textbf{Therapy Overview}

Initial Treatment:

Chemotherapy: Initiated on August 1, 2023, with Docetaxel plus
Trastuzumab. Partial response noted after three months.

Subsequent Treatment: Second-line chemotherapy with carboplatin and
paclitaxel initiated on December 1, 2023, due to disease progression.

Current Status: Progressive disease with lymphatic, pulmonary and
hepatic metastasis.

~

\colorbox{orange}{\textbf{ECOG 2}}

~

\textbf{Comorbidities}

Former smoker 30 py

Hypertension Stage 1

Type 2 Diabetes Mellitus

Hyperlipidemia

Benign Prostatic Hyperplasia (BPH)

\colorbox{orange}{CKD KDIGO G4}

~

\textbf{Medication}

Amlodipine 10 mg 1-0-0

Metformin 1000 mg 1-0-1

Empagliflozin 10mg 1-0-0

Atorvastatin 40 mg 0-0-0-1

Omeprazole 20 mg 1-0-0

Tamsulosin 0.4 mg 1-0-0

Fentanyl TTS 25 mcg every 3 days

Fentanyl s.l. 100 mcg as needed up to 4 times a day

Ibuprofen 600 1-1-1

~

\textbf{Chronological Medical Findings:}

\textbf{June 10, 2023:} Patient presented with persistent facial
swelling and pain. A CT scan of the head and neck revealed a mass in the
left parotid gland measuring approximately 5 cm with extensive local
invasion into the surrounding soft tissues and suspected involvement of
multiple regional lymph nodes in levels II and III

\textbf{June 12, 2023:} Staging CT-scan (chest and abdomen). Multiple
nodular lesions are identified in the right lung, consistent with
metastatic disease. The largest lesion is located in the right lower
lobe, measuring approximately 2.5 cm in diameter. Additional smaller
nodules are noted in the right upper and middle lobes, with the largest
of these measuring up to 1.2 cm. No signs of metastatic involvement in
the abdomen.

\textbf{June 15, 2023:} Brain MRI. No signs of brain metastasis.

\textbf{July 5, 2023:} Ultrasound-guided biopsy confirmed salivary duct
carcinoma with high TMB and specific genetic mutations (HRAS p.Q61R,
PIK3CA p.E545K). FISH positive for HER2 amplification.

\textbf{July 12, 2023:} Started on Docetaxel and Trastuzumab.~

\textbf{October 15, 2023:} Follow-Up imaging: CT scan of the head and
neck showed a reduction in tumor size to approximately 3.5 cm. Regional
lymph nodes remained enlarged but showed decreased metabolic activity on
PET scan. All pulmonary lesions show minimal reduction in size compared
to previous scan. No new metastatic lesions.

\textbf{January 1, 2024:} Follow-up CT scan (neck, chest and abdomen)
indicated disease progression. Primary tumor remains stable in size, as
well as known lymph node metastases. Pulmonary metastases all show tumor
growth with the largest lesion in the right lower lobe now measuring 2.8
cm in diameter. Multiple, previously unknown hypodense lesion within the
left liver lobe, compatible with metastatic disease. PET scan shows high
metabolic activity.

\textbf{January 9, 2024:} Tumor board recommends considering clinical
trial options due to limited response to standard therapies.

\textbf{March 15, 2024:} Routine Labs: Comprehensive blood work
indicated normal organ function except for known reduced kidney
function: \colorbox{orange}{eGFR 21.56 ml/min/1.73m\textsuperscript{2},}\\
\colorbox{orange}{Creatinine 3.0 mg/dl.}\\
\colorbox{orange}{Current ECOG performance status 2.}

~

\begin{center}
\textbf{===== Patient 9.3.2 =====}~~~~~~~~~~~~~~
\end{center}
\textbf{Patient Information~}

Name: Mueller, Max

Born: 25.03.1945

Address: 456 Oak Street, Hamburg, Germany

~

\textbf{Overview of Tumor Diagnosis and Therapy}

Tumor Diagnosis

Diagnosis: Stage IV salivary duct carcinoma

Initial Detection: June 10, 2023, following symptoms of persistent
facial swelling and pain

Biopsy Date: July 5, 2023

Molecular Profile: Tumor Mutational Burden (TMB) of 10.5 Mut/Mb, HRAS
p.Q61R (AF 44\%), PIK3CA p.E545K (AF 39\%). HER2 FISH positive.

~

\textbf{Therapy Overview}

Initial Treatment:

Chemotherapy: Initiated on August 1, 2023, with Docetaxel plus
Trastuzumab. Partial response noted after three months.

Subsequent Treatment: Second-line chemotherapy with carboplatin and
paclitaxel initiated on December 1, 2023, due to disease progression.

Current Status: Progressive disease with lymphatic, pulmonary and
hepatic metastasis.

~

\textbf{ECOG 1}

\colorbox{orange}{\textbf{Active P. jirovecii pneumonia}}

~

\textbf{Comorbidities}

Former smoker 30 py

Hypertension Stage 1

Type 2 Diabetes Mellitus

Hyperlipidemia

Benign Prostatic Hyperplasia (BPH)

~

\textbf{Medication}

\colorbox{orange}{Sulfamethoxazole/Trimethoprim 400mg/80mg 5--5-5-5}~

Amlodipine 10 mg 1-0-0

Metformin 1000 mg 1-0-1

Empagliflozin 10mg 1-0-0

Atorvastatin 40 mg 0-0-0-1

Omeprazole 20 mg 1-0-0

Tamsulosin 0.4 mg 1-0-0

Fentanyl TTS 25 mcg every 3 days

Fentanyl s.l. 100 mcg as needed up to 4 times a day

Ibuprofen 600 1-1-1

~

\textbf{Chronological Medical Findings:}

\textbf{June 10, 2023:} Patient presented with persistent facial
swelling and pain. A CT scan of the head and neck revealed a mass in the
left parotid gland measuring approximately 5 cm with extensive local
invasion into the surrounding soft tissues and suspected involvement of
multiple regional lymph nodes in levels II and III

\textbf{June 12, 2023:} Staging CT-scan (chest and abdomen). Multiple
nodular lesions are identified in the right lung, consistent with
metastatic disease. The largest lesion is located in the right lower
lobe, measuring approximately 2.5 cm in diameter. Additional smaller
nodules are noted in the right upper and middle lobes, with the largest
of these measuring up to 1.2 cm. No signs of metastatic involvement in
the abdomen.

\textbf{June 15, 2023:} Brain MRI. No signs of brain metastasis.

\textbf{July 5, 2023:} Ultrasound-guided biopsy confirmed salivary duct
carcinoma with high TMB and specific genetic mutations (HRAS p.Q61R,
PIK3CA p.E545K). FISH positive for HER2 amplification.

\textbf{July 12, 2023:} Started on Docetaxel and Trastuzumab.~

\textbf{October 15, 2023:} Follow-Up imaging: CT scan of the head and
neck showed a reduction in tumor size to approximately 3.5 cm. Regional
lymph nodes remained enlarged but showed decreased metabolic activity on
PET scan. All pulmonary lesions show minimal reduction in size compared
to previous scan. No new metastatic lesions.

\textbf{January 1, 2024:} Follow-up CT scan (neck, chest and abdomen)
indicated disease progression. Primary tumor remains stable in size, as
well as known lymph node metastases. Pulmonary metastases all show tumor
growth with the largest lesion in the right lower lobe now measuring 2.8
cm in diameter. Multiple, previously unknown hypodense lesion within the
left liver lobe, compatible with metastatic disease. PET scan shows high
metabolic activity.

\textbf{January 9, 2024:} Tumor board recommends considering clinical
trial options due to limited response to standard therapies.

\textbf{March 15, 2024:} Routine Labs: Comprehensive blood work
indicated normal liver and renal function. The patient maintained an
ECOG performance status of 1.

~~~~~~~~~~~~~~
\begin{center}
\textbf{===== Patient 9.3.3 =====}~~~~~~~~~~~~~~
\end{center}
\textbf{Patient Information~}

Name: Mueller, Max

Born: 25.03.1945

Address: 456 Oak Street, Hamburg, Germany

~

\textbf{Overview of Tumor Diagnosis and Therapy}

Tumor Diagnosis

Diagnosis: Stage IV salivary duct carcinoma

Initial Detection: June 10, 2023, following symptoms of persistent
facial swelling and pain

Biopsy Date: July 5, 2023

Molecular Profile: Tumor Mutational Burden (TMB) of 10.5 Mut/Mb, HRAS
p.Q61R (AF 44\%), PIK3CA p.E545K (AF 39\%). HER2 FISH positive.

~

\textbf{Therapy Overview}

Initial Treatment:

Chemotherapy: Initiated on August 1, 2023, with Docetaxel plus
Trastuzumab. Partial response noted after three months.

Subsequent Treatment: Second-line chemotherapy with carboplatin and
paclitaxel initiated on December 1, 2023, due to disease progression.

Current Status: Progressive disease with lymphatic, pulmonary, hepatic
and \colorbox{orange}{brain metastasis}.

~

\textbf{ECOG 1}

~

\textbf{Comorbidities}

Former smoker 30 py

Hypertension Stage 1

Type 2 Diabetes Mellitus

Hyperlipidemia

Benign Prostatic Hyperplasia (BPH)

~

\textbf{Medication}

Amlodipine 10 mg 1-0-0

Metformin 1000 mg 1-0-1

Empagliflozin 10mg 1-0-0

Atorvastatin 40 mg 0-0-0-1

Omeprazole 20 mg 1-0-0

Tamsulosin 0.4 mg 1-0-0

Fentanyl TTS 25 mcg every 3 days

Fentanyl s.l. 100 mcg as needed up to 4 times a day

Ibuprofen 600 1-1-1

~

\textbf{Chronological Medical Findings:}

\textbf{June 10, 2023:} Patient presented with persistent facial
swelling and pain. A CT scan of the head and neck revealed a mass in the
left parotid gland measuring approximately 5 cm with extensive local
invasion into the surrounding soft tissues and suspected involvement of
multiple regional lymph nodes in levels II and III

\textbf{June 12, 2023:} Staging CT-scan (chest and abdomen). Multiple
nodular lesions are identified in the right lung, consistent with
metastatic disease. The largest lesion is located in the right lower
lobe, measuring approximately 2.5 cm in diameter. Additional smaller
nodules are noted in the right upper and middle lobes, with the largest
of these measuring up to 1.2 cm. No signs of metastatic involvement in
the abdomen.

\textbf{June 15, 2023:} Brain MRI. No signs of brain metastasis.

\textbf{July 5, 2023:} Ultrasound-guided biopsy confirmed salivary duct
carcinoma with high TMB and specific genetic mutations (HRAS p.Q61R,
PIK3CA p.E545K). FISH positive for HER2 amplification.

\textbf{July 12, 2023:} Started on Docetaxel and Trastuzumab.~

\textbf{October 15, 2023:} Follow-Up imaging: CT scan of the head and
neck showed a reduction in tumor size to approximately 3.5 cm. Regional
lymph nodes remained enlarged but showed decreased metabolic activity on
PET scan. All pulmonary lesions show minimal reduction in size compared
to previous scan. No new metastatic lesions.

\textbf{January 1, 2024:} Follow-up CT scan (neck, chest and abdomen)
indicated disease progression. Primary tumor remains stable in size, as
well as known lymph node metastases. Pulmonary metastases all show tumor
growth with the largest lesion in the right lower lobe now measuring 2.8
cm in diameter. Multiple, previously unknown hypodense lesion within the
left liver lobe, compatible with metastatic disease. PET scan shows high
metabolic activity.

\textbf{January 4, 2024:} \colorbox{orange}{MRI scan of the brain revealed multiple metastases,}\\
\colorbox{orange}{specifically two lesions in the right hemisphere:}\\
\colorbox{orange}{one in the right frontal lobe and one in the right parietal lobe.}\\
Incidental findings included mild age-related cerebral atrophy and scattered white
matter hyperintensities consistent with chronic microvascular ischemic
changes.

\textbf{January 9, 2024:} Tumor board recommends considering clinical
trial options due to limited response to standard therapies.

\textbf{March 15, 2024:} Routine Labs: Comprehensive blood work
indicated normal liver and renal function. The patient maintained an
ECOG performance status of 1.

~

\begin{center}
\textbf{===== Patient 10.1 =====}~~~~~~~~~~~~~~
\end{center}
Name: Miller, Jane\\
Born: 25.07.1965\\
Address: Main Street 78, Potsdam

~

\textbf{Overview of Tumor Diagnosis and Therapy\\
Tumor Diagnosis\\
}Diagnosis: UICC Stage IV EGFR mutated non-small cell lung cancer
(NSCLC)\\
Initial Detection: January 10, 2023, following symptoms of persistent
cough and chest pain\\
Biopsy Date: February 5, 2023, adenocarcinoma of the lung\\
Molecular Profile: Panel (tumor purity 60\%). EGFR p.E746\_A750del (AF
50\%), EGFR T790M, EGFR p.C797S (AF 29\%), STK11 p.C210* (AF 39\%).

~

\textbf{Therapy Overview\\
}Initial Treatment:\\
Targeted Therapy: Began March 1, 2023, with Osimertinib~ (T790M).
Partial response noted after initial therapy cycle completed by June 15,
2023. Continued therapy until November 2021 (progressive disease).

Subsequent Treatment:\\
Further treatment with Pembrolizumab in combination with
Paclitaxel/Carboplatin/ Bevacizumab and Atezolizumab initiated on
December 1, 2023. Staging CT shows disease progression after 6 months.

Current Status: \textbf{ECOG 1\\
}

\textbf{Comorbidities\\
}Current smoker: 35 py

Hypertension Stage 1\\
Hyperlipidemia: Managed with Simvastatin 20 mg daily\\
COPD GOLD 2

\textbf{Medication}

Losartan 50mg 1-0-0

Simvastatin 20mg daily 0-0-0-1

Albuterol (inhaler) on demand~

Tiotropium (inhaler) on demand

~~~~~~~~~~~

\textbf{Chronological Medical Findings:\\
January 2023}: Complaints of persistent cough and bloody sputum. Weight
loss of -10kg / 5 months. Chest X-ray revealed a mass in the right
lung.~

\textbf{January 10, 2023:} Comprehensive CT scan (chest and abdomen):
Solid, spiculated mass in the right upper lobe of the lung measuring
approximately 3.8 cm. Additionally, three small hypoattenuating hepatic
lesions noted, to be considered as metastases. Enlarged hilar and
subcarinal lymph nodes. Right adrenal gland slightly enlarged,
warranting further investigation for metastatic involvement. No evidence
of pleural effusion or significant vascular invasion was present.
Additional notes: Minor atelectasis in the left lower lobe and mild
emphysematous changes in both lungs, consistent with the patient's
history of chronic obstructive pulmonary disease (COPD). The abdominal
organs, aside from the hepatic lesions and possible adrenal metastasis,
appeared unremarkable.

\textbf{February 5, 2023:} Biopsy and molecular testing confirmed
adenocarcinoma with EGFR T790M mutation. Material sent for further panel
testing. Initiated Osimertinib therapy on March 1, 2021. Patient signed
consent. Patient in clinical good condition.

\textbf{Feb.-June 2023:} Antineoplastic targeted therapy with
Osimertinib.

\textbf{June 15, 2023:} Follow-up CT scan (chest and abdomen), PET CT
scan: Partial response (PR) to treatment. The primary lung mass in the
right upper lobe decreased in size, now measuring approximately 3 cm in
diameter, down from 3.8 cm. The three previously noted hypoattenuating
hepatic lesions have also shown slight reduction in size and decreased
metabolic activity, suggesting a positive response to systemic therapy.
No new metastatic lesions detected in the liver or other abdominal
organs. The previously enlarged hilar and subcarinal lymph nodes have
reduced in size, indicating a favorable response to treatment. The right
adrenal gland remains slightly enlarged but stable, with no significant
change noted, and it continues to show no signs of active disease.
Overall: PR.~

\textbf{July 1, 2023:} Continuation of Osimertinib therapy

\textbf{October 3, 2023:} CT scan Chest/Abd.: PD. The primary lung mass
in the right upper lobe has increased in size, now measuring
approximately 4.5 cm in diameter (prev 3,0 cm). The previously noted
hypoattenuating hepatic lesions have also shown slight growth.
Additional new metastasis in S7. Small pleural effusion on the right
side, minor atelectasis in the left lower lobe has slightly worsened.
Hilar and subcarinal lymph nodes drastically enlarged in size. Right
adrenal gland with second metastasis. No new metastatic lesions were
detected in the liver or other abdominal organs.~

\textbf{December 1, 2023:} Begin Paclitaxel/Carboplatin/Bevacizumab and
Atezolizumab. Received written consent from the patient, ECOG 1.

\textbf{May 10, 2024:} CT Lung/Abdomen: Progressive Disease. Primary
lung mass in the right upper lobe has increased in size, now measuring
approximately 4.5 cm in diameter, up from 3 cm. The previously noted
hypoattenuating hepatic lesions have also shown slight growth. Small
pleural effusion on the right side. Minor atelectasis in the left lower
lobe slightly worsened, likely due to the progressive nature of the
disease and the presence of pleural effusion. The hilar and subcarinal
lymph nodes, prev. reduced in size, now slightly enlarged again. Right
adrenal gland remains slightly enlarged and stable with no significant
change in size. No new metastatic lesions were detected in the liver or
other abdominal organs.

Summary:

~~~~~~~~ Primary Lung Mass: Increased in size to 4.5 cm.

~~~~~~~~ Hepatic Lesions: Slight growth and increased metabolic
activity.

~~~~~~~~ Lymph Nodes: Slightly re-enlarged hilar and subcarinal nodes.

~~~~~~~~ Pleural Effusion: Small right-sided pleural effusion noted.

~~~~~~~~ Atelectasis: Slight worsening of minor atelectasis in the left
lower lobe.

~~~~~~~~ Adrenal Gland: Remains slightly enlarged; increased metabolic
activity.

~

Overall Assessment: Disease progression (PD).

~

\textbf{May 15, 2024:} Tumor board recommends considering clinical trial
options due to limited response to standard therapies.

\textbf{May 18, 2024:} Detailed assessment of health status. ECOG
performance 1. All routine labs, including liver and renal function
tests, within normal limits.

~~~~~~~~~~~~~~~
\begin{center}
\textbf{===== Patient 10.1.1 =====}~~~~~~~~~~~~~~~
\end{center}
Name: Miller, Jane\\
Born: 25.07.1965\\
Address: Main Street 78, Potsdam

~

\textbf{Overview of Tumor Diagnosis and Therapy\\
Tumor Diagnosis\\
}Diagnosis: UICC Stage IV EGFR mutated non-small cell lung cancer
(NSCLC)\\
Initial Detection: January 10, 2023, following symptoms of persistent
cough and chest pain\\
Biopsy Date: February 5, 2023, adenocarcinoma of the lung\\
Molecular Profile: Panel (tumor purity 60\%). EGFR p.E746\_A750del (AF
50\%), EGFR T790M, EGFR p.C797S (AF 29\%), STK11 p.C210* (AF 39\%).

~

\textbf{Therapy Overview\\
}Initial Treatment:\\
Targeted Therapy: Began March 1, 2023, with Osimertinib~ (T790M).
Partial response noted after initial therapy cycle completed by June 15,
2023. Continued therapy until November 2021 (progressive disease).

Subsequent Treatment:\\
Further treatment with Pembrolizumab in combination with
Paclitaxel/Carboplatin/ Bevacizumab and Atezolizumab initiated on
December 1, 2023. Staging CT shows disease progression after 6 months.

Current Status:\\
\textbf{ECOG 1\\
\strut \\
\strut \\
\strut \\
}

\textbf{Comorbidities\\
}Current smoker: 35 py

Hypertension Stage 1\\
Hyperlipidemia: Managed with Simvastatin 20 mg daily\\
COPD GOLD 2

\textbf{Medication}

Losartan 50mg 1-0-0

Simvastatin 20mg daily 0-0-0-1

Albuterol (inhaler) on demand~

Tiotropium (inhaler) on demand

~~~~~~~~~~~

\textbf{Chronological Medical Findings:\\
January 2023}: Complaints of persistent cough and bloody sputum. Weight
loss of -10kg / 5 months. Chest X-ray revealed a mass in the right
lung.~

\textbf{January 10, 2023:} Comprehensive CT scan (chest and abdomen):
Solid, spiculated mass in the right upper lobe of the lung measuring
approximately 3.8 cm. Additionally, three small hypoattenuating hepatic
lesions noted, to be considered as metastases. Enlarged hilar and
subcarinal lymph nodes. Right adrenal gland slightly enlarged,
warranting further investigation for metastatic involvement. No evidence
of pleural effusion or significant vascular invasion was present.
Additional notes: Minor atelectasis in the left lower lobe and mild
emphysematous changes in both lungs, consistent with the patient's
history of chronic obstructive pulmonary disease (COPD). The abdominal
organs, aside from the hepatic lesions and possible adrenal metastasis,
appeared unremarkable.

\textbf{February 5, 2023:} Biopsy and molecular testing confirmed
adenocarcinoma with EGFR T790M mutation. Material sent for further panel
testing. Initiated Osimertinib therapy on March 1, 2021. Patient signed
consent. Patient in clinical good condition.

\textbf{Feb.-June 2023:} Antineoplastic targeted therapy with
Osimertinib.

\textbf{June 15, 2023:} Follow-up CT scan (chest and abdomen), PET CT
scan: Partial response (PR) to treatment. The primary lung mass in the
right upper lobe decreased in size, now measuring approximately 3 cm in
diameter, down from 3.8 cm. The three previously noted hypoattenuating
hepatic lesions have also shown slight reduction in size and decreased
metabolic activity, suggesting a positive response to systemic therapy.
No new metastatic lesions detected in the liver or other abdominal
organs. The previously enlarged hilar and subcarinal lymph nodes have
reduced in size, indicating a favorable response to treatment. The right
adrenal gland remains slightly enlarged but stable, with no significant
change noted, and it continues to show no signs of active disease.
Overall: PR.~

\textbf{July 1, 2023:} Continuation of Osimertinib therapy

\textbf{October 3, 2023:} CT scan Chest/Abd.: PD. The primary lung mass
in the right upper lobe has increased in size, now measuring
approximately 4.5 cm in diameter (prev 3,0 cm). The previously noted
hypoattenuating hepatic lesions have also shown slight growth.
Additional new metastasis in S7. Small pleural effusion on the right
side, minor atelectasis in the left lower lobe has slightly worsened.
Hilar and subcarinal lymph nodes drastically enlarged in size. Right
adrenal gland with second metastasis. No new metastatic lesions were
detected in the liver or other abdominal organs.~

\textbf{December 1, 2023:} Begin Paclitaxel/Carboplatin/Bevacizumab and
Atezolizumab. Received written consent from the patient, ECOG 1.

\colorbox{orange}{\textbf{May 8, 2024:} Stopped platin because of severe neuropathy (CTCAE
III)}

\textbf{May 10, 2024:} CT Lung/Abdomen: Progressive Disease. Primary
lung mass in the right upper lobe has increased in size, now measuring
approximately 4.5 cm in diameter, up from 3 cm. The previously noted
hypoattenuating hepatic lesions have also shown slight growth. Small
pleural effusion on the right side. Minor atelectasis in the left lower
lobe slightly worsened, likely due to the progressive nature of the
disease and the presence of pleural effusion. The hilar and subcarinal
lymph nodes, prev. reduced in size, now slightly enlarged again. Right
adrenal gland remains slightly enlarged and stable with no significant
change in size. No new metastatic lesions were detected in the liver or
other abdominal organs.

Summary:

~~~~~~~~ Primary Lung Mass: Increased in size to 4.5 cm.

~~~~~~~~ Hepatic Lesions: Slight growth and increased metabolic
activity.

~~~~~~~~ Lymph Nodes: Slightly re-enlarged hilar and subcarinal nodes.

~~~~~~~~ Pleural Effusion: Small right-sided pleural effusion noted.

~~~~~~~~ Atelectasis: Slight worsening of minor atelectasis in the left
lower lobe.

~~~~~~~~ Adrenal Gland: Remains slightly enlarged; increased metabolic
activity.

~

Overall Assessment: Disease progression (PD).

\textbf{May 15, 2024:} Tumor board recommends considering clinical trial
options due to limited response to standard therapies.

\textbf{May 18, 2024:} Detailed assessment of health status. ECOG
performance 1. All routine labs, including liver and renal function
tests, within normal limits.

\colorbox{orange}{\textbf{May 20, 2024:} Presentation with shortness of breath via
emergeny.}\\
\colorbox{orange}{CT scan with multiple infiltrates.}\\
\colorbox{orange}{CRP elevated at 190. Started on Meropenem. Admitted for inpatient care.}

~~~~~~~~~~~~~~~
\begin{center}
\textbf{===== Patient 10.1.2 =====}~~~~~~~~~~~~~~~
\end{center}
Name: Miller, Jane\\
Born: 25.07.1965\\
Address: Main Street 78, Potsdam

~

\textbf{Overview of Tumor Diagnosis and Therapy\\
Tumor Diagnosis\\
}Diagnosis: UICC Stage IV EGFR mutated non-small cell lung cancer
(NSCLC)\\
Initial Detection: January 10, 2023, following symptoms of persistent
cough and chest pain\\
Biopsy Date: February 5, 2023, adenocarcinoma of the lung\\
Molecular Profile: Panel (tumor purity 60\%). EGFR p.E746\_A750del (AF
50\%), EGFR T790M, EGFR p.C797S (AF 29\%), STK11 p.C210* (AF 39\%).

~

\textbf{Therapy Overview\\
}Initial Treatment:\\
Targeted Therapy: Began March 1, 2023, with Osimertinib~ (T790M).
Partial response noted after initial therapy cycle completed by June 15,
2023. Continued therapy until November 2021 (progressive disease).

Subsequent Treatment:\\
Further treatment with Pembrolizumab in combination with
Paclitaxel/Carboplatin/ Bevacizumab and Atezolizumab initiated on
December 1, 2023. Staging CT shows disease progression after 6 months.

Current Status:\\
\textbf{ECOG 1\\
}

\textbf{Comorbidities\\
}Current smoker: 35 py

Hypertension Stage 1\\
Hyperlipidemia: Managed with Simvastatin 20 mg daily\\
COPD GOLD 2

\textbf{Medication}

Losartan 50mg 1-0-0

Simvastatin 20mg daily 0-0-0-1

Albuterol (inhaler) on demand~

Tiotropium (inhaler) on demand

~~~~~~~~~~~

\textbf{Chronological Medical Findings:\\
January 2023}: Complaints of persistent cough and bloody sputum. Weight
loss of -10kg / 5 months. Chest X-ray revealed a mass in the right
lung.~

\textbf{January 10, 2023:} Comprehensive CT scan (chest and abdomen):
Solid, spiculated mass in the right upper lobe of the lung measuring
approximately 3.8 cm. Additionally, three small hypoattenuating hepatic
lesions noted, to be considered as metastases. Enlarged hilar and
subcarinal lymph nodes. Right adrenal gland slightly enlarged,
warranting further investigation for metastatic involvement. No evidence
of pleural effusion or significant vascular invasion was present.
Additional notes: Minor atelectasis in the left lower lobe and mild
emphysematous changes in both lungs, consistent with the patient's
history of chronic obstructive pulmonary disease (COPD). The abdominal
organs, aside from the hepatic lesions and possible adrenal metastasis,
appeared unremarkable.

\textbf{February 5, 2023:} Biopsy and molecular testing confirmed
adenocarcinoma with EGFR T790M mutation. Material sent for further panel
testing. Initiated Osimertinib therapy on March 1, 2021. Patient signed
consent. Patient in clinical good condition.

\textbf{Feb.-June 2023:} Antineoplastic targeted therapy with
Osimertinib.

\textbf{June 15, 2023:} Follow-up CT scan (chest and abdomen), PET CT
scan: Partial response (PR) to treatment. The primary lung mass in the
right upper lobe decreased in size, now measuring approximately 3 cm in
diameter, down from 3.8 cm. The three previously noted hypoattenuating
hepatic lesions have also shown slight reduction in size and decreased
metabolic activity, suggesting a positive response to systemic therapy.
No new metastatic lesions detected in the liver or other abdominal
organs. The previously enlarged hilar and subcarinal lymph nodes have
reduced in size, indicating a favorable response to treatment. The right
adrenal gland remains slightly enlarged but stable, with no significant
change noted, and it continues to show no signs of active disease.
Overall: PR.~

\textbf{July 1, 2023:} Continuation of Osimertinib therapy

\textbf{October 3, 2023:} CT scan Chest/Abd.: PD. The primary lung mass
in the right upper lobe has increased in size, now measuring
approximately 4.5 cm in diameter (prev 3,0 cm). The previously noted
hypoattenuating hepatic lesions have also shown slight growth.
Additional new metastasis in S7. Small pleural effusion on the right
side, minor atelectasis in the left lower lobe has slightly worsened.
Hilar and subcarinal lymph nodes drastically enlarged in size. Right
adrenal gland with second metastasis. No new metastatic lesions were
detected in the liver or other abdominal organs.~

\textbf{December 1, 2023:} Begin Paclitaxel/Carboplatin/Bevacizumab and
Atezolizumab. Received written consent from the patient, ECOG 1.

\textbf{May 10, 2024:} CT Lung/Abdomen: Progressive Disease. Primary
lung mass in the right upper lobe has increased in size, now measuring
approximately 4.5 cm in diameter, up from 3 cm. The previously noted
hypoattenuating hepatic lesions have also shown slight growth. Small
pleural effusion on the right side. Minor atelectasis in the left lower
lobe slightly worsened, likely due to the progressive nature of the
disease and the presence of pleural effusion. The hilar and subcarinal
lymph nodes, prev. reduced in size, now slightly enlarged again. Right
adrenal gland remains slightly enlarged and stable with no significant
change in size. No new metastatic lesions were detected in the liver or
other abdominal organs.

Summary:

~~~~~~~~ Primary Lung Mass: Increased in size to 4.5 cm.

~~~~~~~~ Hepatic Lesions: Slight growth and increased metabolic
activity.

~~~~~~~~ Lymph Nodes: Slightly re-enlarged hilar and subcarinal nodes.

~~~~~~~~ Pleural Effusion: Small right-sided pleural effusion noted.

~~~~~~~~ Atelectasis: Slight worsening of minor atelectasis in the left
lower lobe.

~~~~~~~~ Adrenal Gland: Remains slightly enlarged; increased metabolic
activity.

~

Overall Assessment: Disease progression (PD).

\colorbox{orange}{\textbf{May 11, 2024:} Seizures, CT scan: 3 metastases in the brain:}\\
\colorbox{orange}{1x2.5 cm lesion in the left frontal lobe.}\\
\colorbox{orange}{A 1.8 cm lesion in the right parietal lobe. A 1.2 cm lesion in the
cerebellum.}\\
\colorbox{orange}{Initiated Prednisone. Begin with Keppra.}\\
\colorbox{orange}{Consultation with the radiation oncology team, recommended }\\
\colorbox{orange}{whole-brain radiotherapy (30 Gy in 10 fractions).}\\
\colorbox{orange}{Surgical resection deemed not possible due to location of metasases.}~

\textbf{May 15, 2024:} Tumor board recommends considering clinical trial
options due to limited response to standard therapies.

\textbf{May 18, 2024:} Detailed assessment of health status. ECOG
performance 1. All routine labs, including liver and renal function
tests, within normal limits.

~~~~~~~~~~~~~~~~
\begin{center}
\textbf{===== Patient 10.1.3 =====}~~~~~~~~~~~~~~~
\end{center}
Name: Miller, Jane\\
Born: 25.07.1965\\
Address: Main Street 78, Potsdam

~

\textbf{Overview of Tumor Diagnosis and Therapy\\
Tumor Diagnosis\\
}Diagnosis: UICC Stage IV EGFR mutated non-small cell lung cancer
(NSCLC)\\
Initial Detection: January 10, 2023, following symptoms of persistent
cough and chest pain\\
Biopsy Date: February 5, 2023, adenocarcinoma of the lung\\
Molecular Profile: Panel (tumor purity 60\%). EGFR p.E746\_A750del (AF
50\%), EGFR T790M, EGFR p.C797S (AF 29\%), STK11 p.C210* (AF 39\%).

~

\textbf{Therapy Overview\\
}Initial Treatment:\\
Targeted Therapy: Began March 1, 2023, with Osimertinib~ (T790M).
Partial response noted after initial therapy cycle completed by June 15,
2023. Continued therapy until November 2021 (progressive disease).

Subsequent Treatment:\\
Further treatment with Pembrolizumab in combination with
Paclitaxel/Carboplatin/ Bevacizumab and Atezolizumab initiated on
December 1, 2023. Staging CT shows disease progression after 6 months.

Current Status:\\
\textbf{ECOG 1\\
\strut \\
Comorbidities\\
}Current smoker: 35 py

Hypertension Stage 1\\
Hyperlipidemia: Managed with Simvastatin 20 mg daily\\
COPD GOLD 2
\colorbox{orange}{
Diabetes Mellitus (II)~

Diabetic Retinopathy}

\textbf{Medication}

Losartan 50mg 1-0-0

Simvastatin 20mg daily 0-0-0-1

\colorbox{orange}{Metformin~ (800 mg 1-0-1)}

Albuterol (inhaler) on demand~

Tiotropium (inhaler) on demand

\colorbox{orange}{Lucentis}~

~~~~~~~~~~~

\textbf{Chronological Medical Findings:\\
January 2023}: Complaints of persistent cough and bloody sputum. Weight
loss of -10kg / 5 months. Chest X-ray revealed a mass in the right
lung.~

\textbf{January 10, 2023:} Comprehensive CT scan (chest and abdomen):
Solid, spiculated mass in the right upper lobe of the lung measuring
approximately 3.8 cm. Additionally, three small hypoattenuating hepatic
lesions noted, to be considered as metastases. Enlarged hilar and
subcarinal lymph nodes. Right adrenal gland slightly enlarged,
warranting further investigation for metastatic involvement. No evidence
of pleural effusion or significant vascular invasion was present.
Additional notes: Minor atelectasis in the left lower lobe and mild
emphysematous changes in both lungs, consistent with the patient's
history of chronic obstructive pulmonary disease (COPD). The abdominal
organs, aside from the hepatic lesions and possible adrenal metastasis,
appeared unremarkable.

\textbf{February 5, 2023:} Biopsy and molecular testing confirmed
adenocarcinoma with EGFR T790M mutation. Material sent for further panel
testing. Initiated Osimertinib therapy on March 1, 2021. Patient signed
consent. Patient in clinical good condition.

\textbf{Feb.-June 2023:} Antineoplastic targeted therapy with
Osimertinib.

\textbf{June 15, 2023:} Follow-up CT scan (chest and abdomen), PET CT
scan: Partial response (PR) to treatment. The primary lung mass in the
right upper lobe decreased in size, now measuring approximately 3 cm in
diameter, down from 3.8 cm. The three previously noted hypoattenuating
hepatic lesions have also shown slight reduction in size and decreased
metabolic activity, suggesting a positive response to systemic therapy.
No new metastatic lesions detected in the liver or other abdominal
organs. The previously enlarged hilar and subcarinal lymph nodes have
reduced in size, indicating a favorable response to treatment. The right
adrenal gland remains slightly enlarged but stable, with no significant
change noted, and it continues to show no signs of active disease.
Overall: PR.~

\textbf{July 1, 2023:} Continuation of Osimertinib therapy

\textbf{October 3, 2023:} CT scan Chest/Abd.: PD. The primary lung mass
in the right upper lobe has increased in size, now measuring
approximately 4.5 cm in diameter (prev 3,0 cm). The previously noted
hypoattenuating hepatic lesions have also shown slight growth.
Additional new metastasis in S7. Small pleural effusion on the right
side, minor atelectasis in the left lower lobe has slightly worsened.
Hilar and subcarinal lymph nodes drastically enlarged in size. Right
adrenal gland with second metastasis. No new metastatic lesions were
detected in the liver or other abdominal organs.~

\textbf{December 1, 2023:} Begin Paclitaxel/Carboplatin/Bevacizumab and
Atezolizumab. Received written consent from the patient, ECOG 1.

\textbf{May 10, 2024:} CT Lung/Abdomen: Progressive Disease. Primary
lung mass in the right upper lobe has increased in size, now measuring
approximately 4.5 cm in diameter, up from 3 cm. The previously noted
hypoattenuating hepatic lesions have also shown slight growth. Small
pleural effusion on the right side. Minor atelectasis in the left lower
lobe slightly worsened, likely due to the progressive nature of the
disease and the presence of pleural effusion. The hilar and subcarinal
lymph nodes, prev. reduced in size, now slightly enlarged again. Right
adrenal gland remains slightly enlarged and stable with no significant
change in size. No new metastatic lesions were detected in the liver or
other abdominal organs.

Summary:

~~~~~~~~ Primary Lung Mass: Increased in size to 4.5 cm.

~~~~~~~~ Hepatic Lesions: Slight growth and increased metabolic
activity.

~~~~~~~~ Lymph Nodes: Slightly re-enlarged hilar and subcarinal nodes.

~~~~~~~~ Pleural Effusion: Small right-sided pleural effusion noted.

~~~~~~~~ Atelectasis: Slight worsening of minor atelectasis in the left
lower lobe.

~~~~~~~~ Adrenal Gland: Remains slightly enlarged; increased metabolic
activity.

~

Overall Assessment: Disease progression (PD).

\textbf{May 15, 2024:} Tumor board recommends considering clinical trial
options due to limited response to standard therapies.

\textbf{May 18, 2024:} Detailed assessment of health status.
\\
\colorbox{orange}{Patient progressively in worse conditions, currently ECOG performance 2.}\\
All routine labs, including liver and renal function tests, within normal
limits.

~~~~~~~~~~~~~~~~
\begin{center}
\textbf{===== Patient 11.1 =====}~~~~~~~~~~~~~~~~
\end{center}
\textbf{Patient Information}

Name: Smith, Anna

Born: 10.03.1980

Address: Another Avenue 89, Augsburg

\textbf{Overview of Tumor Diagnosis and Therapy}

\textbf{Tumor Diagnosis}

\textbf{Diagnosis}: UICC Stage IV well-differentiated, non-functioning
neuroendocrine tumor of the pancreas

Initial Detection: June 5, 2021, following symptoms of abdominal pain
and weight loss

Biopsy Date: July 1, 2021, well-differentiated neuroendocrine tumor of
the pancreas

Molecular Profile: High Tumor Mutational Burden (TMB-H, >=10 mut/Mb,
F1CDx assay)

Therapy Overview

\textbf{Initial Treatment:}

Chemotherapy: Began July 20, 2021, with Capecitabine and Temozolomide
regimen. Partial response noted after initial chemotherapy cycle
completed by October 10, 2021. Continued therapy until January 2022
(progressive disease).

Subsequent Treatment:

Further chemotherapy with Everolimus initiated on February 1, 2022.
Staging CT shows disease progression after 6 months.

Current Status: ECOG 1

\textbf{Comorbidities}

Hypertension Stage 2

History of appendectomy 2001

Medication

Ramipril 10mg 1-0-0

Amlodipin 5mg 1-0-0

Hydrochlorothiazide 12.5mg 1-0-0~~~~~~~~

~

\textbf{Chronological Medical Findings:}

\textbf{May 10, 2021:} Complained of abdominal pain and weight loss.
Abdominal ultrasound revealed a mass in the pancreas. Referred to
oncologist. No symptoms attributable to carcinoid syndrome.

\textbf{June 5, 2021:} CT scan of the abdomen and pelvis: Revealed a
well-defined mass in the head of the pancreas measuring approximately
4.5 cm with surrounding lymphadenopathy. No signs of vascular invasion,
but multiple small hepatic lesions were identified, indicating
metastatic disease. No evidence of bowel obstruction, but slight
dilation of the pancreatic duct was observed.

\textbf{July 1, 2021:} Biopsy and molecular testing confirmed a
well-differentiated neuroendocrine tumor with a high tumor mutational
burden (TMB-H, >=10 mut/Mb, F1CDx assay), Grade 3. Initiated Capecitabine
and Temozolomide chemotherapy regimen on July 20, 2021.

\textbf{October 10, 2021:} Follow-up CT scan (chest, abdomen, and
pelvis): Partial response to treatment observed, with the primary
pancreatic mass reducing to 3.5 cm in diameter. Hepatic lesions showed
slight reduction in size and metabolic activity. No new metastatic
lesions were detected, but mild ascites persisted.

\textbf{January 1, 2022:} Continued Capecitabine and Temozolomide
therapy, with staging scans showing moderate disease progression.
Primary mass increased to 4.0 cm, with new peritoneal nodules.

\textbf{February 1, 2022:} Initiated Everolimus therapy due to
progression on previous regimen.

\textbf{August 1, 2022:} Follow-up MRI scan (abdomen and pelvis): MRI
indicated further disease progression, with the primary tumor enlarging
to 4.8 cm and increased involvement of adjacent hepatic structures.
Peritoneal nodules showed slight growth, and moderate ascites was
present. There was no evidence of bowel obstruction or significant
vascular invasion.

\textbf{November 15, 2022:} Tumor board recommends considering clinical
trial options due to limited response to standard therapies.

\textbf{December 1, 2022:} Detailed assessment of health status. ECOG
performance status 1. All routine labs, including liver and renal
function tests, within normal limits.

~~~~~~~~~~~~~~~~
\begin{center}
\textbf{===== Patient 11.1.1 =====}~~~~~~~~~~~~~~~~
\end{center}
\textbf{Patient Information}

Name: Smith, Anna

Born: 10.03.1980

Address: Another Avenue 89, Augsburg

\textbf{Overview of Tumor Diagnosis and Therapy}

\textbf{Tumor Diagnosis}

\textbf{Diagnosis}: UICC Stage IV well-differentiated, non-functioning
neuroendocrine tumor of the pancreas

Initial Detection: June 5, 2021, following symptoms of abdominal pain
and weight loss

Biopsy Date: July 1, 2021, well-differentiated neuroendocrine tumor of
the pancreas

Molecular Profile: High Tumor Mutational Burden (TMB-H, >=10 mut/Mb,
F1CDx assay)

\textbf{Therapy Overview}

\textbf{Initial Treatment:}

Chemotherapy: Began July 20, 2021, with Capecitabine and Temozolomide
regimen. Partial response noted after initial chemotherapy cycle
completed by October 10, 2021. Continued therapy until January 2022
(progressive disease).

\textbf{Subsequent Treatment:}

Further chemotherapy with Everolimus initiated on February 1, 2022.
Staging CT shows disease progression after 6 months.

Current Status: ECOG 1

\textbf{Comorbidities}

Hypertension Stage 2

History of appendectomy 2001

\colorbox{orange}{Systemic lupus erythematodes (last systemic therapy 09/2020)}

~

\textbf{Medication}

Ramipril 10mg 1-0-0

Amlodipin 5mg 1-0-0

\colorbox{orange}{Hydrochlorothiazide 12.5mg 1-0-0~

Hydroxychloroquine 200mg 1-0-0}~~~

~

\textbf{Chronological Medical Findings:}

\textbf{May 10, 2021:} Complained of abdominal pain and weight loss.
Abdominal ultrasound revealed a mass in the pancreas. Referred to
oncologist. No symptoms attributable to carcinoid syndrome.

\textbf{June 5, 2021:} CT scan of the abdomen and pelvis: Revealed a
well-defined mass in the head of the pancreas measuring approximately
4.5 cm with surrounding lymphadenopathy. No signs of vascular invasion,
but multiple small hepatic lesions were identified, indicating
metastatic disease. No evidence of bowel obstruction, but slight
dilation of the pancreatic duct was observed.

\textbf{July 1, 2021:} Biopsy and molecular testing confirmed a
well-differentiated neuroendocrine tumor with a high tumor mutational
burden (TMB-H, >=10 mut/Mb, F1CDx assay), Grade 3. Initiated Capecitabine
and Temozolomide chemotherapy regimen on July 20, 2021.

\textbf{October 10, 2021:} Follow-up CT scan (chest, abdomen, and
pelvis): Partial response to treatment observed, with the primary
pancreatic mass reducing to 3.5 cm in diameter. Hepatic lesions showed
slight reduction in size and metabolic activity. No new metastatic
lesions were detected, but mild ascites persisted.

\textbf{January 1, 2022:} Continued Capecitabine and Temozolomide
therapy, with staging scans showing moderate disease progression.
Primary mass increased to 4.0 cm, with new peritoneal nodules.

\textbf{February 1, 2022:} Initiated Everolimus therapy due to
progression on previous regimen.

\textbf{August 1, 2022:} Follow-up MRI scan (abdomen and pelvis): MRI
indicated further disease progression, with the primary tumor enlarging
to 4.8 cm and increased involvement of adjacent hepatic structures.
Peritoneal nodules showed slight growth, and moderate ascites was
present. There was no evidence of bowel obstruction or significant
vascular invasion.

\textbf{November 15, 2022:} Tumor board recommends considering clinical
trial options due to limited response to standard therapies.

\textbf{December 1, 2022:} Detailed assessment of health status. ECOG
performance status 1. All routine labs, including liver and renal
function tests, within normal limits.

~
\begin{center}
\textbf{===== Patient 11.1.2 =====}~~~~~~~~~~~~~~~~~
\end{center}
\textbf{Patient Information}

Name: Smith, Anna

Born: 10.03.1980

Address: Another Avenue 89, Augsburg

\textbf{Overview of Tumor Diagnosis and Therapy}

\textbf{Tumor Diagnosis}

\textbf{Diagnosis}: UICC Stage IV well-differentiated, non-functioning
neuroendocrine tumor of the pancreas

Initial Detection: June 5, 2021, following symptoms of abdominal pain
and weight loss

Biopsy Date: July 1, 2021, well-differentiated neuroendocrine tumor of
the pancreas

Molecular Profile: High Tumor Mutational Burden (TMB-H, >=10 mut/Mb,
F1CDx assay)

\textbf{Therapy Overview}

\textbf{Initial Treatment:}

Chemotherapy: Began July 20, 2021, with Capecitabine and Temozolomide
regimen. Partial response noted after initial chemotherapy cycle
completed by October 10, 2021. Continued therapy until January 2022
(progressive disease).

Subsequent Treatment:

Further chemotherapy with Everolimus initiated on February 1, 2022.
Staging CT shows disease progression after 6 months.

\textbf{Current Status:}

~

ECOG 1

~

\textbf{Comorbidities}

Smoker 35 py

\colorbox{orange}{Alcohol dependence}

\colorbox{orange}{Hepatitis C}

Hypertension Stage 2

History of appendectomy 2001

~

\textbf{Medication}

\colorbox{orange}{Glecaprevir/Pibrentasvir 100mg/40mg 3-0-0}

Ramipril 10mg 1-0-0

Amlodipin 5mg 1-0-0

Hydrochlorothiazide 12.5mg 1-0-0~~~~~~~~

~

\textbf{Chronological Medical Findings:}

\textbf{May 10, 2021:} Complained of abdominal pain and weight loss.
Abdominal ultrasound revealed a mass in the pancreas. Referred to
oncologist. No symptoms attributable to carcinoid syndrome.

\textbf{June 5, 2021:} CT scan of the abdomen and pelvis: Revealed a
well-defined mass in the head of the pancreas measuring approximately
4.5 cm with surrounding lymphadenopathy. No signs of vascular invasion,
but multiple small hepatic lesions were identified, indicating
metastatic disease. No evidence of bowel obstruction, but slight
dilation of the pancreatic duct was observed.

July 1, 202\textbf{1:} Biopsy and molecular testing confirmed a
well-differentiated neuroendocrine tumor with a high tumor mutational
burden (TMB-H, >=10 mut/Mb, F1CDx assay), Grade 3. Initiated Capecitabine
and Temozolomide chemotherapy regimen on July 20, 2021.

\textbf{October 10, 2021:} Follow-up CT scan (chest, abdomen, and
pelvis): Partial response to treatment observed, with the primary
pancreatic mass reducing to 3.5 cm in diameter. Hepatic lesions showed
slight reduction in size and metabolic activity. No new metastatic
lesions were detected, but mild ascites persisted.

\textbf{January 1, 2022:} Continued Capecitabine and Temozolomide
therapy, with staging scans showing moderate disease progression.
Primary mass increased to 4.0 cm, with new peritoneal nodules.

\textbf{February 1, 2022:} Initiated Everolimus therapy due to
progression on previous regimen.

\textbf{August 1, 2022:} Follow-up MRI scan (abdomen and pelvis): MRI
indicated further disease progression, with the primary tumor enlarging
to 4.8 cm and increased involvement of adjacent hepatic structures.
Peritoneal nodules showed slight growth, and moderate ascites was
present. There was no evidence of bowel obstruction or significant
vascular invasion.

\textbf{November 15, 2022:} Tumor board recommends considering clinical
trial options due to limited response to standard therapies.

\textbf{December 1, 2022:} Detailed assessment of health status. ECOG
performance status 1. Routine labs show elevated liver enzymes: ALT 100
U/L, AST 89 U/L, total bilirubin 2.8 mg/dl, direct bilirubin 1.6 mg/dl,
Albumin 3.0 g/dl .

~~~~~~~~~~~~~~~~~~
\begin{center}
\textbf{===== Patient 11.1.3 =====}~~~~~~~~~~~~~~~~~~
\end{center}
\textbf{Patient Information}

Name: Smith, Anna

Born: 10.03.1980

Address: Another Avenue 89, Augsburg

\textbf{~}

\textbf{Overview of Tumor Diagnosis and Therapy}

\textbf{Tumor Diagnosis}

\textbf{Diagnosis}: UICC Stage IV well-differentiated, non-functioning
neuroendocrine tumor of the pancreas

Initial Detection: June 5, 2021, following symptoms of abdominal pain
and weight loss

Biopsy Date: July 1, 2021, well-differentiated neuroendocrine tumor of
the pancreas

Molecular Profile: High Tumor Mutational Burden (TMB-H, >=10 mut/Mb,
F1CDx assay)

~

\textbf{Therapy Overview}

\textbf{Initial Treatment:}

Chemotherapy: Began July 20, 2021, with Capecitabine and Temozolomide
regimen. Partial response noted after initial chemotherapy cycle
completed by October 10, 2021. Continued therapy until January 2022
(progressive disease).

Subsequent Treatment:

Further chemotherapy with Everolimus initiated on February 1, 2022.
Staging CT shows disease progression after 6 months.

Current Status: ECOG 1

\textbf{~}

\textbf{Comorbidities}

Hypertension Stage 2

History of appendectomy 2001

\colorbox{orange}{History of active tuberculosis 2003}

~

\textbf{Medication}

Ramipril 10mg 1-0-0

Amlodipin 5mg 1-0-0

Hydrochlorothiazide 12.5mg 1-0-0~~~~~~~~

~

\textbf{Chronological Medical Findings:}

\textbf{May 10, 2021:} Complained of abdominal pain and weight loss.
Abdominal ultrasound revealed a mass in the pancreas. Referred to
oncologist. No symptoms attributable to carcinoid syndrome.

\textbf{June 5, 2021:} CT scan of the abdomen and pelvis: Revealed a
well-defined mass in the head of the pancreas measuring approximately
4.5 cm with surrounding lymphadenopathy. No signs of vascular invasion,
but multiple small hepatic lesions were identified, indicating
metastatic disease. No evidence of bowel obstruction, but slight
dilation of the pancreatic duct was observed.

\textbf{July 1, 2021:} Biopsy and molecular testing confirmed a
well-differentiated neuroendocrine tumor with a high tumor mutational
burden (TMB-H, >=10 mut/Mb, F1CDx assay), Grade 3. Initiated Capecitabine
and Temozolomide chemotherapy regimen on July 20, 2021.

\textbf{October 10, 2021:} Follow-up CT scan (chest, abdomen, and
pelvis): Partial response to treatment observed, with the primary
pancreatic mass reducing to 3.5 cm in diameter. Hepatic lesions showed
slight reduction in size and metabolic activity. No new metastatic
lesions were detected, but mild ascites persisted.

\textbf{January 1, 2022:} Continued Capecitabine and Temozolomide
therapy, with staging scans showing moderate disease progression.
Primary mass increased to 4.0 cm, with new peritoneal nodules.

\textbf{February 1, 2022:} Initiated Everolimus therapy due to
progression on previous regimen.

\textbf{August 1, 2022:} Follow-up MRI scan (abdomen and pelvis): MRI
indicated further disease progression, with the primary tumor enlarging
to 4.8 cm and increased involvement of adjacent hepatic structures.
Peritoneal nodules showed slight growth, and moderate ascites was
present. There was no evidence of bowel obstruction or significant
vascular invasion.

\colorbox{orange}{\textbf{August 5, 2022:} Brain MRI. Multiple metastases, specifically three lesions}\\
\colorbox{orange}{in the right hemisphere: two in the right parietal lobe,
and one in the}\\
\colorbox{orange}{right occipital lobe.}
Incidental findings included
scattered white matter hyperintensities consistent with chronic
microvascular ischemic changes.

\textbf{November 15, 2022:} Tumor board recommends considering clinical
trial options due to limited response to standard therapies.

\textbf{December 1, 2022:} Detailed assessment of health status. ECOG
performance status 1. All routine labs, including liver and renal
function tests, within normal limits.

\end{document}


\begin{longtable}{@{} l p{12cm} @{}} 

\caption*{{\Large Supplementary Table 1}}\\
\toprule
\textbf{No.} & \textbf{Prompts and Expected Trial} \\
\midrule
\endfirsthead

\toprule
\textbf{No.} & \textbf{Prompts and Expected Trial} \\
\midrule
\endhead

1 & \textbf{Prompt:} Please find a Phase 1 clinical trial targeting the patients KRAS G13D mutation in the USA.
\newline
\textbf{Target Trial:} NCT05254184
\\
\bottomrule

2 & \textbf{Prompt:} Find a clinical trial for urachal adenocarcinoma in the United States. Ignore all mutations.
\newline
\textbf{Target Trial:} NCT05756569
\\
\bottomrule

3 & \textbf{Prompt:} Find a clinical trial for urachal adenocarcinoma in the United States. Ignore all mutations.
\newline
\textbf{Target Trial:} NCT03866382
\\
\bottomrule

4 & \textbf{Prompt:} Please find a clinical trial for the thymic cancer in Germany, that would fit the patient. Also check for trials targeting the BRCA mutation.
\newline
\textbf{Target Trial:} NCT01078662
\\
\bottomrule

5 & \textbf{Prompt:} Please find a clinical trial in Dresden for the patient targeting PIK3CA and MAPK1 mutations.
\newline
\textbf{Target Trial:} NCT04551521
\\
\bottomrule

6 & \textbf{Prompt:} Could you please find a clinical trial for the patient here in Germany focusing on MET as a target?
\newline
\textbf{Target Trial:} NCT05488314
\\
\bottomrule

7 & \textbf{Prompt:} Please find a clincial trial targeting the patients tumor and its KRAS mutation in Germany.
\newline
\textbf{Target Trial:} NCT05358249
\\
\bottomrule

8 & \textbf{Prompt:} Please find a clincial trial targeting the patients tumor and its KRAS mutation in Germany.
\newline
\textbf{Target Trial:} NCT04685135
\\
\bottomrule

9 & \textbf{Prompt:} Please find a clincial trial targeting the patients tumor and its KRAS mutation in Italy.
\newline
\textbf{Target Trial:} NCT05789082
\\
\bottomrule

10 & \textbf{Prompt:} Please find a clincial trial targeting the patients tumor and its KRAS mutation in Germany.
\newline
\textbf{Target Trial:} NCT04699188
\\
\bottomrule

11 & \textbf{Prompt:} The patient will move to France soon. Can you find a trial for the patient there?
\newline
\textbf{Target Trial:} NCT06054555
\\
\bottomrule

12 & \textbf{Prompt:} Could you please find a clincial trial for the patients molecular tumor alterations in Germany?
\newline
\textbf{Target Trial:} NCT04526106
\\
\bottomrule

13 & \textbf{Prompt:} Please find a trial by checking all of the patients molecular alterations in Germany.
\newline
\textbf{Target Trial:} NCT04551521
\\
\bottomrule

14 & \textbf{Prompt:} Please find a trial by checking all of the patients molecular alterations in Germany.
\newline
\textbf{Target Trial:} NCT02628067
\\
\bottomrule

15 & \textbf{Prompt:} Please find a clinical trial (Phase 1) for the patient's lung cancer in Germany.
\newline
\textbf{Target Trial:} NCT04077463
\\
\bottomrule
\end{longtable}